\documentclass[10pt,twocolumn,letterpaper]{article}

\usepackage{iccv}
\usepackage{times}
\usepackage{epsfig}
\usepackage{graphicx}
\usepackage{amsmath}
\usepackage{amssymb}
\usepackage[accsupp]{axessibility}  

\usepackage{xparse}
\usepackage[numbers]{natbib}
\usepackage[resetlabels,labeled]{multibib}
\newcites{S}{References for Supplement}
\newcites{M}{References}
\newcommand{\citeApp}[1]{\citeS{#1_2}}
\newcommand{\citeMain}[1]{\cite{#1}}

\let\oriCiteS\citeS
\RenewDocumentCommand{\citeS}{O{} O{} m}{%
  \renewcommand{\citenumfont}[1]{S##1}%
  \oriCiteS[#1][#2]{#3}%
  \renewcommand{\citenumfont}[1]{##1}%
}

\usepackage{subcaption}

\usepackage[pagebackref=true,breaklinks=true,letterpaper=true,colorlinks,bookmarks=false,pdftex]{hyperref}
\usepackage{cleveref}
\crefname{section}{Sec.}{Secs.}
\crefname{figure}{Fig.}{Figs.}
\crefname{table}{Tab.}{Tabs.}
\crefname{equation}{Eq.}{Eqs.}
\Crefname{section}{Section}{Sections}

\iccvfinalcopy 

\def\iccvPaperID{10062} 

\ificcvfinal\fi

\input{commands}

\begin{document}

\title{Studying How to Efficiently and Effectively Guide Models with Explanations}

\author{Sukrut Rao$^*$, Moritz Böhle$^*$, Amin Parchami-Araghi, Bernt Schiele\\
Max Planck Institute for Informatics, Saarland Informatics Campus, Saarbrücken, Germany\\
{\tt\small \{sukrut.rao,mboehle,mparcham,schiele\}@mpi-inf.mpg.de}
}

\maketitle
\def\thefootnote{*}\footnotetext{Equal contribution.}
\def\thefootnote{\arabic{footnote}}
\ificcvfinal\thispagestyle{empty}\fi

\begin{abstract}
Despite being highly performant, deep neural networks
might base their decisions on features that spuriously correlate with the provided labels, thus hurting generalization.
To mitigate this, `model guidance' has recently gained popularity, \ie the idea of regularizing the models' explanations to ensure that they are ``right for the right reasons''
\citeMain{ross2017right}.
While various techniques to achieve such model guidance have been proposed,
experimental validation of these approaches has thus far 
been limited to relatively simple and / or synthetic datasets.
To better understand the \textbf{effectiveness}
of the various design choices that have been explored in the context of model guidance,
in this work we conduct an in-depth evaluation across various loss functions, attribution methods, models, and `guidance depths' on the \voc and \coco datasets. 
As annotation costs for model guidance can limit its applicability, we also place a particular focus on \textbf{efficiency}. Specifically, we guide the models via bounding box annotations, which are much cheaper to obtain than the commonly used segmentation masks, and evaluate the robustness of model guidance under limited (\eg with only $1\%$ of annotated images) or overly coarse annotations.
Further, we
propose using the EPG score as an additional evaluation metric and loss function (`Energy loss'). We show that optimizing for the \energyloss loss leads to models that exhibit a distinct focus on object-specific features, despite only using bounding box annotations that also include background regions.
Lastly, we show that such model guidance can improve generalization under distribution shifts. {Code available at: \href{https://github.com/sukrutrao/Model-Guidance}{https://github.com/sukrutrao/Model-Guidance}}
\end{abstract}

\vspace{-.75em}
\begin{figure}[!ht]
    \centering 
    \begin{subfigure}[c]{.9\columnwidth}
    \includegraphics[width=\textwidth]{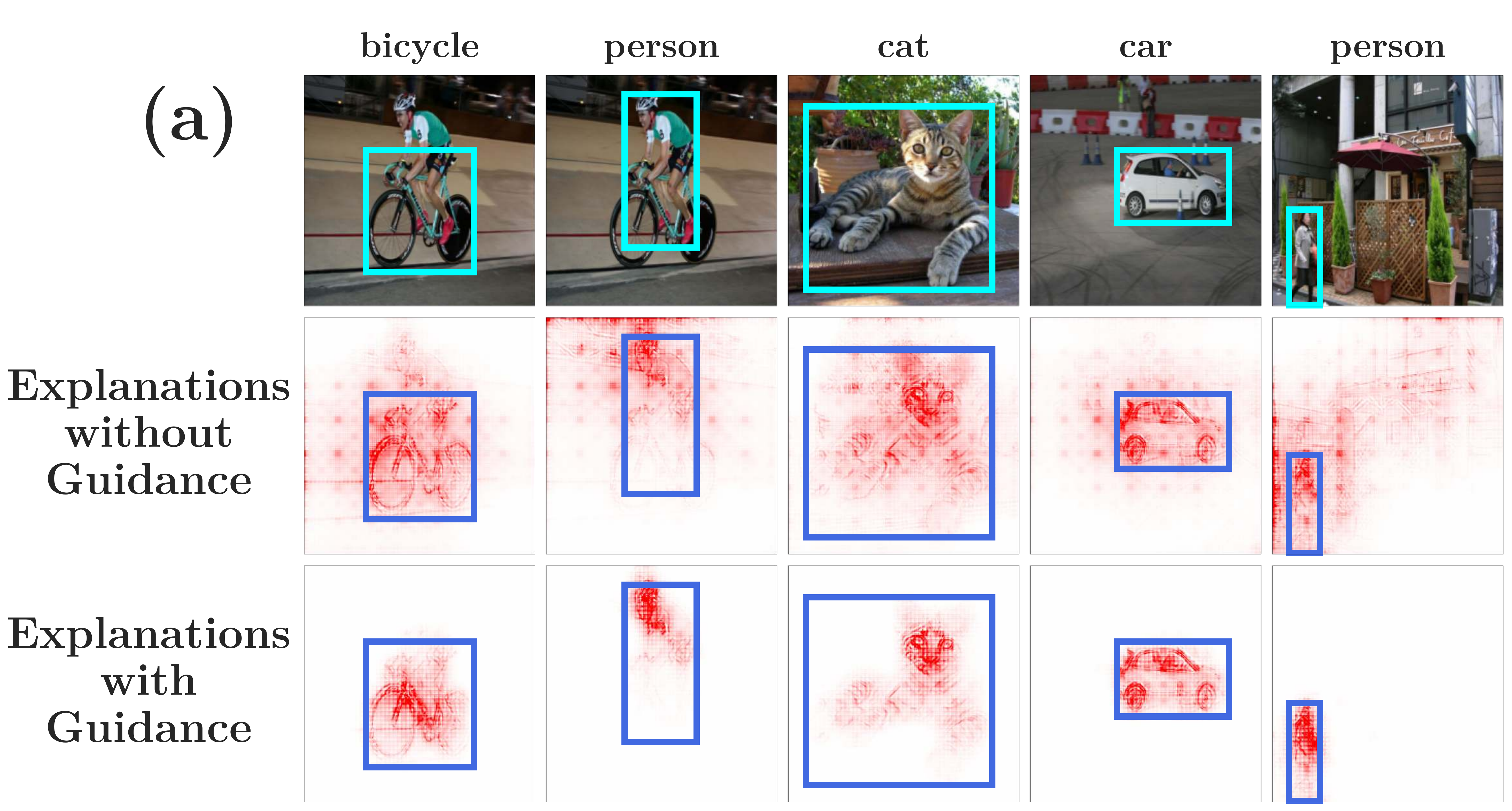}
    \end{subfigure}
    \vspace{.15cm}

    \begin{subfigure}[c]{\columnwidth}
    \includegraphics[width=\textwidth]{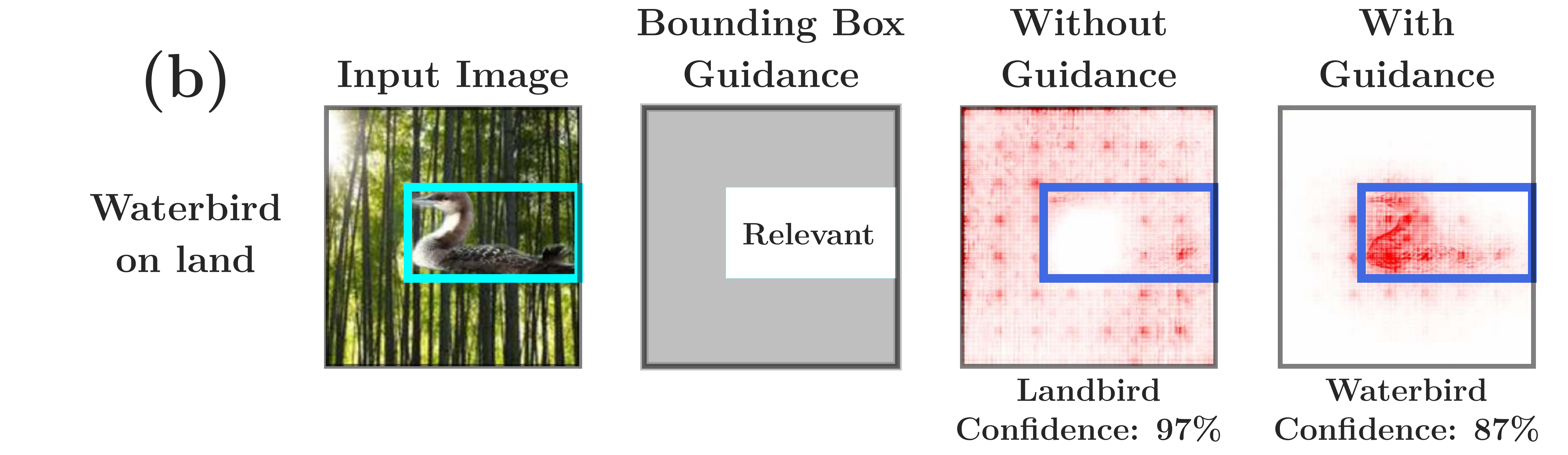}
    \end{subfigure}
    \caption{\textbf{(a) Model guidance increases object focus.}
    Models may rely on irrelevant background features or spurious correlations (\eg presence of person provides positive evidence for bicycle, center row, col. 1). Guiding the model via bounding box annotations can mitigate this and consistently increases the focus on object features (bottom row).
    \textbf{(b) Model guidance can improve accuracy.} In the presence of spurious correlations in the training data, non-guided models might focus on the wrong features. In the example image in (b), the waterbird is incorrectly classified to be a landbird due to the background (col.~3). Guiding the model via bounding box annotation (as shown in col.~2), the model can be guided to focus on the bird features for classification (col.~4).
    }
    \label{fig:teaser}
\end{figure}
\section{Introduction}
\label{sec:introduction}

\begin{figure*}[!ht]
    \centering 
    \includegraphics[width=.95\textwidth]{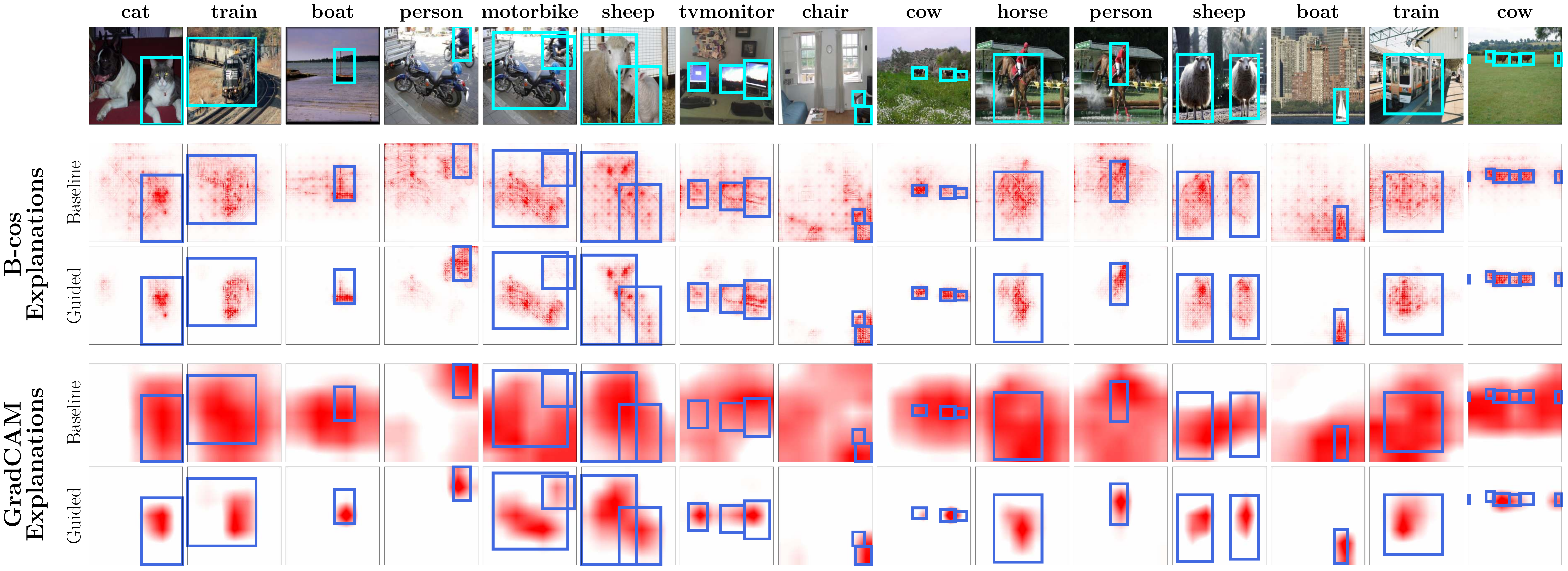}
    \caption{\textbf{Qualitative results of model guidance.} We show model-inherent \bcos explanations (input layer) of a \bcos \resnet and \gradcam explanations (final layer) of a conventional \resnet before (`Standard') and after optimization (`Guided') for images from the \vocs test set, using our proposed \epgloss loss (\cref{eq:energyloss}). Guiding the model via bounding box annotations consistently increases the focus on object features for both methods. 
    Specifically, we find that background attributions are consistently suppressed in both cases.
    }
    \label{fig:teaser2}
    \vspace{-1em}
\end{figure*}
 
Deep neural networks (DNNs) excel at learning predictive features that allow them to correctly classify a set of training images with ease. The features learnt on the training set, however, do not necessarily transfer to unseen images: \ie, instead of learning the actual class-relevant features, DNNs might memorize individual images {(\cf.~\citeMain{feldman2020neural})}
or exploit spurious correlations in the training data (cf.~\citeMain{xiao2021noise}).
For example, if bikes are highly correlated with people in the training data, a model might learn to associate the presence of a person in an image as positive evidence for a bike {(\eg \cref{fig:teaser}{\color{red}a}, col.\ 1, rows 1-2)}, which can limit how well it generalizes. Similarly, a bird classifier might rely on background features from the bird's habitat, and fail to correctly classify in a different habitat (cf.~\cref{fig:teaser}{\color{red}b} cols.\ 1-3 and \citeMain{petryk2022guiding}).

{To detect 
such behaviour, recent advances in model interpretability have provided attribution methods (\eg \citeMain{selvaraju2017grad,sundararajan2017axiomatic,shrikumar2017learning,bohle2022b})
to understand a model's reasoning. These methods typically provide attention maps that highlight regions of importance in an input to explain the model's decisions and can help identify incorrect reasoning such as reliance on spurious or irrelevant features, see for example \cref{fig:teaser}{\color{red}b}.}

As many attribution methods are in fact themselves differentiable (\eg \citeMain{shrikumar2017learning,sundararajan2017axiomatic,selvaraju2017grad,bohle2022b}), recent work \citeMain{ross2017right,shen2021human,gao2022aligning,gao2022res,teso2019explanatory,teso2019toward} has explored the idea of using them to guide the models to
make them ``right for the right reasons'' \citeMain{ross2017right}. Specifically, models can be guided by jointly optimizing for correct classification as well as for attributing importance to regions deemed relevant by humans.
This can help the model focus on the relevant features of a class, and correct errors in reasoning (\cref{fig:teaser}{\color{red}b}, col.\ 4). Such guidance has the added benefit of providing well-localized 
explanations that are thus easier to understand for end users (\eg \cref{fig:teaser2}).
 
While model guidance has shown promising results, a detailed study of how to do this most \emph{effectively} is crucially missing. In particular, model guidance has so far been studied for a limited set of attribution methods and models and usually on relatively simple and/or synthetic datasets; further, the evaluation settings between approaches can significantly differ, which makes a fair comparison difficult.

Therefore, in this work, we perform an in-depth evaluation of model guidance on large scale, real-world datasets, to better understand the effectiveness of a variety of design choices.  
Specifically, we evaluate model guidance along the following dimensions: the model architecture,
the guidance \emph{depth}\footnote{The layer at which guidance is applied, \eg typically at the last convolutional layer for \gradcam \cite{selvaraju2017grad} or the first layer for \ixg\cite{shrikumar2017learning}.}, the attribution method,
and the loss function. In this context, we propose using the EPG score \cite{wang2020score}---an evaluation metric that has thus far been used to evaluate the quality of attribution methods---as an additional loss function (which we call the \energyloss loss) as it is fully differentiable.

Further, as annotation costs can be a major hurdle for making model guidance practical, we place a particular focus on \emph{efficient} guidance. Specifically, we use bounding boxes instead of semantic segmentation masks, and evaluate the robustness of guidance techniques under limited or overly coarse annotations to reduce data collection costs.

We find that our \energyloss loss lends itself well to those settings. On the one hand, it exhibits a high degree of robustness to limited or noisy bounding box annotations (cf.~\cref{fig:coarse_annotations,fig:lim_local_input}). On the other hand, despite the coarseness of bounding box guidance, it maintains a clear focus on object-specific features inside the bounding boxes, see \cref{fig:teaser}{\color{red}a}, row 3. In contrast, prior approaches often regularize for a uniform distribution of the attribution values inside the annotation masks, and thus
tend to exhibit much lower attribution granularity (cf.~\cref{fig:loss_comp}).

\myparagraph{Contributions.} \textbf{(1)} We perform an in-depth evaluation of model guidance on challenging large scale, multi-label classification datasets (\voc \citeMain{everingham2009pascal}, \coco \citeMain{lin2014microsoft}), assessing the impact of attribution methods, model architectures, guidance depths, and loss functions. Further, we show that, despite being relatively coarse,  bounding box supervision can provide sufficient guidance to the models whilst being much cheaper to obtain than semantic segmentation masks. 
\textbf{(2)} We propose using the Energy Pointing Game (EPG) score \citeMain{wang2020score} as an alternative to the IoU metric for evaluating the effectiveness of such guidance and show that the EPG score constitutes a good loss function for model guidance, particularly when using bounding boxes. \textbf{(3)} We show that model guidance can be performed cost-effectively by using annotation masks that are noisy or are available for only a small fraction (\eg $1\%$) of the training data. \textbf{(4)} We show through experiments on the \waterbirds dataset \citeMain{sagawa2019distributionally,petryk2022guiding} that model guidance with a small number of annotations suffices to improve the model's generalization under distribution shifts at test time.
\section{Related Work}
\label{sec:related}

\myparagraph{Attribution Methods} \citeMain{simonyan2013deep,springenberg2014striving,sundararajan2017axiomatic,shrikumar2017learning,selvaraju2017grad,wang2020score,ramaswamy2020ablation,jiang2021layercam,chattopadhyay2018grad,petsiuk2018rise,fong2017interpretable,zeiler2013visualizing,ribeiro2016should,dabkowski2017real,bach2015pixel} are often used to explain black-box models by generating heatmaps that highlight input regions important to the model's decision. However, such methods are often not faithful to the model \citeMain{adebayo2018sanity,rao2022towards,kim2021sanity,zhou2022feature,adebayo2022post} and risk misleading users. Recent work proposes inherently interpretable models \citeMain{boehle2021convolutional,bohle2022b} that address this by providing model-faithful explanations by design. In our work, we use both popular post-hoc and model-inherent attribution methods to guide models and discuss their effectiveness.

\myparagraph{Attribution Priors:} Several approaches have been proposed for training better models by enforcing desirable properties on their attributions. These include enforcing consistency against augmentations \citeMain{pillai2021explainable,pillai2022consistent,guo2019visual}, smoothness \citeMain{erion2021improving,moshe2022improving,kiritoshi2019l1}, separation of classes \citeMain{zhang2022correct,pillai2022consistent,sun2020fixing,nakka2020towards,singh2020don}, or constraining the model's attention \citeMain{fukui2019attention,asgari2022masktune}.
In contrast, in this work, we focus on providing explicit human guidance to the model using bounding box annotations. {This constitutes more explicit guidance but allows fine-grained control over the model's reasoning even with few annotations.}

\myparagraph{Model Guidance:} In contrast to the indirect regularization effect achieved by attribution priors, various approaches have been proposed (cf.~\citeMain{friedrich2022typology,teso2022leveraging}) to actively guide models by regularizing their attributions, for tasks such as classification \citeMain{ross2017right,gao2022aligning,gao2022res,rieger2020interpretations,petryk2022guiding,hagos2022identifying,teney2020learning,mitsuhara2019embedding,teso2019explanatory,teso2019toward,schramowski2020making,shao2021right,linsley2018learning,shen2021human,yang2022improving,fel2022harmonizing}, segmentation \citeMain{li2018tell}, VQA \citeMain{selvaraju2019taking,teney2020learning}, and knowledge distillation \citeMain{fernandes2022learning}. The goal of such approaches is not only to improve performance, but also make sure that the model is ``right for the right reasons'' \citeMain{ross2017right}. For classifiers, this typically involves jointly optimizing both for classification performance and localization to object features.
While various benefits of model guidance have been reported, most prior work evaluate on simple datasets \citeMain{ross2017right,shao2021right,gao2022aligning,gao2022res} and, thus far, no common evaluation setting has emerged. Recently, \citeMain{chefer2022optimizing} has extended model guidance to ImageNet, showing that its benefits can scale to large scale problems. In contrast to \citeMain{chefer2022optimizing}, who investigated one particular attribution method \citeMain{chefer2021beyond}, our focus lies on a better understanding of the impact of the different design choices for model guidance.

To distill the most effective techniques for model guidance, in this work, we conduct an in-depth evaluation on challenging,commonly used real-world multi-label classification datasets (\voc, \coco). 
Specifically, we perform a comprehensive comparison across multiple dimensions of interest: the loss function, the model architecture, the guidance depth, and the attribution method.
For this, we evaluate the localization losses introduced in the closest related work, \ie \rrr \citeMain{ross2017right}, \haics \citeMain{shen2021human}, and \gradia \citeMain{gao2022aligning}; additionally, we propose using the EPG metric \cite{wang2020score} as a loss function and show that it has various desirable properties, in particular when guiding models via bounding box annotations. 

{Finally, model guidance has also been used to mitigate reliance on spurious features using language guidance \citeMain{petryk2022guiding}, and we show that using a small number of coarse bounding box annotations can be similarly effective.}

\myparagraph{Evaluating Model Guidance:} The benefits of model guidance have typically been shown via improvements in classification performance (\eg \citeMain{ross2017right,rieger2020interpretations}) or an increase in \iou between object masks and attribution maps (\eg \citeMain{gao2022res,li2018tell}). In addition to these metrics, we also evaluate on the \energypg metric \citeMain{wang2020score}, which has thus far only been used to evaluate the quality of the attribution methods themselves. We further show that it lends itself well to being used as a guidance loss, as it places only minor constraints on the model, and, in contrast to the IoU metric, it is fully differentiable.
\section{Guiding Models Using Attributions}
\label{sec:method}

\begin{figure}
    \centering
    \includegraphics[width=\linewidth]{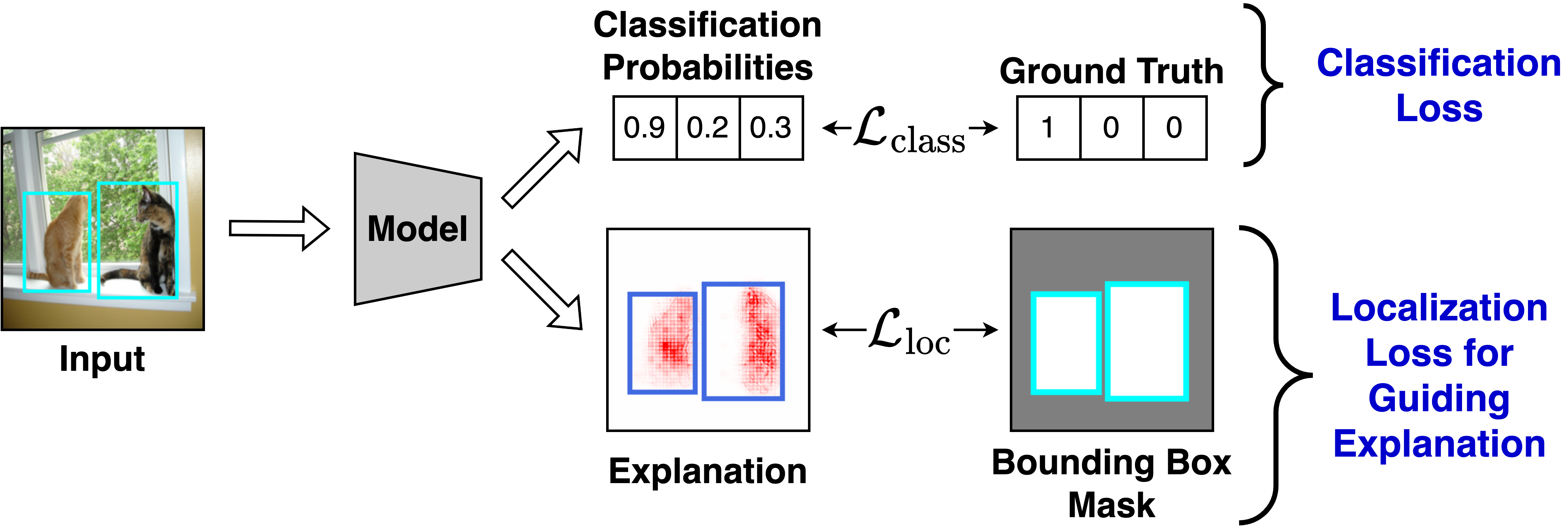}
    \caption{\textbf{Model guidance overview.} 
    We jointly optimize for classification ($\mathcal L_\text{class}$) and localization of attributions to human-annotated bounding boxes ($\mathcal L_\text{loc}$), to guide the model to focus on object features. Various localization loss functions can be used, see \cref{sec:method:losses}.
    }
    \label{fig:pipeline}
\end{figure}

In this section, we provide an overview of the model guidance approach that jointly optimizes for classification and localization (\cref{sec:method:procedure}). Specifically, we describe the attribution methods (\cref{sec:method:attributions}), metrics (\cref{sec:method:metrics}), and localization loss functions (\cref{sec:method:losses}) that we evaluate in \cref{sec:results}.
In \cref{sec:method:efficient} we discuss our strategy to train for localization in the presence of multiple ground truth classes.

\myparagraph{Notation:} We consider a multi-label classification problem with $K$ classes with $X\myin\mathbb{R}^{C\times H\times W}$ the input image and $y\myin\{0,1\}^K$ the one-hot encoding of the image labels. With $A_k\myin\mathbb{R}^{H\times W}$ we denote an attribution map for a class $k$ for $X$ using a classifier $f$; $A_{k}^+$ denotes the positive component of the attributions, $\hat{A}_k=\frac{A_k}{\max(\text{abs}(A_k))}$ normalized attributions, and $\hat{A}_{k}^+=\frac{A_{k}^+}{\max(A_{k}^+)}$ normalized positive attributions. Finally, $M_k \myin \{0,1\}^{H\times W}$ denotes the binary mask for class $k$, which is given by the union of bounding boxes of all occurrences of class $k$ in $X$.

\subsection{Model Guidance Procedure}\label{sec:method:procedure}

Following prior work (\eg \citeMain{ross2017right,shen2021human,gao2022aligning,gao2022res}), the model is trained jointly for classification and localization (cf.~\cref{fig:pipeline}):
\begin{equation}\label{eq:overall}
\textstyle
    \mathcal{L}=\mathcal{L}_\text{class} + \lambda_\text{loc}\mathcal{L}_\text{loc}\;.
\end{equation}
\Ie, the loss consists of a classification loss ($\mathcal{L}_\text{class}$), for which we use binary cross-entropy, and a localization loss ($\mathcal{L}_\text{loc}$), which we discuss in \cref{sec:method:losses}; here, the hyperparameter $\lambda_\text{loc}$ controls the weight given to each of the objectives.

\subsection{Attribution Methods}
\label{sec:method:attributions}

In contrast to prior work that typically use \gradcam \citeMain{selvaraju2017grad} attributions, we perform an evaluation over a selection of popularly used differentiable\footnote{Differentiability is necessary for optimizing attributions via gradient descent, so non-differentiable methods (\eg \citeMain{ribeiro2016should,petsiuk2018rise}) are not considered.} attribution methods which have been shown to localize well \citeMain{rao2022towards}: \ixg \citeMain{shrikumar2017learning}, \intgrad \citeMain{sundararajan2017axiomatic}, and \gradcam \citeMain{selvaraju2017grad}. We further evaluate model-inherent explanations of the recently proposed \bcos models \citeMain{bohle2022b}. To ensure comparability across attribution methods \citeMain{rao2022towards}, we evaluate all attribution methods at the input, various intermediate, and the final spatial layer. 

\myparagraph{\ixg \citeMain{shrikumar2017learning}} computes the element-wise product $\odot$ of the input and the gradients of the $k$-th output w.r.t.~the input, \ie $X\odot\nabla_X f_k(X)$. For piece-wise linear models such as DNNs with ReLU activations \citeMain{nair2010rectified}, this faithfully computes the linear contributions of a given input pixel to the model output.

\myparagraph{\gradcam \citeMain{selvaraju2017grad}} computes importance attributions as a ReLU-thresholded, gradient-weighted sum of activation maps. In detail, it is given by
$\text{ReLU}(\sum_c \alpha_c^k \odot U_c)$ with $c$ denoting the channel dimension, and $\alpha^k$ the average-pooled gradients of the output for class $k$ with respect to the activations $U$ of the last convolutional layer in the model.

\myparagraph{\intgrad \citeMain{sundararajan2017axiomatic}} takes an axiomatic approach and is formulated as the integral of gradients over a straight line path from a baseline input to the given input $X$. Approximating this integral requires several gradient computations, making it computationally expensive for use in model guidance. To alleviate this, when optimizing with \intgrad, we use the recently proposed \xdnn models \citeMain{hesse2021fast} that 
allow for an exact computation of \intgrad in a single backward pass.

\myparagraph{\bcos \citeMain{bohle2022b}} attributions are generated using the inherently-interpretable \bcos networks, which promote alignment between the input $\mathbf x$ and a dynamic weight matrix $\mathbf W(\mathbf x)$ during optimization. 
In our experiments, 
we use the contribution maps given by the element-wise product of the dynamic weights with the input ($\mathbf W^T_k(\mathbf x)\odot \mathbf x$), which faithfully represent the contribution of each pixel to class $k$. To be able to guide \bcos models, we developed a differentiable implementation of \bcos explanations, see supplement.

\subsection{Evaluation Metrics}
\label{sec:method:metrics}

We evaluate the models' performance on both our training objectives: classification and localization. For classification, we use the F1 
score and mean average precision (\map). We discuss the localization metrics below.

\myparagraph{Intersection over Union (IoU)} is a commonly used metric (cf.~\citeMain{gao2022res}) that computes the intersection between the ground truth annotation masks and the binarized attribution maps, normalized by their union; for binarization, a threshold parameter needs to be chosen. 
In this work, the ground truth masks are taken to be the union of
all bounding boxes of a class in the image and, following prior work \citeMain{fong2017interpretable}, the threshold parameter is selected via a heldout set. 

\myparagraph{Energy-based Pointing Game (EPG) \citeMain{wang2020score}} measures the concentration of attribution energy within the mask, \ie the fraction of positive attributions inside the bounding boxes:
\begin{equation}\label{eq:epg}
    \text{EPG}_k = \frac{\sum_{h=1}^H\sum_{w=1}^W M_{k,hw} A^+_{k,hw}}{\sum_{h=1}^H\sum_{w=1}^W A^+_{k,hw}}\;.
\end{equation}
In contrast to \iou,
\epg more faithfully takes into account the relative importance given to each input region,
since it does not binarize the attributions. Like \iou, the scores lie in $[0,1]$, with higher scores indicating better localization. 

\subsection{Localization Losses}\label{sec:method:losses}
We evaluate the most commonly used localization losses ($\mathcal{L}_\text{loc}$ in \cref{eq:overall}) from prior work. We describe these losses as applied on attribution maps of an image for a single class $k$, as well as the proposed EPG-derived \energyloss loss.

\myparagraph{\loneloss loss (\citeMain{gao2022aligning,gao2022res}, \cref{eq:lone})}
minimizes the $L_1$ distance between annotation masks and normalized positive attributions $\hat A_{k}^+$, guiding the model towards uniform attributions inside the mask and suppressing attributions outside of it. 
\begin{equation}\label{eq:lone}
    \textstyle
    \mathcal{L}_{\text{loc},k} = \frac{1}{H\times W}\sum_{h=1}^H\sum_{w=1}^W\lVert M_{k,hw} - \hat{A}_{k,hw}^+ \rVert_1
\end{equation}

\myparagraph{Per-pixel cross entropy (\ppceloss) loss (\citeMain{shen2021human}, \cref{eq:ppce})}
applies a binary cross entropy loss between the mask and the normalized positive annotations $\hat A_{k}^+$, thus guiding the model to maximize the attributions inside the mask:
\begin{equation}\label{eq:ppce}
\textstyle
    \mathcal{L}_{\text{loc},k} = -\frac{1}{\lVert M_k \rVert_1}\sum_{h=1}^H\sum_{w=1}^W M_{k,hw}\log(\hat{A}_{k,hw}^+) \;.
\end{equation}
As \ppceloss does not constrain attributions outside the mask, there is no explicit pressure to avoid spurious features.
    
\myparagraph{\rrrloss loss (\citeMain{ross2017right}, \cref{eq:rrr}).}
\citeMain{ross2017right} introduced the RRR loss to regularize the normalized input gradients $\hat{A}_{k,hw}$ as 
\begin{equation}\label{eq:rrr}
    \textstyle \mathcal{L}_{\text{loc},k} = \sum_{h=1}^H\sum_{w=1}^W (1-M_{k,hw}) \hat{A}_{k,hw}^2 \;.
\end{equation}
To extend it to our setting, we 
take $\hat{A}_{k,hw}$ to be given by an arbitrary attribution method (\eg \intgrad); 
we denote this generalized version by \rrrloss.
In contrast to the \ppceloss loss, \rrrloss only regularizes attributions \emph{outside} the ground truth masks. While it thus does not introduce a uniformity prior similar to the \loneloss loss, it also does not explicitly promote high importance attributions inside the masks.

\myparagraph{\energyloss Loss.}\label{sec:method:energyloss}
In addition to the losses described in prior work, we propose to also evaluate using the \epg score (\citeMain{wang2020score}, \cref{eq:epg}) as a loss function for model guidance, as it is fully differentiable. In particular, we simply define it as
\begin{equation}\label{eq:energyloss}
\textstyle
    \mathcal{L}_{\text{loc},k} = -\text{EPG}_k.
\end{equation}
Unlike existing localization losses that either (i) do not constrain attributions across the entire input (\rrrloss, \ppceloss), or (ii) force the model to attribute uniformly within the mask even if it includes irrelevant background regions (\loneloss, \ppceloss), maximizing the \epg score jointly optimizes for higher attribution energy within the mask and lower attribution energy outside the mask. 
By not enforcing a uniformity prior, we find that the \energyloss loss is able to provide effective guidance while allowing the model to learn freely what to focus on within the bounding boxes (\cref{sec:results}).

\subsection{Efficient Optimization}\label{sec:method:efficient}
In contrast to prior work \citeMain{ross2017right,shen2021human,gao2022aligning,gao2022res}, we perform model guidance on a multi-label classification setting, and consequently there are multiple ground truth classes whose attribution localization could be optimized. Computing and optimizing for several attributions within an image would add a significant overhead to the computational cost of training (multiple backward passes).
Hence, for efficiency, we sample one ground truth class $k$ per image at random for every batch and only optimize for localization of that class, \ie, $\mathcal{L}_\text{loc}\myeq\mathcal{L}_{\text{loc},k}$. We find that this still provides effective model guidance while keeping the training cost tractable.
\section{Experimental Setup}
\label{sec:experiments}
\begin{figure}[t]
    \centering 
    \hspace{-.04\columnwidth}
    \includegraphics[width=1.025\columnwidth]{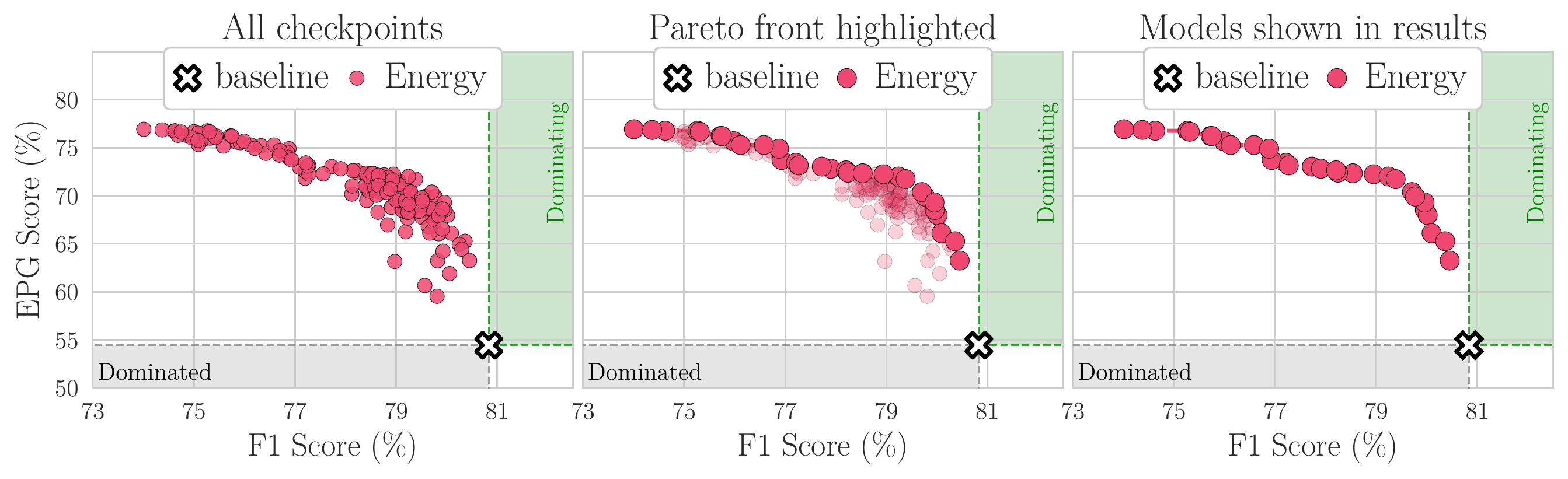}
    \caption{\textbf{Selecting models for evaluation.} For each configuration, we evaluate every model at every checkpoint and measure its performance across various metrics (\fone, \epg, \iou) on the validation set; \ie every point in the left graph corresponds to one model (for \bcos models optimized via the \epgloss loss at the input layer). Instead of evaluating a single model on the test set, we evaluate \emph{all Pareto-dominant} models, as indicated in the center and right plot.
    }
    \label{fig:pareto_example}
\end{figure}

In this section, we describe our experimental setup
and how we select the best models across metrics; for full details, see supplement.
We evaluate across all possible
choices for each category, and discuss our results in \cref{sec:results}. 

\myparagraph{Datasets:} We evaluate on \voc \citeMain{everingham2009pascal} and \coco \citeMain{lin2014microsoft} for multi-label image classification. {In \cref{sec:results:waterbirds}, to understand the effectiveness of model guidance in mitigating spurious correlations, we also evaluate on the synthetically constructed Waterbirds-100 dataset \citeMain{sagawa2019distributionally,petryk2022guiding}, where landbirds are perfectly correlated with land backgrounds on the training and validation sets, but are equally likely to occur on land or water in the test set (similar for waterbirds and water). With this dataset, we evaluate model guidance for suppressing undesired features.}

\myparagraph{Attribution Methods and Architectures:} As described in \cref{sec:method:attributions}, we evaluate with \ixg \citeMain{shrikumar2017learning}, \intgrad \citeMain{sundararajan2017axiomatic}, \bcos \citeMain{bohle2022b,bohle2023b}, and \gradcam \citeMain{selvaraju2017grad} using models with a \resnet \citeMain{he2016deep} backbone. For \intgrad, we use an \xdnn \resnet \citeMain{hesse2021fast} to reduce the computational cost, and a \bcos \resnet for the \bcos attributions. To emphasize that the results generalize across different backbones, we further provide results for a \bcos ViT-S \citeMain{Dosovitskiy2021image,bohle2023b} and a \bcos DenseNet-121 \citeMain{huang2017densely,bohle2023b}. 
We evaluate optimizing the attributions at different network layers, such as at the input image and the last convolutional layers' output\footnote{As typically used in \ixg (input) and \gradcam (final) respectively.}, as well as at multiple intermediate layers. Within the main paper, we highlight some of the most representative and insightful results, the full set of results can be found in the supplement.
All models were pretrained on \imagenet \citeMain{imagenet}, and model guidance was applied when fine-tuning the models on the target dataset.

\myparagraph{Localization Losses:} As described in \cref{sec:method:losses}, we compare four localization losses in our evaluation: (i) \energyloss, (ii) \loneloss \citeMain{gao2022aligning,gao2022res}, (iii) \ppceloss \citeMain{shen2021human}, and (iv) \rrrloss (cf.~\cref{sec:method:losses}, \citeMain{ross2017right}).

\myparagraph{Evaluation Metrics:} As discussed in \cref{sec:method:metrics}, we evaluate both for classification and localization performance of the models. For classification, we report the F1 scores, similar results with \map scores can be found in the supplement. For localization, we evaluate using the \epg and \iou scores.

\myparagraph{Selecting the best models:} As we evaluate for two distinct objectives (classification + localization), it is not trivial to decide which models perform `the best', \eg a model that provides the best classification performance might provide significantly worse localization than a model that provides only slightly lower classification performance. 
Finding the right balance and deciding which of those models in fact constitutes the `better' model depends on the preference of the end user. 
Hence, instead of selecting models based on a single metric, we select the set of Pareto-dominant models \citeMain{pareto1894massimo,pareto2008maximum,backhaus1980pareto} across three metrics---F1, \epg, and \iou---for each training configuration, as defined by a combination of attribution method, layer, and loss. Specifically, as shown in \cref{fig:pareto_example}, we train each configuration using three different choices of $\lambda_\text{loc}$, and select the set of Pareto-dominant models among all checkpoints (epochs and $\lambda_\text{loc}$). This provides a more holistic view of the general trends on the effectiveness of model guidance for each configuration.
\section{Experimental Results}
\label{sec:results}

\begin{figure*}[t]
    \centering 

    \begin{subfigure}[c]{.95\textwidth}
    \includegraphics[width=\textwidth]{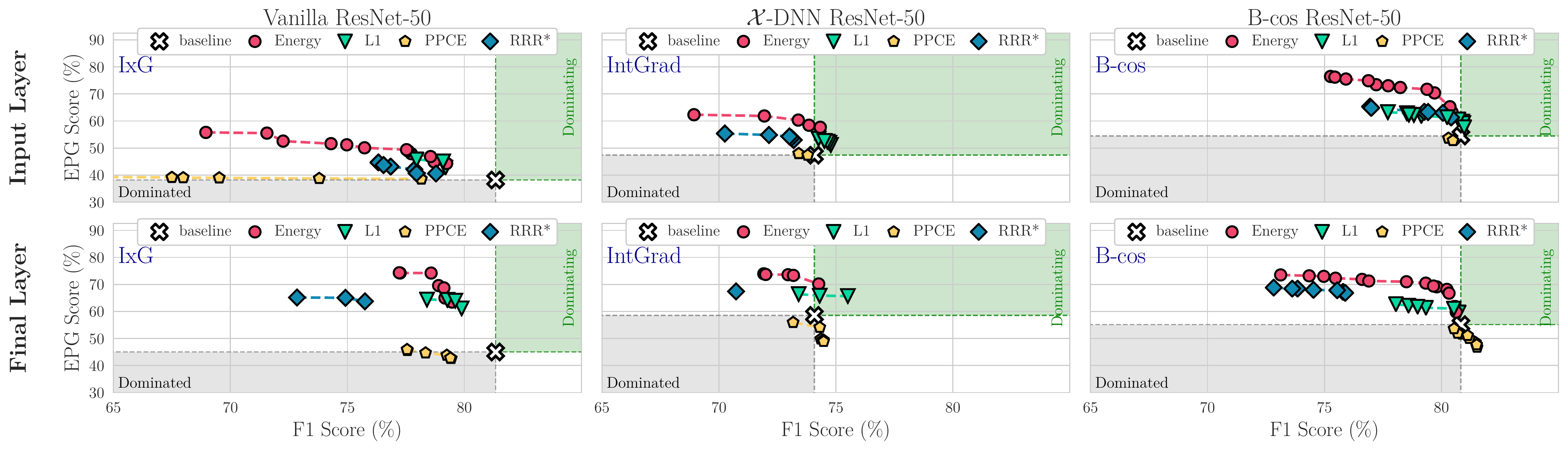}
    \caption{\textbf{PASCAL VOC results for \energypg vs.~\fone.} 
    }
    \label{fig:epg_voc}
    \end{subfigure}
    \begin{subfigure}[c]{.95\textwidth}
    \includegraphics[width=\textwidth]{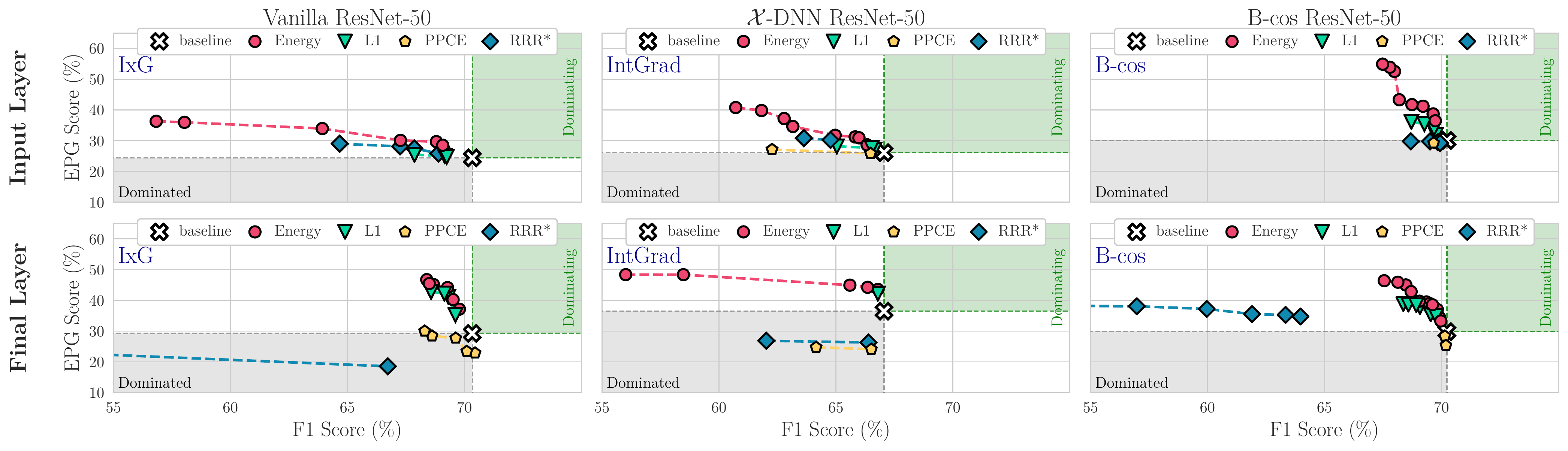}
    \caption{\textbf{MS COCO results for \energypg vs.~\fone.}
    }
    \label{fig:epg_coco}
    \end{subfigure}
    \caption{\textbf{EPG vs.~\fone,} for different datasets (\textbf{(a)}: VOC; \textbf{(b)}: COCO), losses (\textbf{markers}) and models (\textbf{columns}), optimized at different layers (\textbf{rows}); additionally, we show the performance of the baseline model before fine-tuning and demarcate regions that strictly dominate (are strictly dominated by) the baseline performance in green (grey). 
    For each configuration, we show the Pareto fronts (cf.\ \cref{fig:pareto_example}) across regularization strengths $\lambda_\text{loc}$ and epochs (cf.\ \cref{sec:results} and \cref{fig:pareto_example}). 
    We find the \epgloss loss to give the best trade-off between \epg and \fone.
    }
    \label{fig:epg_results}
        \vspace{-.75em}
\end{figure*}

\begin{figure*}[t]
    \centering 

    \begin{subfigure}[c]{.95\textwidth}
    {\includegraphics[width=\textwidth]{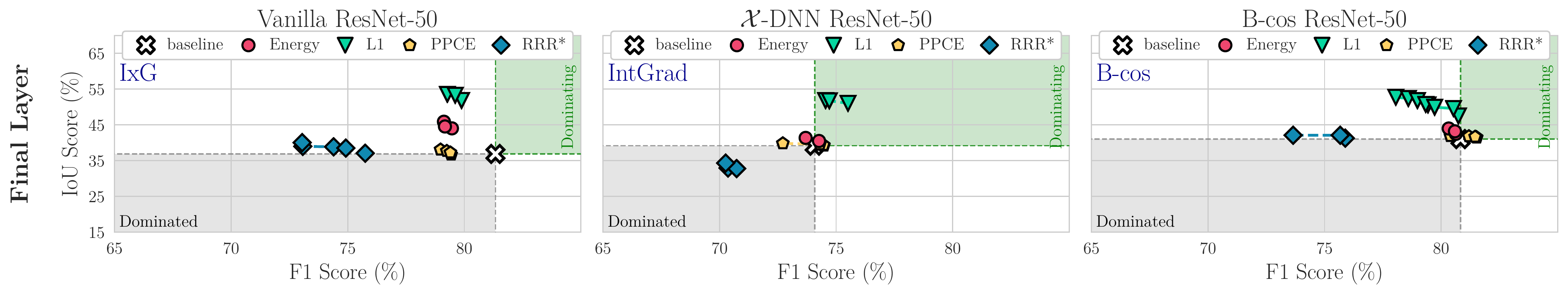}}
    \end{subfigure}
    \caption{\textbf{\iou vs.~\fone,} for different 
    losses (\textbf{markers}) and models (\textbf{columns}) for \vocs; results for \cocos are in the supplement. Additionally, we show the performance of the baseline model before fine-tuning and demarcate regions that strictly dominate (are strictly dominated by) the baseline model in green (grey). 
    For each configuration, we show the Pareto fronts (\cref{fig:pareto_example}) across regularization strengths $\lambda_\text{loc}$ and all epochs; for details, see \cref{sec:experiments,,sec:results}. 
    Across all configurations, we find the \lone loss to provide the largest gains in \iou at the lowest cost.
    }
    \label{fig:iou_results}
        \vspace{-1em}
\end{figure*}

\begin{figure}[t]
    \vspace{-.1em}
    \centering
    \includegraphics[width=\linewidth]{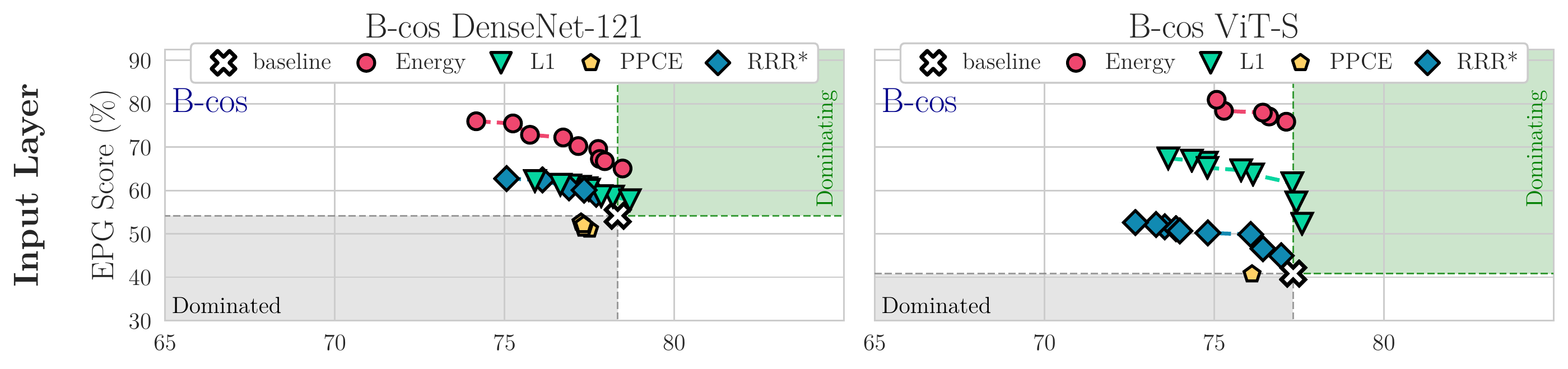}
    \caption{\textbf{\epg vs.~\fone on VOC.} We observe the same trends as in \cref{fig:epg_voc} for different backbone architectures, specifically a B-cos DenseNet-121 and a B-cos ViT-S. For \iou results, see supplement.}
    \label{fig:moremodels}
\end{figure}

\begin{figure*}[t]
    \vspace{-.1em}
    \centering
    \begin{subfigure}[c]{.95\textwidth}
    \includegraphics[width=\linewidth]{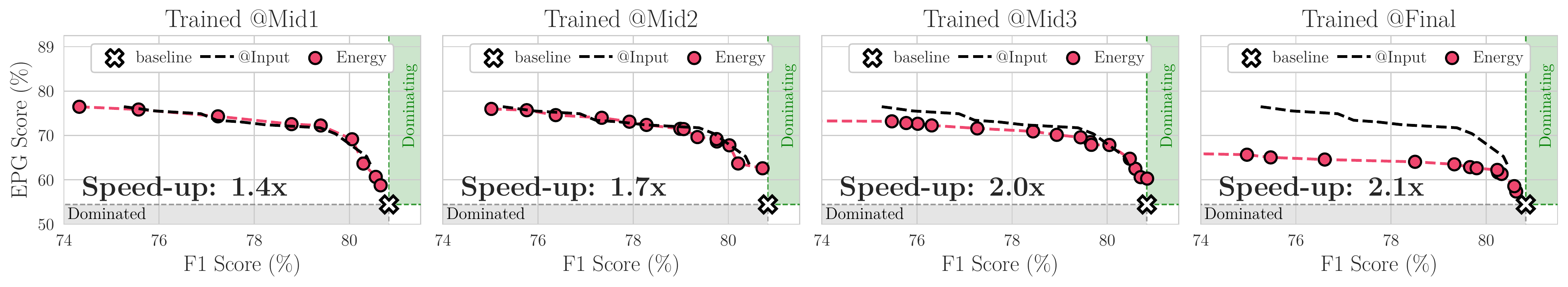}
    \end{subfigure}
    \caption{
    \textbf{Faster training by guiding at later layers.} While input-level attributions tend to be more detailed (cf.~\cref{fig:teaser2}), they are costlier to compute than attributions at later layers. However, we find that guidance at later layers (\eg @Mid3) also significantly improves input-level attributions, yielding similar \epg results as input-level guidance (@Input) at up to twice the training speed; for \iou results, see supplement.
    }
    \label{fig:speedup}
\end{figure*}
In this section, we discuss our experimental findings. In particular, in \cref{sec:results:epg+iou}, we first discuss the impact of the loss functions on the \epg and \iou scores of the models; in \cref{sec:results:layers+models}, we then analyze the impact of the models and attribution methods; further in \cref{sec:results:accuracy}, we show that 
guiding the models via their explanations can lead to improved classification accuracy. In \cref{sec:results:ablations}, we present additional studies in which we evaluate and discuss the cost of model guidance approaches: in particular, we study model guidance with limited additional labels, with increasingly coarse bounding boxes, and at deep layers in the network.
Finally, in \cref{sec:results:waterbirds}, we show the utility of model guidance in improving accuracy in the presence of distribution shifts.
For easier reference, we label our individual findings as \finding1--\finding9.

\myparagraph{Note.} To draw conclusive insights and highlight general and reliable trends in the experiments, we compare the Pareto curves (see \cref{fig:pareto_example}) of individual configurations. If the Pareto curve of a specific loss (\eg \epgloss in \cref{fig:epg_results}) consistently Pareto-dominates the Pareto curves of all other losses, we can confidently conclude that for the combination of evaluated metrics (\eg \epg vs.~\fone), this loss is the best choice.

\subsection{Comparing loss functions for model guidance}
\label{sec:results:epg+iou}

\begin{figure}[t]
    \centering 
    \includegraphics[width=\columnwidth]{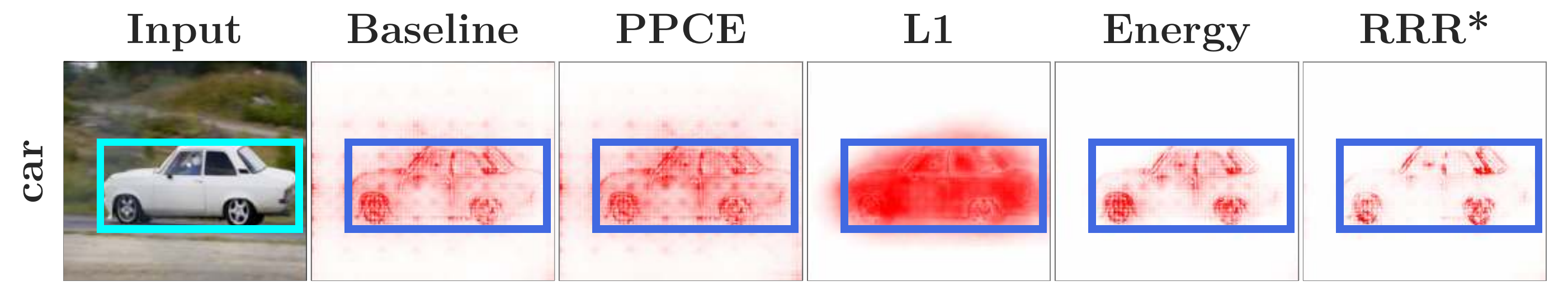}
    \caption{\textbf{Loss comparison} for input attributions (atts.) of a \bcos model. We show atts.~before (baseline, col.~2) and after guidance (cols.~3-6) for a specific image (col.~1) and its bounding box annotation. We find that \epgloss and \rrrloss yield sparse atts, whereas \lone yields smoother atts, as it is optimized to fill the entire bounding box. For \ppceloss we observe only a minor effect on the atts.
    }
    \label{fig:loss_comp}
\end{figure}

In the following, we highlight the main insights gained from the \emph{quantitative} evaluations. For a \emph{qualitative} comparison between the losses, please see \cref{fig:loss_comp}; note that we show examples for a \bcos model as the differences become clearest; full results can be found in the supplement.

\myparagraph{\finding1 The \energyloss loss yields the best \epg scores.} 
In \cref{fig:epg_results}, we plot the Pareto curves for \epg vs.~\fone scores for a wide range of configurations (see \cref{sec:experiments}) on \vocs (a) and \cocos (b); specifically, we group the results by model type (\vanilla, \xdnn, \bcos), the layer depths at which the attribution was regularized (Input / Final), and the loss used during optimization (\energyloss, \loneloss, \ppceloss, \rrrloss). 
From these results it becomes apparent that the optimization with the \energyloss loss yields the best trade-off between accuracy (\fone) and the \epg score: \eg, when looking at the upper right plot in \cref{fig:epg_voc} we can see that the \epgloss loss (red dots) improves over the baseline \bcos model (white cross) by improving the localization in terms of \epg score with only a minor cost in classification performance (\ie \fone score). Further trading off F1 scores yields even higher \epg scores. Importantly, the \epgloss loss Pareto-dominates all the other losses (\rrrloss: blue diamonds; \lone: green triangles; \ppceloss: yellow pentagons). This is is also true for the other network types (\vanilla \resnet, \cref{fig:epg_voc} (top left), and \xdnn, \cref{fig:epg_voc} (top center)) and at the final layer (bottom row), and generalizes across backbone architectures (\cref{fig:moremodels}). When comparing \cref{fig:epg_voc} and \cref{fig:epg_coco}, we also find these results to be highly consistent between datasets.

\myparagraph{\finding2 The \lone loss yields the best \iou performance.} Similarly, in \cref{fig:iou_results}, we plot the Pareto curves of \iou vs.~\fone scores for various configurations at the final layer; for the \iou results at the input layer and on the \cocos dataset, please see the supplement. 
For \iou, the \lone loss provides the best trade-off and, with few exceptions, \lone-guided models Pareto-dominate all other models in all configurations.

\myparagraph{\finding3 The \epgloss loss focuses best on on-object features.}
By not forcing the models to highlight the entire bounding boxes (see \cref{sec:method:energyloss}),
we find that the \energyloss loss also suppresses background features \emph{within} the bounding boxes, thus better preserving fine details of the explanations (cf.~\cref{fig:loss_comp,,fig:dilation_comp}). To quantify this, we evaluate the distribution of \energyloss (\cref{eq:epg}) just within the bounding boxes. For this, 
we take advantage of the segmentation mask annotations available for a subset of the \vocs test set. Specifically, we measure the \energyloss contained in the segmentation masks versus the entire bounding box, which indicates how much of the attributions actually highlight on-object features. We find that the \energyloss loss outperforms \lone across all models and configurations; see supplement for details.

In short, we find that the \epgloss loss works best for improving the \epg metric, whereas the \lone loss yields the highest gains in terms of \iou; depending on the use case, either of these losses could thus be recommendable. However, we find that the \epgloss loss is more robust to annotation errors (\finding8, \cref{sec:results:ablations}), and, as discussed in \finding3, the \epgloss loss more reliably focuses on object-specific features.

\subsection{Comparing models and attribution methods}
\label{sec:results:layers+models}
In the following, we highlight our findings regarding different attribution methods and models. Given the similarity of the results between \gradcam and \ixg, and since \bcos attributions performed better than \gradcam for \bcos models, we show \gradcam results in the supplement.

\myparagraph{\finding4 At the input layer, \bcos explanations perform best.} We find that the \bcos models not only achieve the highest \epg/\iou performance before applying model guidance,
(`baselines') 
but also obtain the highest gains in \epg and \iou and thus the highest overall performance (for \epg see \cref{fig:epg_results}, right; for \iou, see supplement): \eg, an \energyloss-based \bcos model achieves an \epg score of 71.7 @ 79.4\% \fone, thus significantly outperforming the best \epg scores of both other model types at a much lower cost in \fone (\vanilla: 55.8 @ 69.0\%, \xdnn: 62.3 @ 68.9\%). This is also observed \emph{qualitatively}, as we show in the supplement.

\myparagraph{\finding5 Regularizing at the final layer yields consistent gains.} As can be seen in \cref{fig:epg_results} (bottom) and \cref{fig:iou_results}, all models can be guided well via regularization at the final layer, \ie all models show improvements in \iou and \epg score.

In short, we find model guidance to work well across all tested models when optimizing at the final layer (\finding5), highlighting its wide applicability. However, to obtain highly detailed and well-localized attributions at the input layer, the model-inherent explanations of the \bcos models seem to lend themselves much better to such guidance (\finding4).

\subsection{Improving accuracy with model guidance}
\label{sec:results:accuracy}

\myparagraph{\finding6 Model guidance {can} improve accuracy.} For both the \vanilla models (final layer) and the \xdnns (input+final), we found models that improve the localization metrics \emph{and} the \fone score. These improvements are particularly pronounced for the \xdnn: \eg, we find models that improve the \epg and \fone scores by $\Delta\myeq7.2$ p.p.\  and $\Delta\myeq1.4$ p.p.\ respectively (\cref{fig:epg_results}, center top), or the \iou and \fone scores by $\Delta\myeq11.9$ p.p.\ and $\Delta\myeq1.4$ p.p.\ (\cref{fig:iou_results}, center).

However, overall we observe a trade-off between localization and accuracy (\cref{fig:epg_results,fig:iou_results}). Given the similarity of the training and test distributions, focusing on the object need not improve classification performance, as spurious features are also present at test time. Further, the guided model is discouraged from relying on contextual features, making the classification more challenging. In \cref{sec:results:waterbirds}, we show that guidance can significantly improve performance when there is a distribution shift between training and test.

\subsection{Efficiency and robustness considerations}
\label{sec:results:ablations}

While bounding boxes decrease the data collection cost with respect to segmentation masks, they can nonetheless be expensive to obtain, especially when expert knowledge is required. To further reduce those costs, in this section, we assess the robustness of guiding the model with a limited  number (\finding7) or increasingly coarse annotations (\finding8).
Apart from \emph{data efficiency}, we further explore how \emph{training efficiency} can be improved for fine-grained (\ie input-level) explanations (\finding9), as explanations at early layers are more costly to obtain than those at later layers.

\begin{figure}[ht]
    \centering 
    \hspace{-0.05\columnwidth}
    \includegraphics[width=1.025\columnwidth]{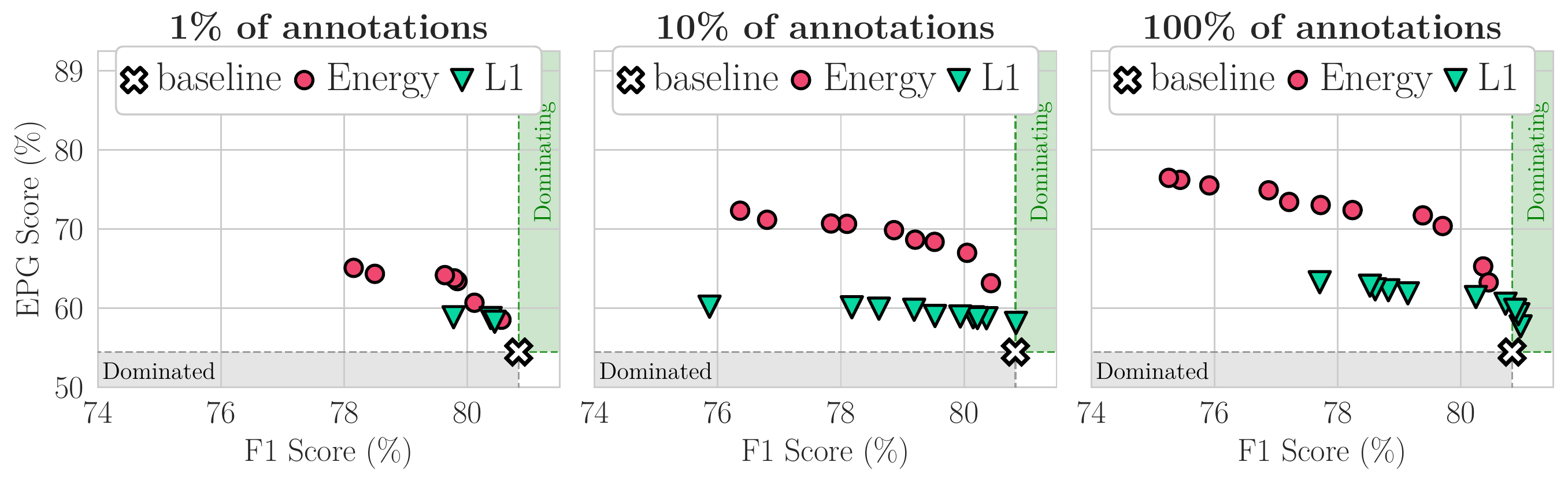}
    \caption{\textbf{\epg results with limited annotations} for a \bcos model at the input layer, optimized with the \epgloss and the \lone loss. Using bounding box annotations for as little as 1\% (left) of the images yields significant improvements in \epg, and with 10\% (center) similar gains as in the fully annotated setting (right) are obtained.
    }
    \label{fig:lim_local_input}
    \vspace{-.05em}
\end{figure}

\myparagraph{\finding7 Model guidance requires only few add.~annotations.} In \cref{fig:lim_local_input}, we show that the \epg score can be significantly improved with a very limited number of annotations; for \iou results, see supplement. Specifically, we find that when using only 1\% of the training data (25 annotated images) for \vocs, improvements of up to $\Delta\myeq23.0$ p.p.\ ($\Delta\myeq1.4$) in \epg (\iou) can be obtained, at a minor drop in \fone ($\Delta\myeq0.3$ p.p.\ and $\Delta\myeq2.5$ p.p.\ respectively). When annotating up to 10\% of the images, very similar results can be achieved as with full annotation (see \eg cols.~2+3 in \cref{fig:lim_local_input}).

\begin{figure}[ht]
    \centering 
    \hspace{-.04\columnwidth}\includegraphics[width=1.025\columnwidth]{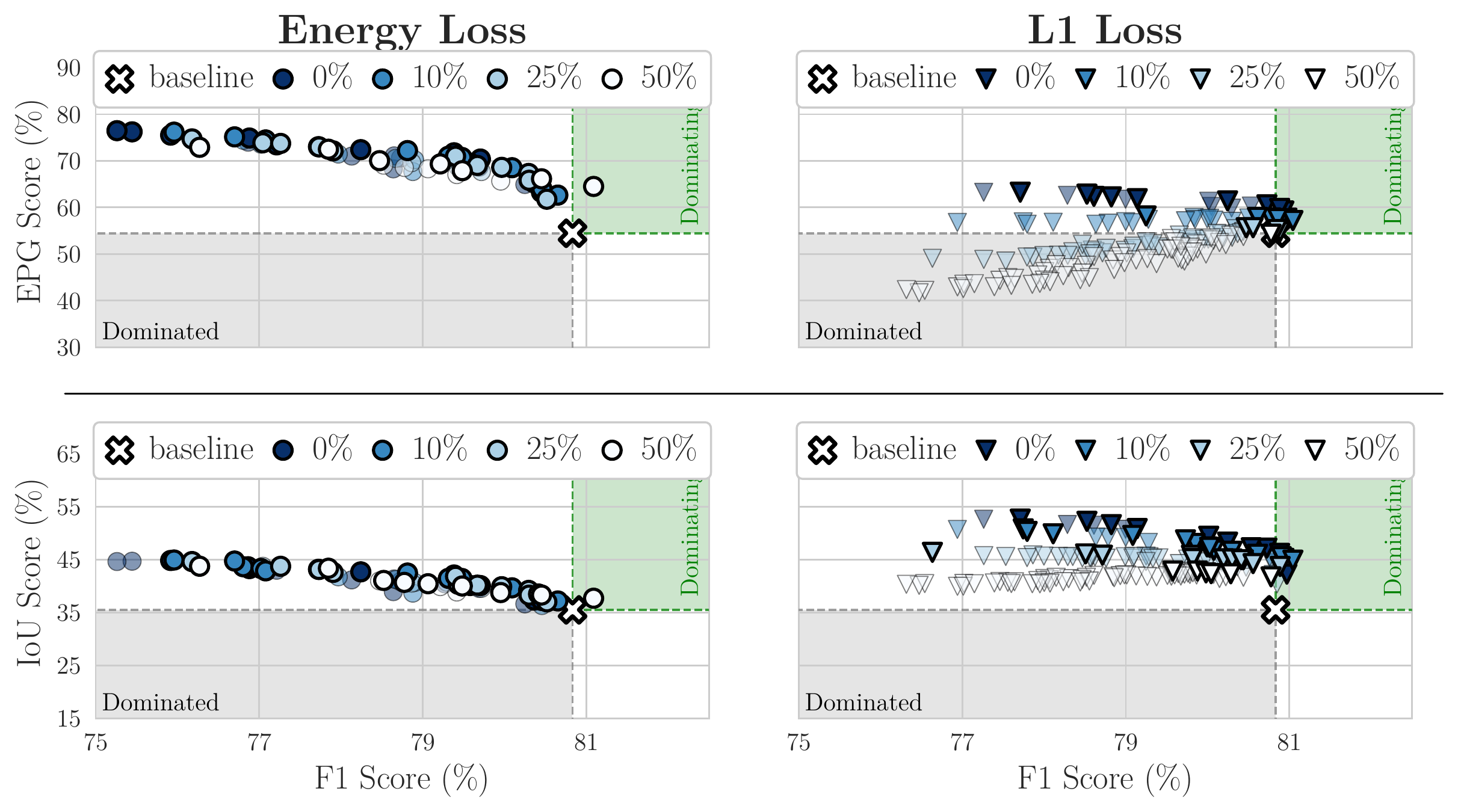}
    \caption{\textbf{Quantitative results for dilated bounding boxes} for a \bcos model at the input layer.
    We show \epg  and \iou (top and bottom) results for models trained with various amounts of annotation errors (increasingly large bounding boxes, see \cref{fig:dilation_comp}). The \epgloss loss yields highly consistent results despite training with heavily dilated bounding boxes (left), whereas the results of the \lone loss (right) worsen markedly; best viewed in color.
    }
    \label{fig:coarse_annotations}
\end{figure}

\begin{figure}[ht]
    \centering 
    \includegraphics[width=\columnwidth]{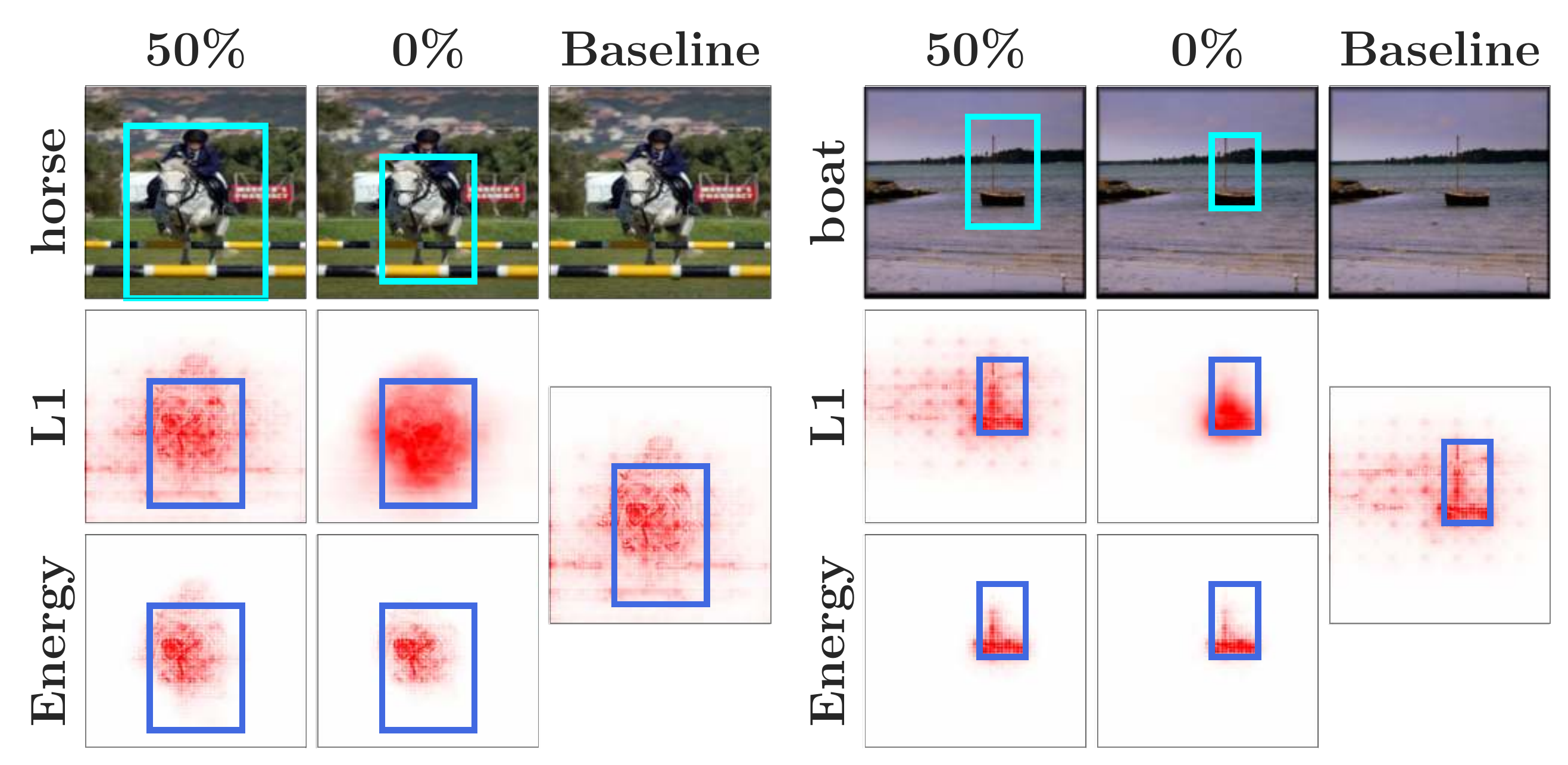}
    \caption{\textbf{Qualitative results for dilated bounding boxes} for a \bcos model at input.
    Examples for attributions (rows 2+3) of models trained with dilated bounding boxes (row 1). In contrast to \lone, models trained with \epgloss show significant gains in object focus even with significant noise (\eg `Baseline' vs.\ `50\%').
    }
    \label{fig:dilation_comp}
\end{figure}

\myparagraph{\finding8 The \energyloss loss is highly robust to annotation errors.} As discussed in \cref{sec:method:energyloss}, the \energyloss loss only directs the model on which features \emph{not} to use and does not impose a uniform prior on the attributions within the bounding boxes. As a result, we find it to be much more stable to annotation errors: \eg, in \cref{fig:coarse_annotations}, we visualize how the \epg (top) and \iou (bottom) scores of the best performing models under the \epgloss (left) and \lone loss (right) evolve when using coarser bounding boxes; for this, we simply dilate the bounding box size by $p\myin\{10, 25, 50\}$\% during training, see \cref{fig:dilation_comp}. While the models optimized via the \lone loss achieve increasingly worse results (right),  the \epgloss-optimized models are essentially unaffected by the coarseness of the annotations.

In short, we find that the models can be guided effectively at a low cost in terms of annotation effort, as only few annotations (\eg 25 for \vocs) are required (cf.\ \finding7), and, especially for the \epgloss loss, these annotations can be very coarse and do not have to be `pixel-perfect' (cf.\ \finding8).

\myparagraph{\finding9 Guidance at deep layers can be effective.}
While guided input-level explanations of \bcos networks exhibit a high degree of detail, 
regularizing those explanations 
comes at an added training cost. 
In particular, optimizing at the input layer requires backpropagating through the entire network to compute the attributions. In an effort to reduce training costs whilst
maintaining the benefits of fine-grained explanations at input resolution,
we evaluate if input-level attributions benefit from an optimization at deeper layers.

Specifically, we regularize \bcos attributions at the final and at three intermediate layers (Mid\{1,2,3\}), and evaluate the localization of attributions at the input. 
We find (\cref{fig:speedup}) that training at a deeper layer can provide significant speed-ups in training time with often a negligible cost in localization performance. \Eg, since we do not have to compute a full backward pass through the entire model during training, optimizing at Mid2 (col.~2 in \cref{fig:speedup}) provides similar gains in localization but with a 1.7x speed-up in training time.

\subsection{Effectiveness against spurious correlations}
\label{sec:results:waterbirds}
To evaluate the potential for mitigating spurious correlations, we evaluate model guidance with the \epgloss and \lone losses on the synthetically constructed \waterbirds dataset \citeMain{sagawa2019distributionally,petryk2022guiding}. We perform model guidance under two settings: (1) the conventional setting to classify between landbirds and waterbirds, using the region within the bounding box as the mask; and (2) the reversed setting \cite{petryk2022guiding} to classify the background, \ie, land vs.\ water, using the region outside the bounding box as the mask. To simulate a limited annotation budget, we only use bounding boxes for a random 1\% of the training set, and report results averaged over four runs. We show the results for the worst-group accuracy (\ie, images containing a waterbird on land) and the overall accuracy using \bcos models in \cref{tab:waterbirds}; full results for all attributions and models can be found in the supplement.

\begin{figure}[t]
    \centering
    \includegraphics[width=\linewidth]{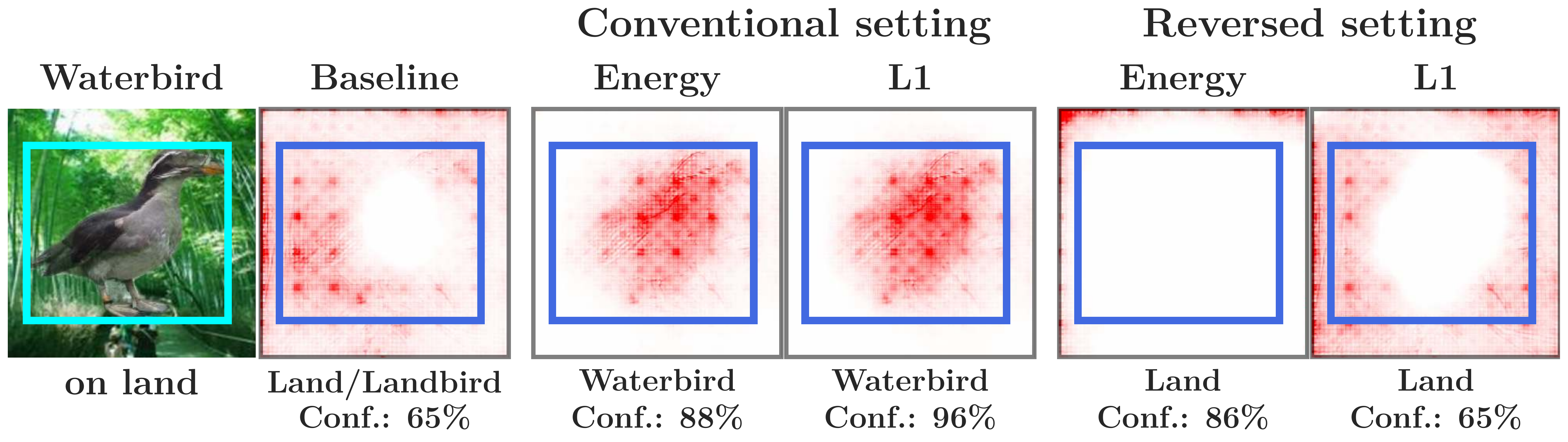}
    \caption{\textbf{Qualitative \waterbirds results.} Without guidance, a model might focus on the background to classify birds (baseline) and thus misclassify waterbirds on land (col.~2). Guided models can correct such errors and focus on the desired feature: in cols.~3+4 (5+6) the model is guided to classify by using the bird (background) features and arrives at the desired prediction. Model predictions and confidence scores are indicated below the images.
    }
    \label{fig:waterbirds}
\end{figure}

Both losses consistently and significantly improve the accuracy in the conventional and the reversed settings by guiding the model to select the `right' features, \ie birds (conventional) or background (reversed). This guidance can also be observed qualitatively (cf.~\cref{fig:waterbirds}).

{
\setlength\tabcolsep{.5em}
\begin{table}[t]
    \centering
    \begin{tabular}{c|cccc}
	    & \multicolumn{2}{c}{{\footnotesize Conventional}} & \multicolumn{2}{c}{{\footnotesize Reversed}} \\
	    \hline
        {\footnotesize Model} & {\footnotesize Worst} & {\footnotesize Overall} & {\footnotesize Worst} &  {\footnotesize Overall} \\
        {\footnotesize Baseline }
        &\footnotesize{ 43.4 \scriptsize($\pm$2.4)  }&\footnotesize{ 68.7 \scriptsize($\pm$0.2)}
        &\footnotesize{ 56.6 \scriptsize($\pm$2.4)  }&\footnotesize{ 80.1 \scriptsize($\pm$0.2)}
        \\
        {\footnotesize Energy }
        &\footnotesize{ \textbf{56.1 \scriptsize($\pm$4.0)} }&\footnotesize{ \textbf{71.2 \scriptsize($\pm$0.1)}}
        &\footnotesize{ \textbf{62.8 \scriptsize($\pm$2.1)} }&\footnotesize{ \textbf{83.6 \scriptsize($\pm$1.1)}}
        \\
        {\footnotesize $L_1$ }
        &\footnotesize{ 51.1 \scriptsize($\pm$1.9) }&\footnotesize{ 69.5 \scriptsize($\pm$0.2)}
        &\footnotesize{ 58.8 \scriptsize($\pm$5.0) }&\footnotesize{ 82.2 \scriptsize($\pm$0.9)}
        \\
	\end{tabular}
	\caption{\textbf{Waterbirds-100 results.} We find that model guidance is effective in improving both worst-group (`Waterbird on Land')
 and overall accuracy in the conventional (Landbird vs.\ Waterbird) and reversed (Land vs.\ Water) settings; full results in the supplement.}
        \label{tab:waterbirds}
\end{table}
}

\section{Discussion And Conclusion}
\label{sec:conclusion}
In this work, we comprehensively evaluated various models, attribution methods, and loss functions for their utility in guiding models to be ``right for the right reasons''.

In summary, we find that
guiding models via bounding boxes can significantly improve \epg and \iou performance of the optimized attribution method, with the \epgloss loss working best to improve the \epg score (\finding1) and the \lone loss yielding the highest gains in \iou scores (\finding2). While the \bcos models achieve the best results in \iou and \epg score at the input layer (\finding4), all tested model types (\vanilla, \xdnn, \bcos) lend themselves well to being optimized at the final layer (\finding5), which can even improve attribution maps at early layers (\finding9).
Further, we find that regularizing the explanations of the models and thereby `telling them where to look' can increase the object recognition performance (mAP/accuracy) of some models (\finding6), especially when strong spurious correlations are present (\cref{sec:results:waterbirds}).
Interestingly, those gains (\epg, \iou), can be achieved with relatively little additional annotation (\finding7).
Lastly, we find that by not assuming a uniform prior over the attributions within the annotated bounding boxes, training with the energy loss is more robust to annotation errors (\finding8) and results in models that produce attribution maps that are more focused on class-specific features (\finding3).

{\small
\bibliographystyle{ieee_fullname}
\bibliography{references}
}

\clearpage

\appendix

\renewcommand\thesection{\Alph{section}}
\numberwithin{equation}{section}
\numberwithin{figure}{section}
\numberwithin{table}{section}
\renewcommand{\thefigure}{\thesection\arabic{figure}}
\renewcommand{\thetable}{\thesection\arabic{table}}
\crefname{appendix}{Sec.}{Secs.}

{\onecolumn 
{\begin{center}
\Large\bf
\phantom{skip}\\[.25em]
{Studying How to Efficiently and Effectively Guide Models with Explanations}\\[1em]
\large
Appendix
\end{center}
}
\newcommand{\additem}[2]{%
\item[\textbf{(\ref{#1})}] 
    \textbf{#2} \dotfill\makebox{\textbf{\pageref{#1}}}
}

\newcommand{\addsubitem}[2]{%
    \textbf{(\ref{#1})}\hspace{1em}
    #2\\[.1em] 
}

\newcommand{\adddescription}[1]{\newline
\begin{adjustwidth}{.25cm}{.25cm}
#1
\end{adjustwidth}
}


{\vspace{2em}\bf\large Table of Contents\\[1em]}

In this supplement to our work on using explanations to guide models, we provide:
\\[1em]

\begin{adjustwidth}{1cm}{1cm}
\begin{enumerate}[label={({\arabic*})}, topsep=1em, itemsep=.25em]
    \additem{supp:sec:quali}{
    Additional qualitative results (VOC and COCO)
    }
    \adddescription{ 
    In this section, we present additional \emph{qualitative} results. In particular, we provide:
    
    \addsubitem{supp:sec:main:qualitative:examples}{Detailed comparisons between \textbf{models, layers, and losses}. (\vocs + \cocos).}
    \addsubitem{supp:sec:main:qualitative:dilation}{Additional visualizations for training with \textbf{dilated bounding boxes} (\vocs + \cocos).}
    }    
    \additem{supp:sec:quanti}{
    Additional quantitative results (VOC and COCO)
    }
    \adddescription{ 
    In this section, we present additional \emph{quantitative} results. In particular, we show:
    
    \addsubitem{supp:sec:quantitative:classvsloc}{The remaining \textbf{localization vs.\ accuracy comparisons} (\vocs + \cocos).}
    \addsubitem{supp:sec:quantitative:gradcam}{The results of guiding models via \textbf{\gradcam}. (\vocs + \cocos).}
    \addsubitem{supp:sec:quantitative:intermediate}{Results for optimizing at \textbf{intermediate layers} (\vocs).}
    \addsubitem{supp:sec:quantitative:segmentepg}{Results for measuring \textbf{on-object \epg scores} (\vocs).}
    \addsubitem{supp:sec:quantitative:limited}{Additional analyses regarding training with \textbf{few annotated images} (\vocs).}
    \addsubitem{supp:sec:quantitative:dilation}{Additional analyses regarding the usage of \textbf{coarse bounding boxes} (\vocs).}
    \addsubitem{supp:sec:quantitative:morearchs}{Additional results using other \textbf{model backbones}.}
    }
    \additem{supp:sec:waterbirds}{
    Additional results on the Waterbirds dataset
    }
    \adddescription{ 
    In this section, we provide additional results for the \waterbirds dataset. In particular, we provide full results regarding \textbf{classification performance} with and without model guidance as well as \textbf{additional qualitative visualizations} of the attribution maps.\\
    
    }
    \additem{supp:sec:implementation}{
    Implementation details
    }
    \adddescription{ In this section, we provide relevant implementation details; note that all code will be made available upon publication. In particular, we provide:
    
    \addsubitem{supp:sec:implementation:training}{Training details across datasets (\vocs + \cocos + Waterbirds).}
    \addsubitem{supp:sec:implementation:bcos}{Implementation details for twice-differentiable \bcos models.}
    }
    \additem{supp:sec:full}{
    Full results across all experiments.
    }
    \adddescription{ 
    Given the large amount of experimental results, in each of the preceding sections we show only a sub-selection of those results for improved readability. In section \ref{supp:sec:full}, we provide the \emph{full} results across datasets, models, layers, experiments, and metrics, to peruse at the reader's convenience.    
    ~\\
    }
\end{enumerate}
\end{adjustwidth}
}

\setlength{\parskip}{.5em}
\clearpage
\section{Additional Qualitative Results (\vocs and \cocos)}
\label{supp:sec:quali}

\subsection{Qualitative Examples Across Losses, Attribution Methods, and Layers}
\label{supp:sec:main:qualitative:examples}

\begin{figure}
    \centering
    \begin{subfigure}[c]{\textwidth}
    \centering
    \textbf{\large \voc.}\\\vspace{.25cm}
    \begin{subfigure}[c]{.475\columnwidth}
    \centering
    \textbf{Input}\\\vspace{.25cm}
    \includegraphics[width=\textwidth]{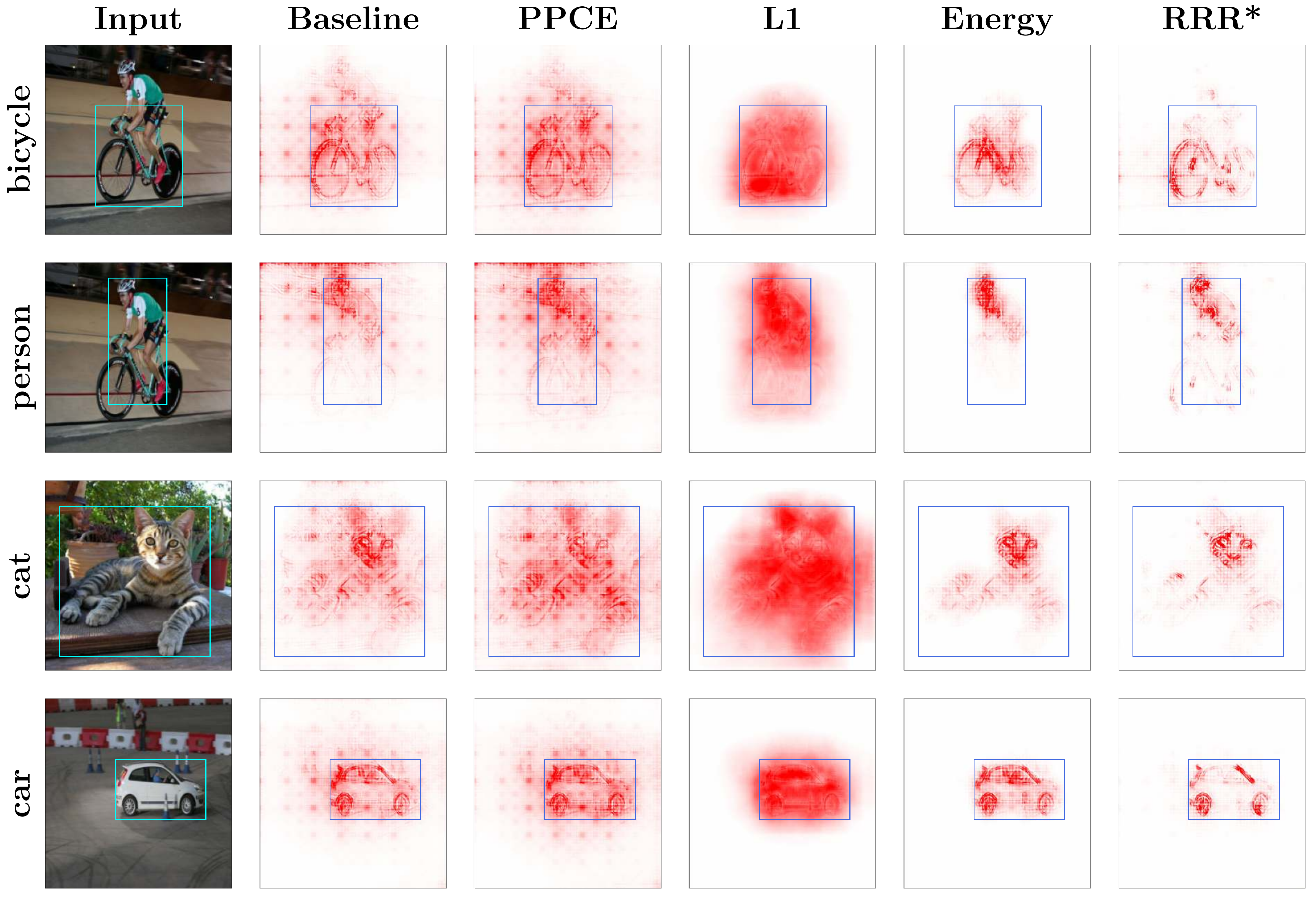}
    \end{subfigure}
    \begin{subfigure}[c]{.475\columnwidth}
    \centering
    \textbf{Final}\\\vspace{.25cm}
    \includegraphics[width=\textwidth]{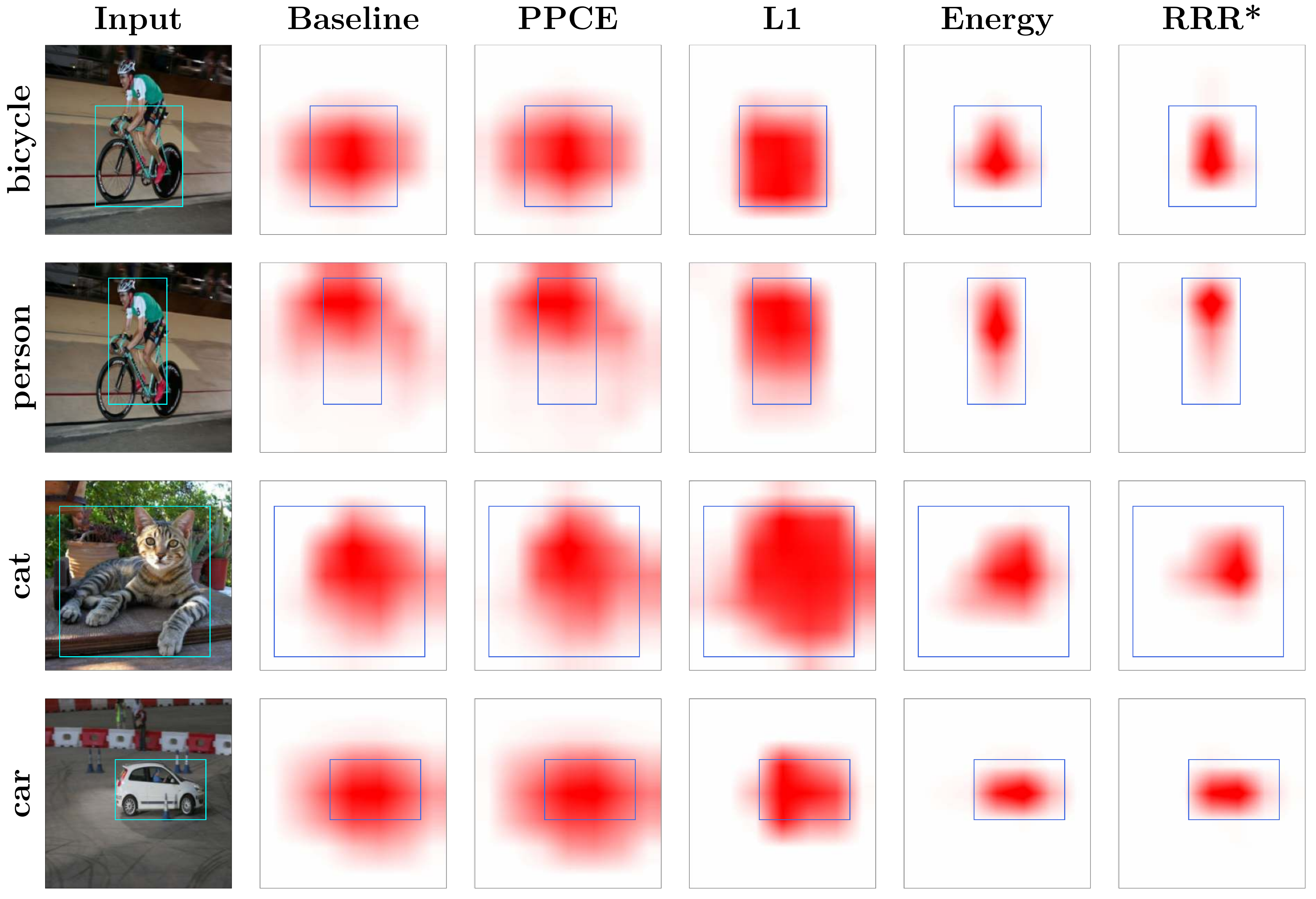}
    \end{subfigure}
    \caption{\textbf{\bcos \resnet}.}
    \label{fig:supp:quali_voc_1:bcos}
    \end{subfigure}
    \begin{subfigure}[c]{\textwidth}
    \centering
    \begin{subfigure}[c]{.475\columnwidth}
    \includegraphics[width=\textwidth]{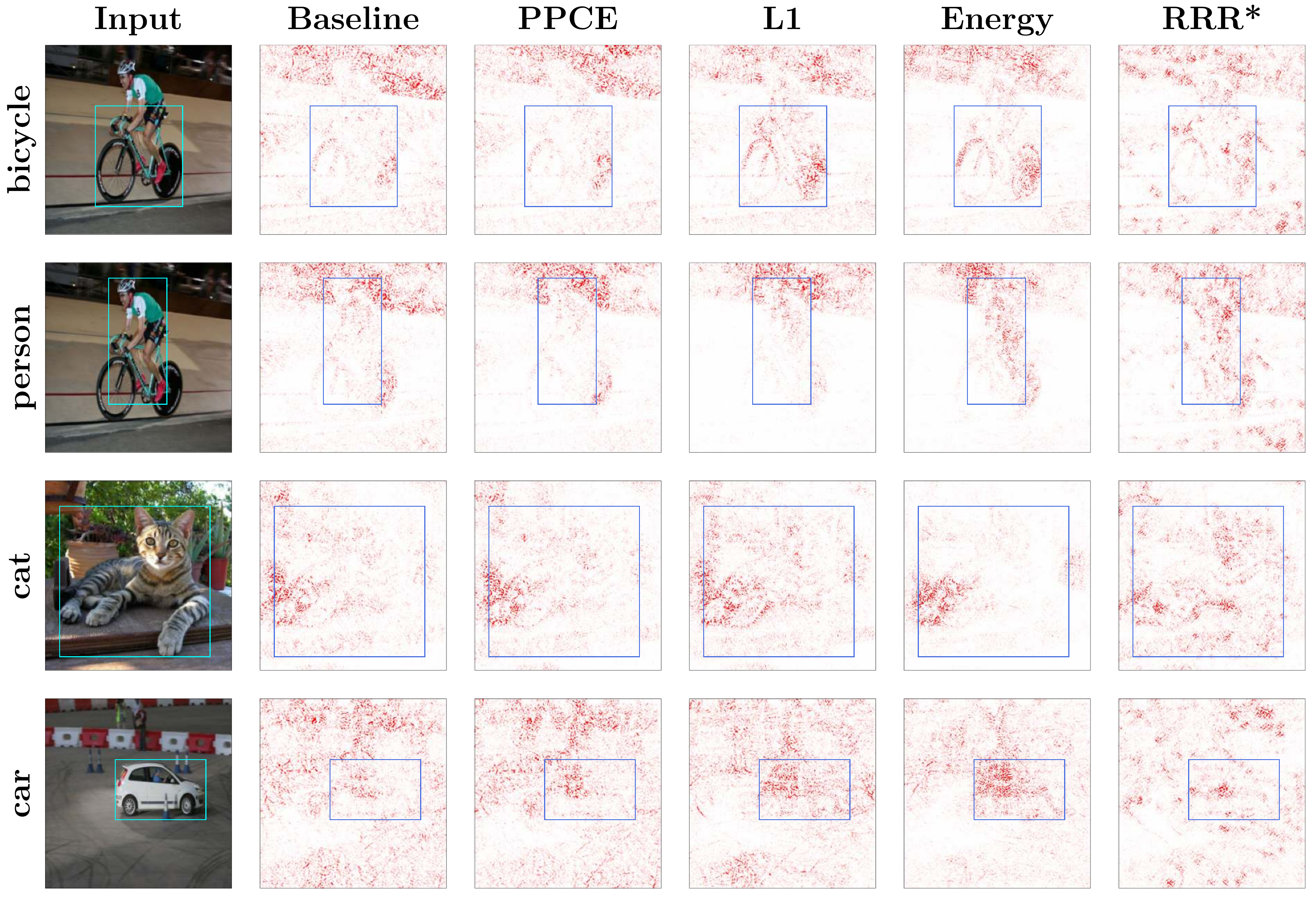}
    \end{subfigure}
    \begin{subfigure}[c]{.475\columnwidth}
    \includegraphics[width=\textwidth]{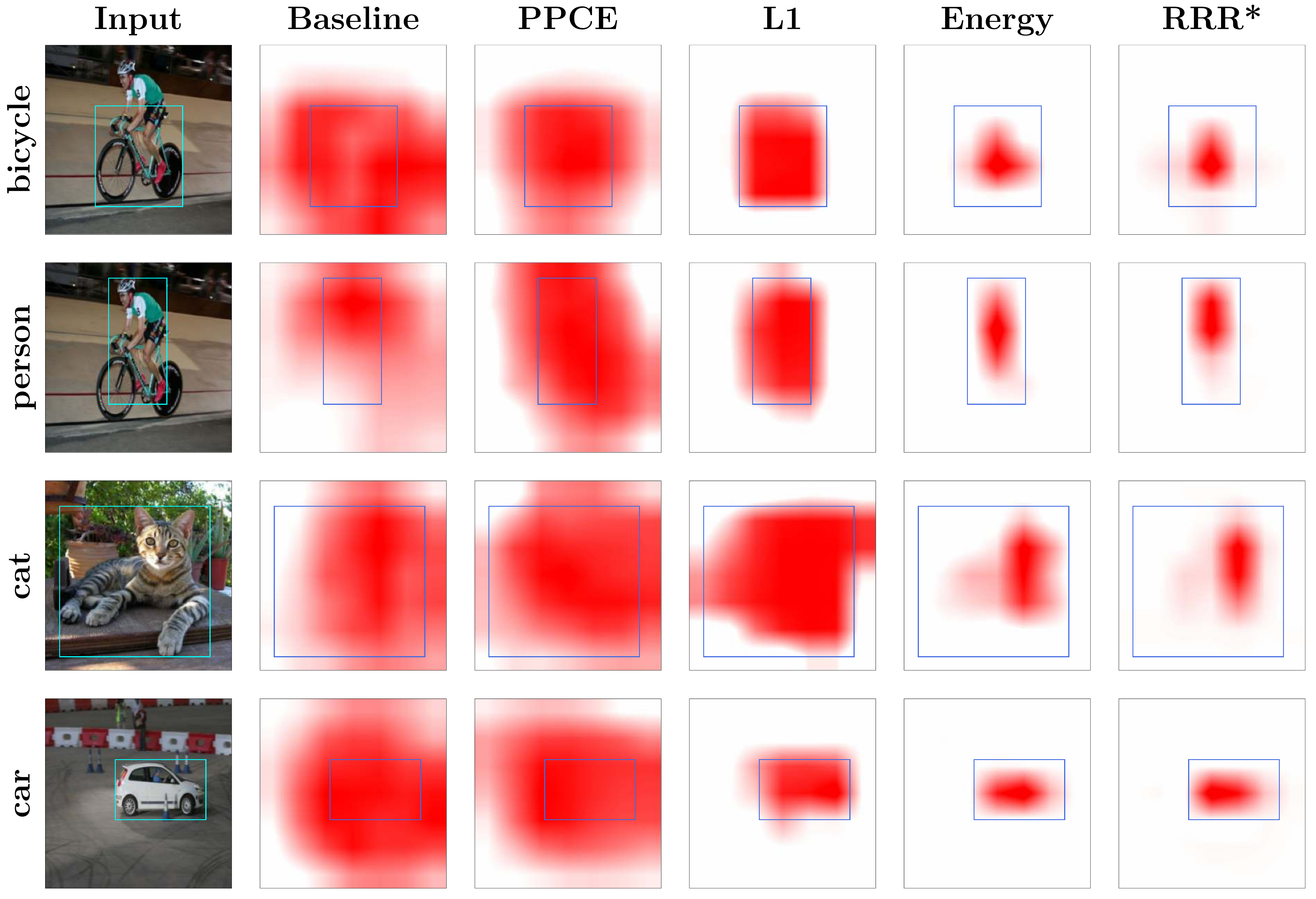}
    \end{subfigure}
    \caption{\textbf{\vanilla \resnet}.}
    \label{fig:supp:quali_voc_1:vanilla}
    \end{subfigure}
    \begin{subfigure}[c]{\textwidth}
    \centering
    \begin{subfigure}[c]{.475\columnwidth}
    \includegraphics[width=\textwidth]{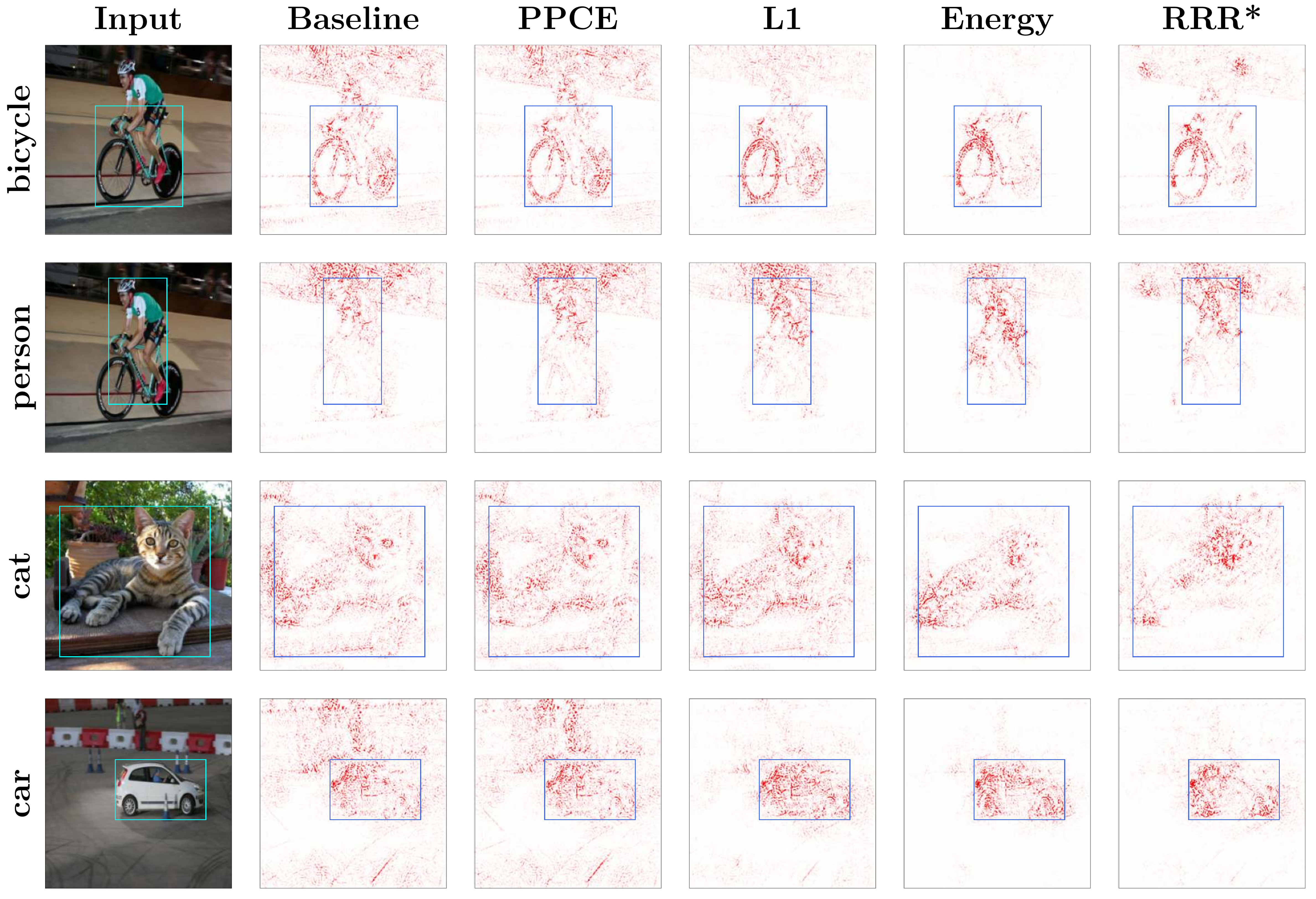}
    \end{subfigure}
    \begin{subfigure}[c]{.475\columnwidth}
    \includegraphics[width=\textwidth]{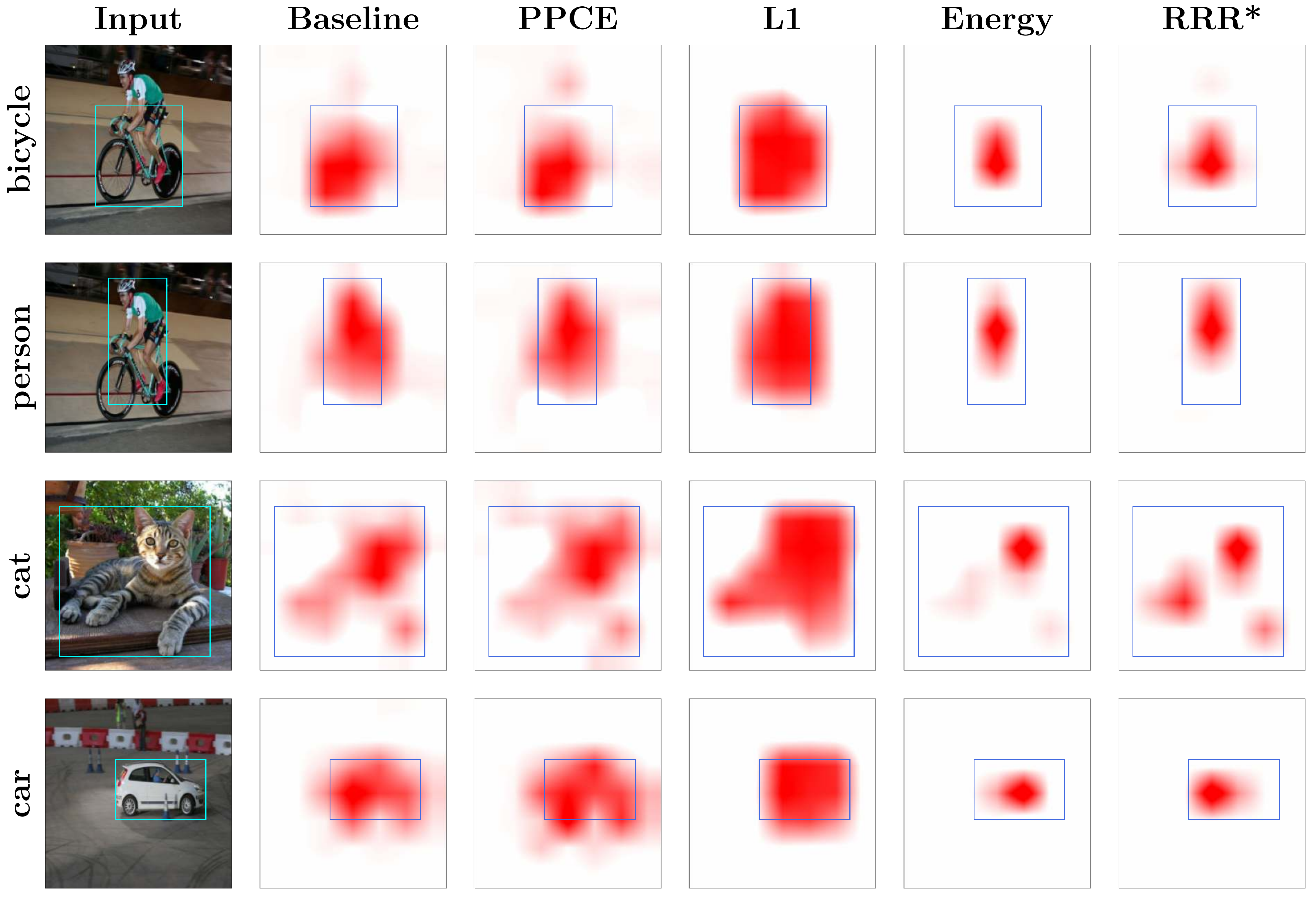}
    \end{subfigure}
    \caption{\textbf{\xdnn \resnet}.}
    \label{fig:supp:quali_voc_1:xdnn}
    \end{subfigure}
    \caption{Qualitative examples from the \textbf{\vocs dataset}. In particular, this figure allows to compare between models (\textbf{major rows}, \ie (a), (b), and (c)) losses (\textbf{major columns}) and layers (\textbf{left+right}) for multiple images (\textbf{minor rows}).}
    \label{fig:supp:quali_voc_1}
\end{figure}
\begin{figure}
    \centering
    \begin{subfigure}[c]{\textwidth}
    \centering
    \textbf{\large \coco.}\\\vspace{.25cm}
    \begin{subfigure}[c]{.475\columnwidth}
    \centering
    \textbf{Input}\\\vspace{.25cm}
    \includegraphics[width=\textwidth]{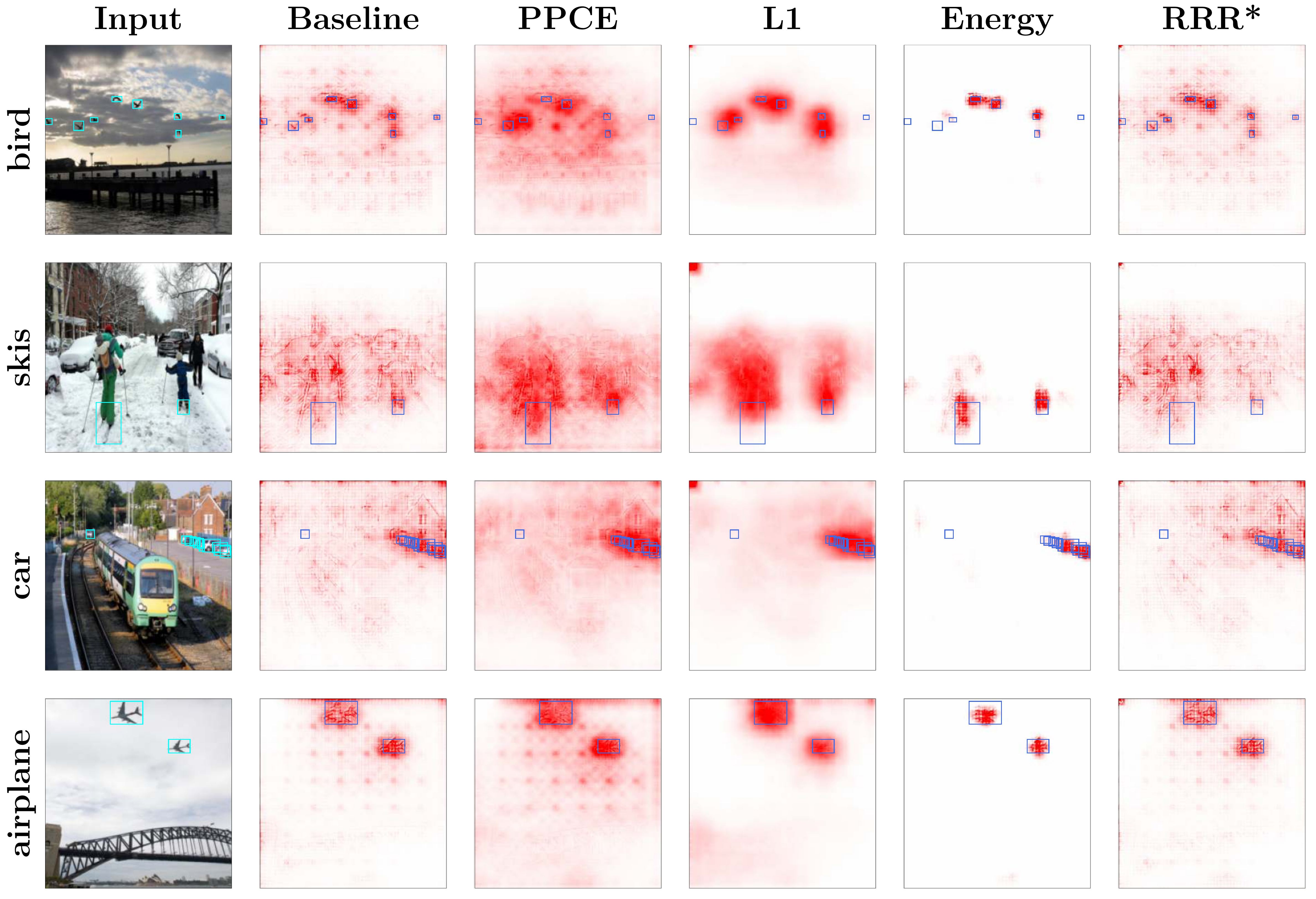}
    \end{subfigure}\hfill
    \begin{subfigure}[c]{.475\columnwidth}
    \centering
    \textbf{Final}\\\vspace{.25cm}
    \includegraphics[width=\textwidth]{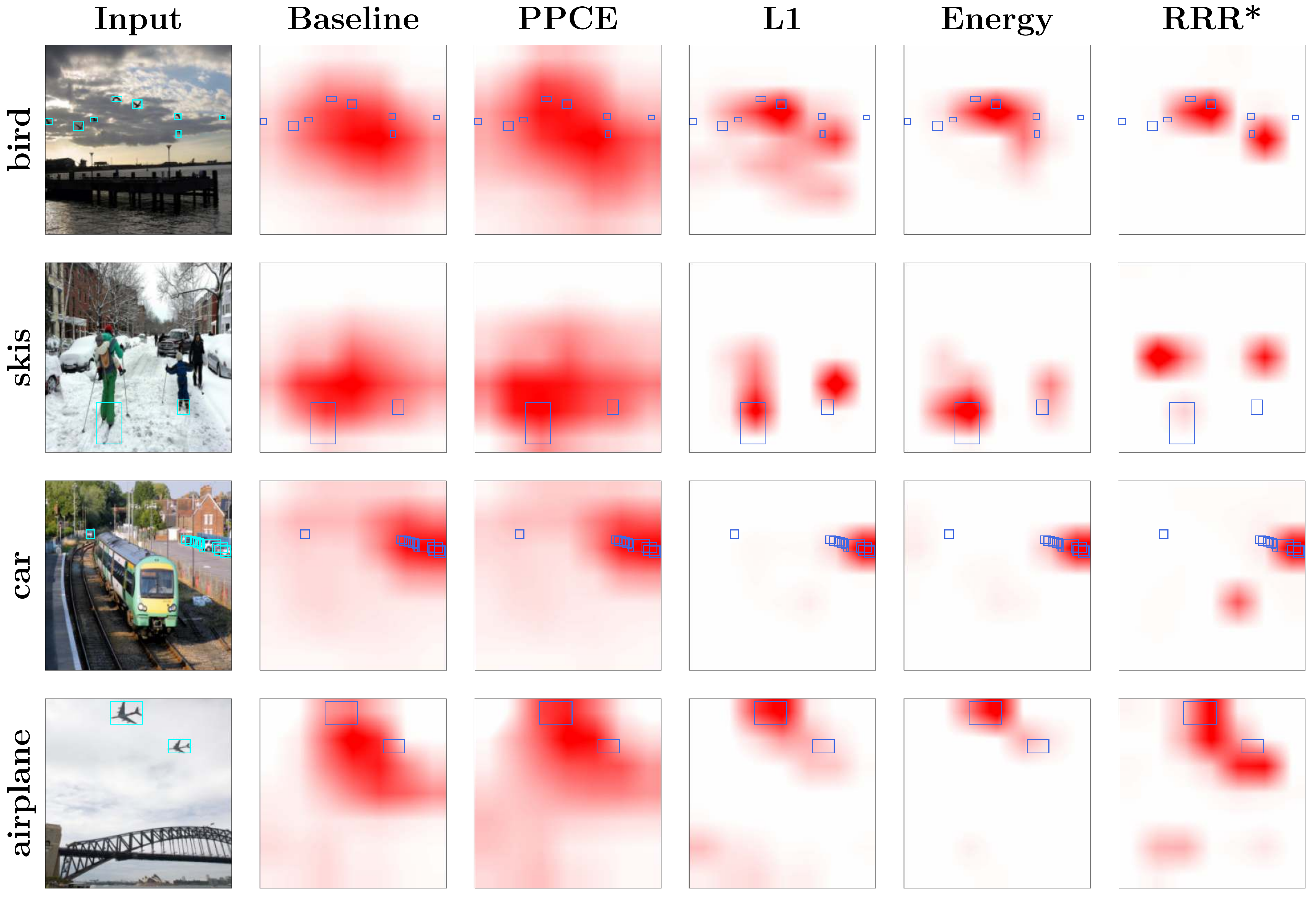}
    \end{subfigure}
    \caption{\textbf{\bcos \resnet}.}
    \label{fig:supp:quali_coco_1:bcos}
    \end{subfigure}
    \begin{subfigure}[c]{\textwidth}
    \centering
    \begin{subfigure}[c]{.475\columnwidth}
    \includegraphics[width=\textwidth]{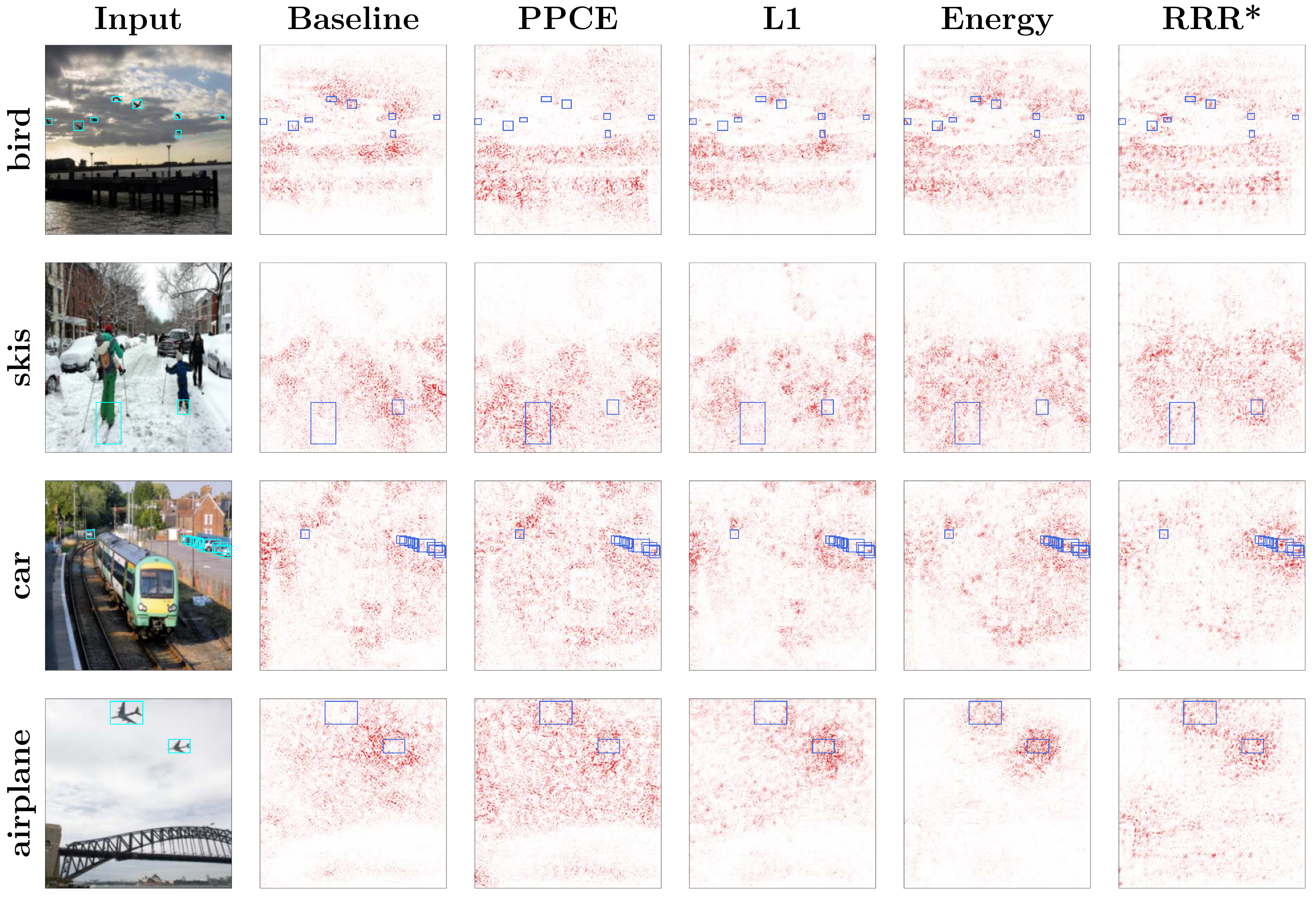}
    \end{subfigure}\hfill
    \begin{subfigure}[c]{.475\columnwidth}
    \includegraphics[width=\textwidth]{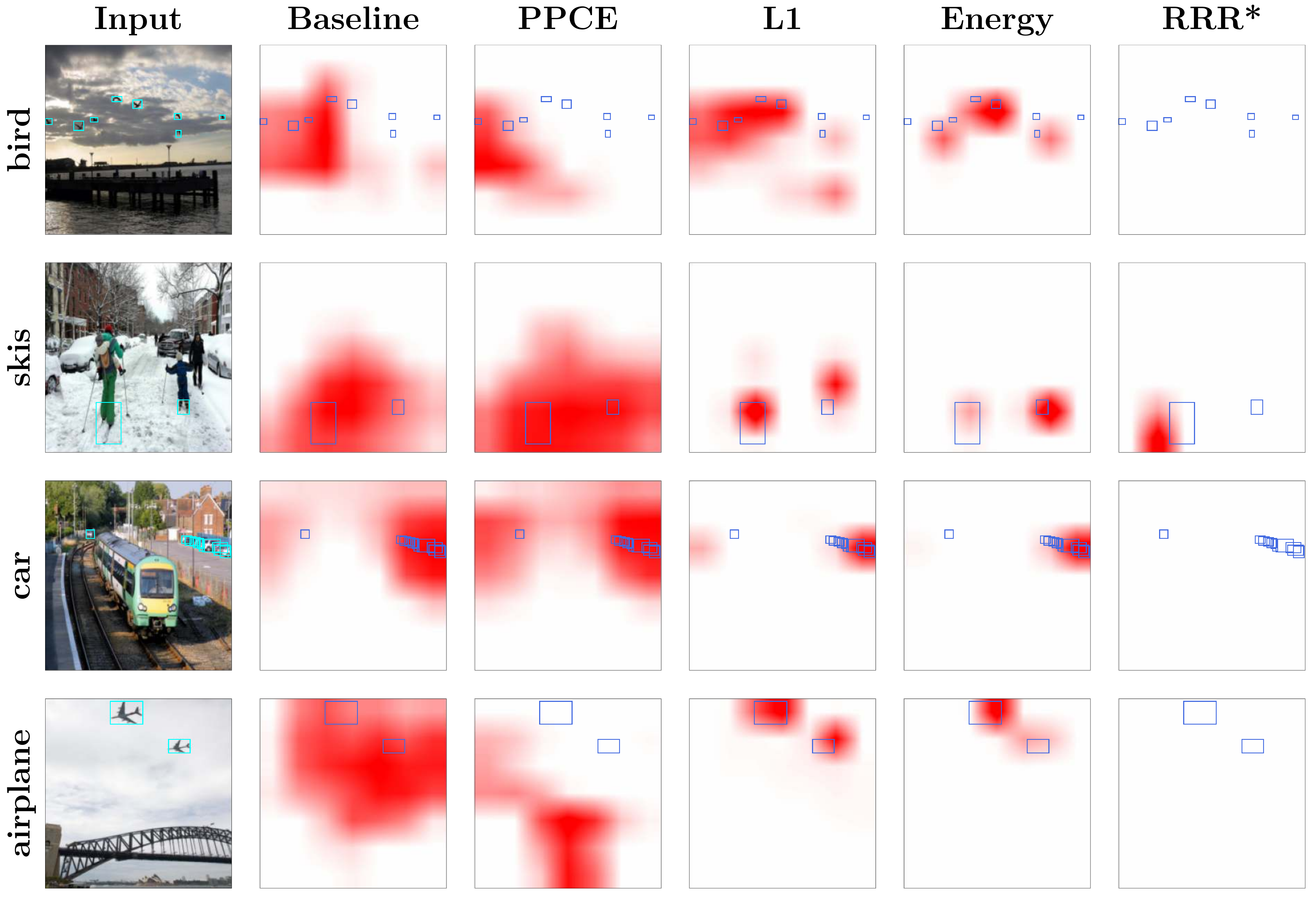}
    \end{subfigure}
    \caption{\textbf{\vanilla \resnet}.}
    \label{fig:supp:quali_coco_1:vanilla}
    \end{subfigure}
    \begin{subfigure}[c]{\textwidth}
    \centering
    \begin{subfigure}[c]{.475\columnwidth}
    \includegraphics[width=\textwidth]{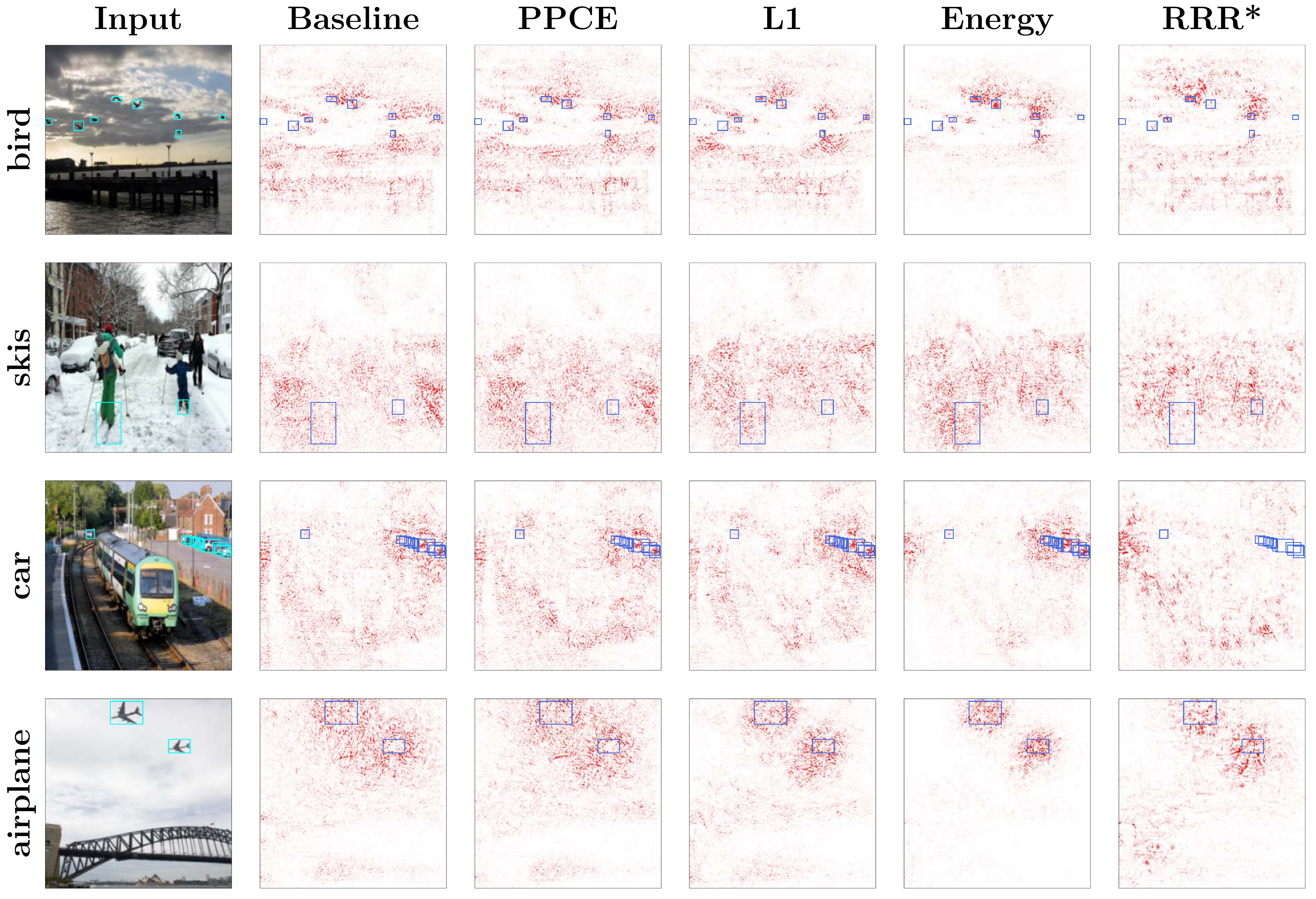}
    \end{subfigure}\hfill
    \begin{subfigure}[c]{.475\columnwidth}
    \includegraphics[width=\textwidth]{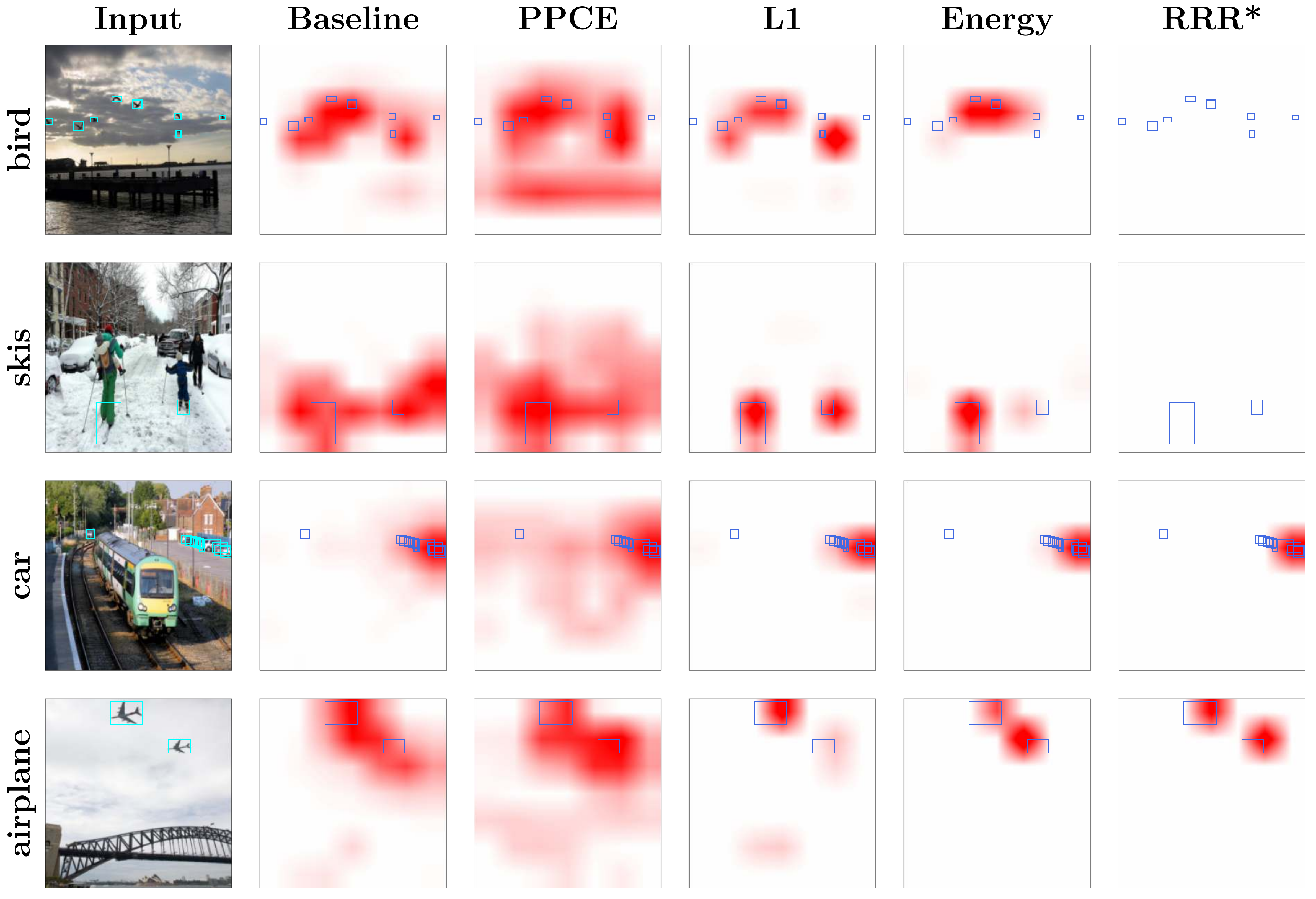}
    \end{subfigure}
    \caption{\textbf{\xdnn \resnet}.}
    \label{fig:supp:quali_coco_1:xdnn}
    \end{subfigure}
    \caption{Qualitative examples from the \textbf{\cocos dataset}. In particular, this figure allows to compare between models (\textbf{major rows}, \ie (a), (b), and (c)) losses (\textbf{major columns}) and layers (\textbf{left+right}) for multiple images (\textbf{minor rows}).}
    \label{fig:supp:quali_coco_1}
\end{figure}

\begin{figure}
    \centering
    \begin{subfigure}[c]{\textwidth}
    \centering
    \textbf{\large Additional qualitative examples.}\\\vspace{.25cm}
    \begin{subfigure}[c]{.485\columnwidth}
    \centering
    \textbf{\voc}\\\vspace{.25cm}
    \includegraphics[width=\textwidth]{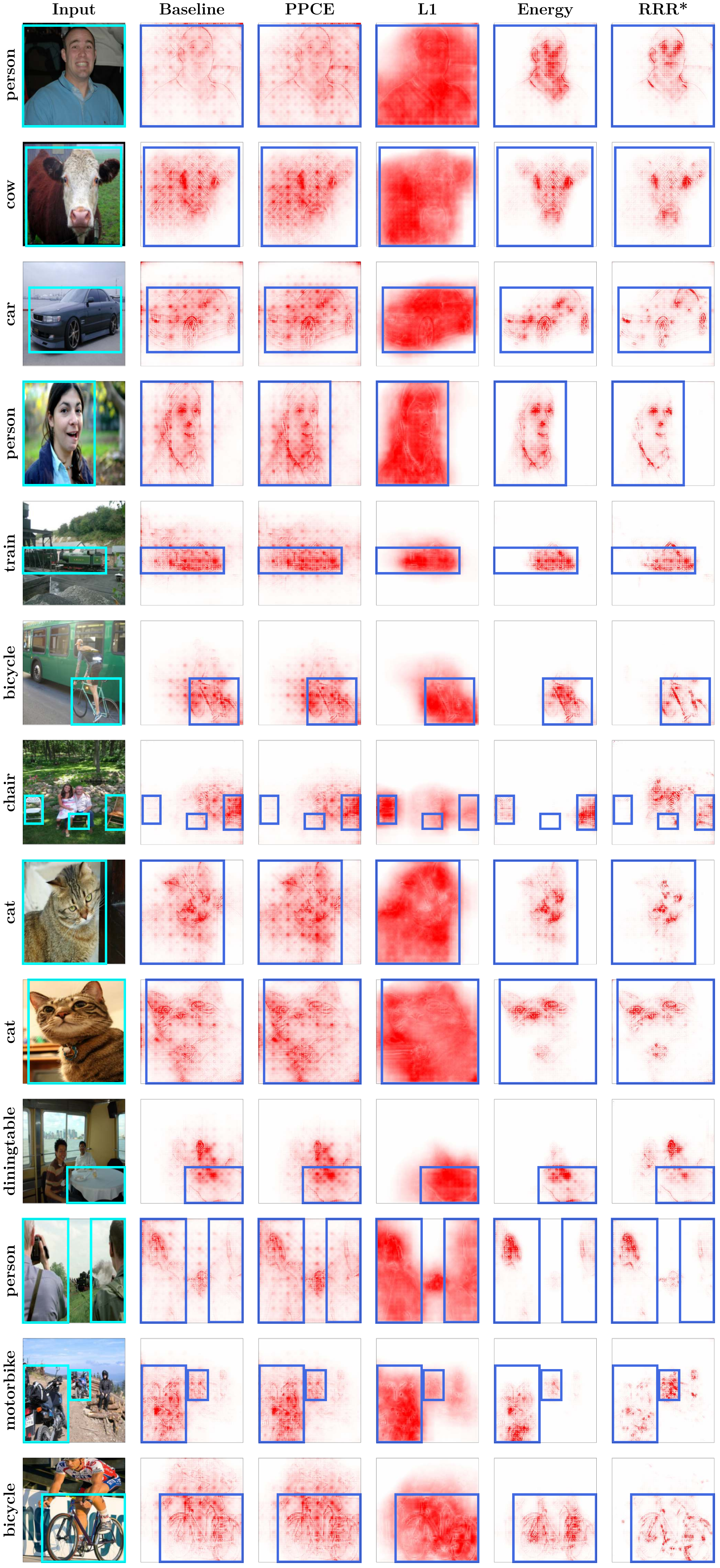}
    \end{subfigure}\hfill
    \begin{subfigure}[c]{.485\columnwidth}
    \centering
    \textbf{\coco}\\\vspace{.25cm}
    \includegraphics[width=\textwidth]{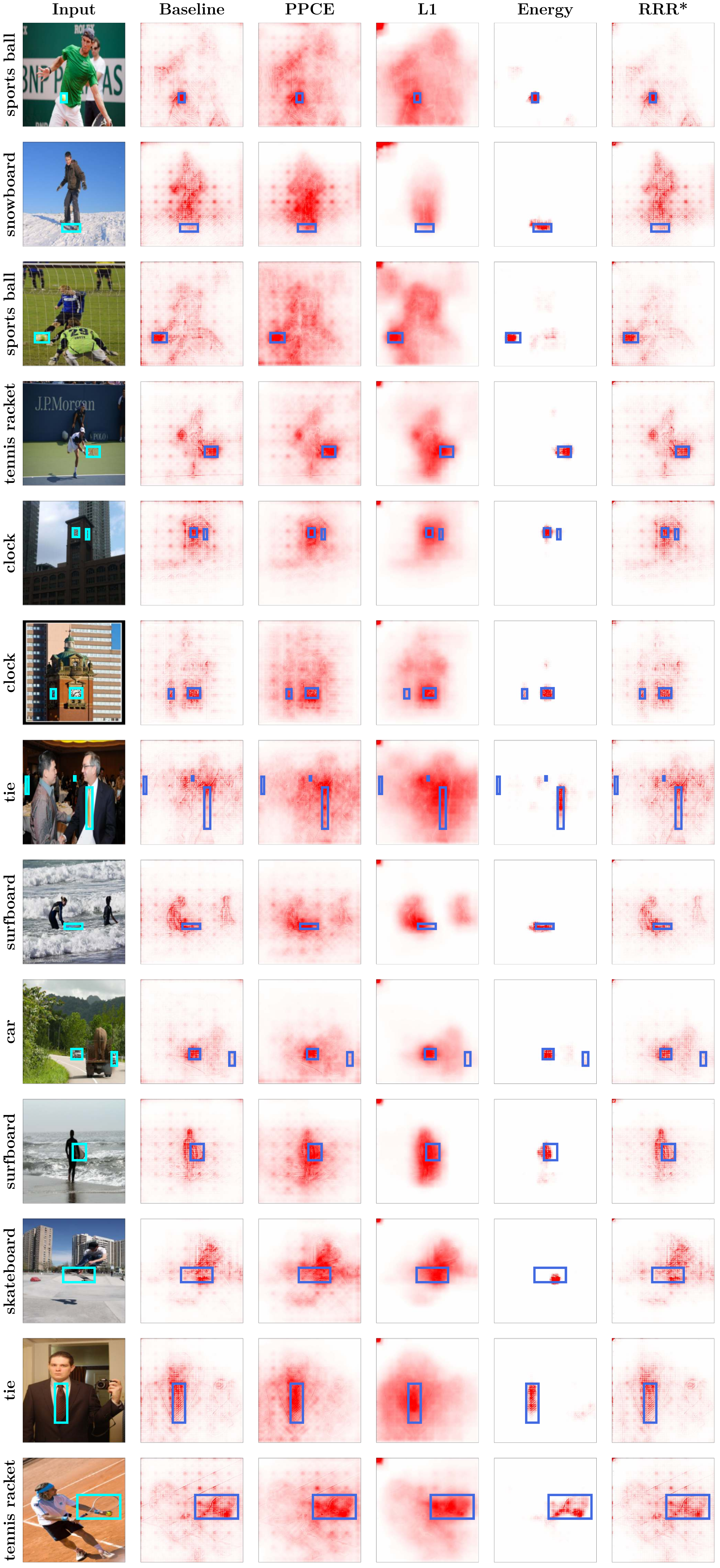}
    \end{subfigure}
    \end{subfigure}
    \caption{Qualitative examples from the \textbf{\vocs (left) and \cocos (right)} datasets. In particular, here we just show additional examples for the \bcos models with input attributions, as this configuration exhibits the most detail. We show results for such models trained with different losses (\textbf{columns}) for multiple images (\textbf{rows}).}
    \label{fig:supp:quali_combined}
\end{figure}

In \cref{fig:supp:quali_voc_1,fig:supp:quali_coco_1}, we visualize attributions across losses, attribution methods, and layers for the same set of examples from the \vocs and \cocos datasets respectively. As discussed in the main paper, we make the following observations.

First, when guiding models at the \emph{final layer}, we observe a marked improvement in the granularity of the attribution maps for all losses (\finding5), except for \ppceloss, for which we do not observe notable differences. The improvements are particularly noticeable on the \cocos dataset (\cref{fig:supp:quali_coco_1}, ``Final'' column), in which the objects tend to be smaller. \Eg, when looking at the airplane image (last row per model), we observe much fewer attributions in the background after applying model guidance.

Second, as the \lone loss optimizes for uniform coverage \emph{within} the bounding boxes, it provides coarser attributions that tend to fill the entire bounding box (cf.~\finding3). This can be observed particularly well for the large objects from the \vocs dataset: \eg, whereas models trained with the \epgloss and the \rrr loss highlight just a relatively small area within the bounding box of the cat (\cref{fig:supp:quali_voc_1}, "Final" column, third row), the \lone loss yields  much more distributed attributions for all models.

Third, at the input layer, the \bcos models show the most notable qualitative improvements (cf.~\finding4). In particular, although the \xdnn models show some reduction in noisy background attributions (\eg last rows in \cref{fig:supp:quali_voc_1:xdnn} and \cref{fig:supp:quali_coco_1:xdnn}), the attributions remain rather noisy for many of the images; for the \vanilla models, the improvements are even less pronounced (\cref{fig:supp:quali_voc_1:vanilla}, \cref{fig:supp:quali_coco_1:vanilla}). The \bcos models, on the other hand, seem to lend themselves better to such guidance being applied to the attributions at the input layer (\cref{fig:supp:quali_voc_1:bcos}, \cref{fig:supp:quali_coco_1:bcos}) and the resulting attributions show much more detail (\epgloss + \rrr) or an increased focus on the entire bounding box (\lone). Especially with the \epgloss, the \bcos models are able to clearly focus on even small objects, see \cref{fig:supp:quali_coco_1:bcos}.

For additional results from both the \vocs as well as the \cocos dataset, please see \cref{fig:supp:quali_combined}.

\clearpage
\subsection{Additional visualizations for training with coarse bounding boxes}
\label{supp:sec:main:qualitative:dilation}
\begin{figure}[h]
    \centering
    \begin{subfigure}[c]{.475\textwidth}
    \includegraphics[width=\textwidth]{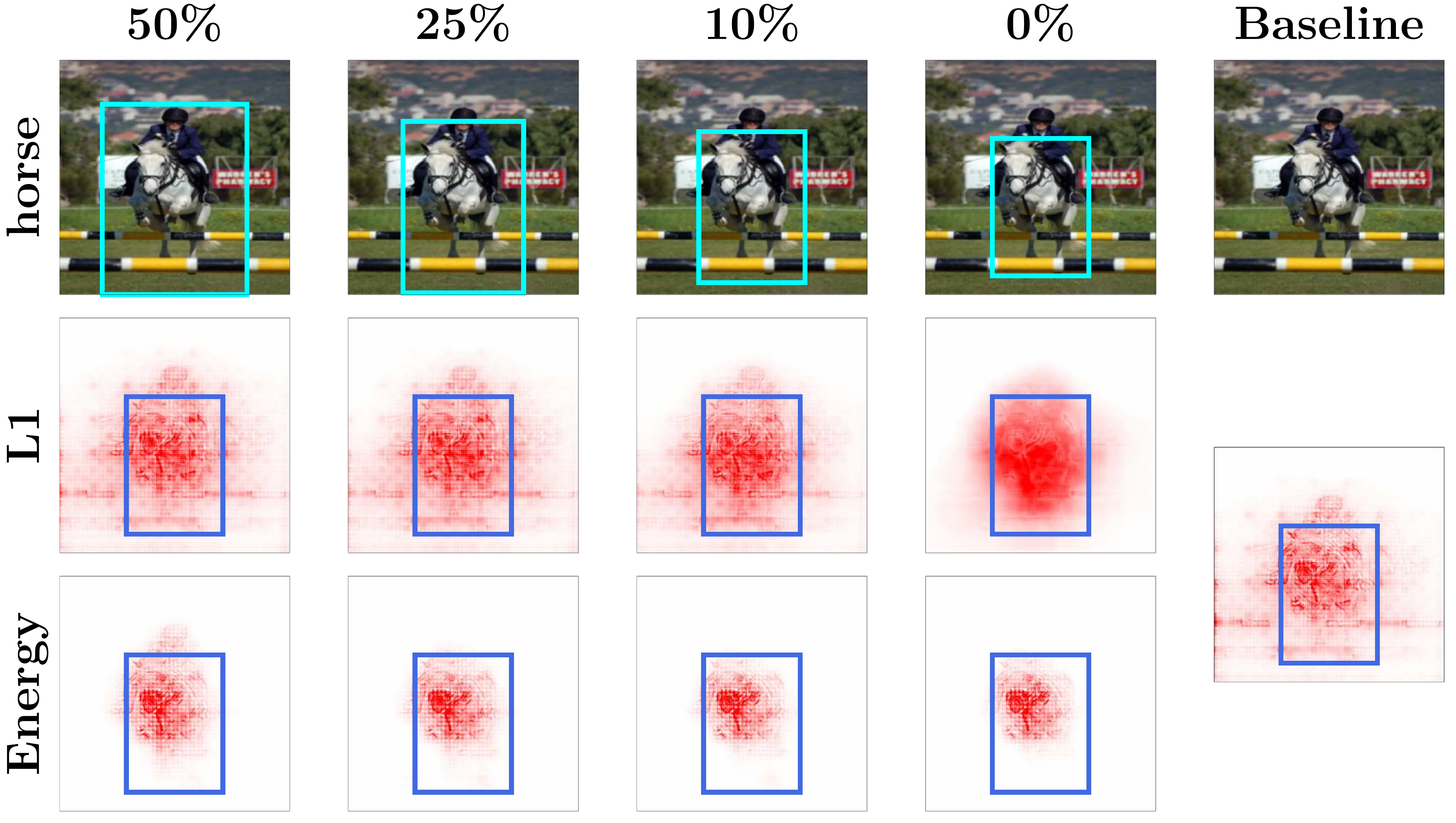}
    \end{subfigure}\hfill
    \begin{subfigure}[c]{.475\textwidth}
    \includegraphics[width=\textwidth]{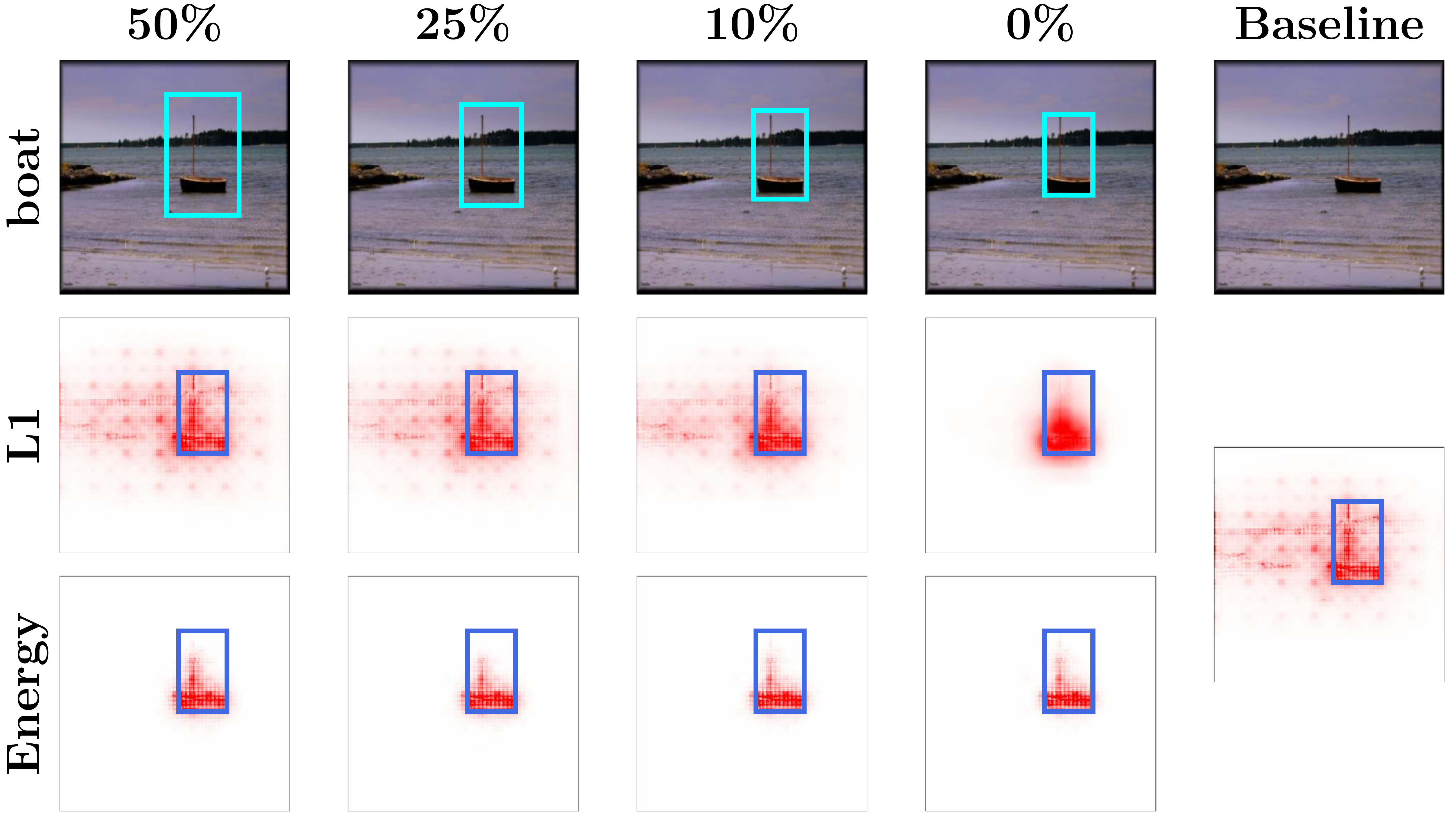}
    \end{subfigure}\hfill
    \begin{subfigure}[c]{.475\textwidth}
    \includegraphics[width=\textwidth]{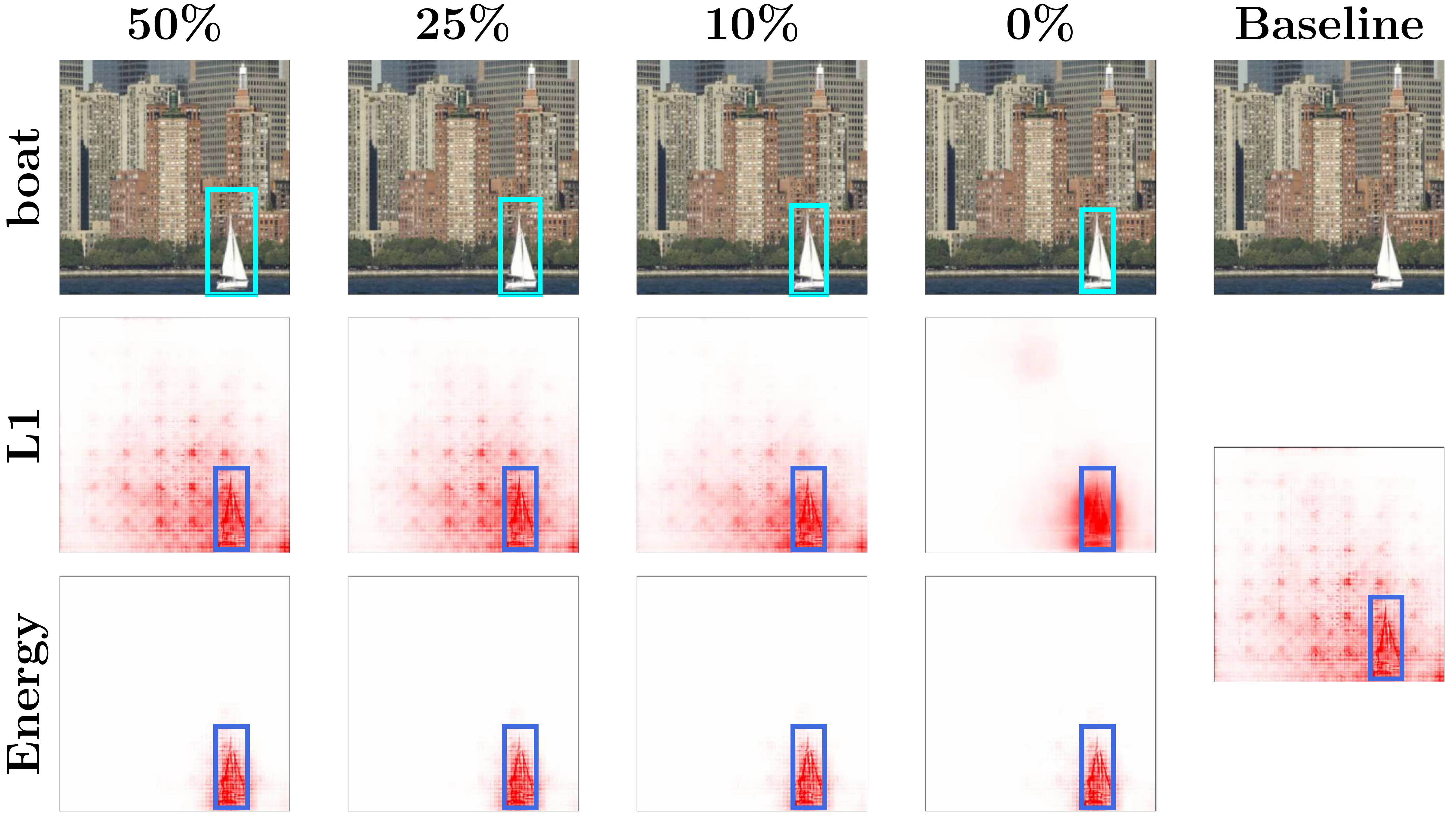}
    \end{subfigure}\hfill
    \begin{subfigure}[c]{.475\textwidth}
    \includegraphics[width=\textwidth]{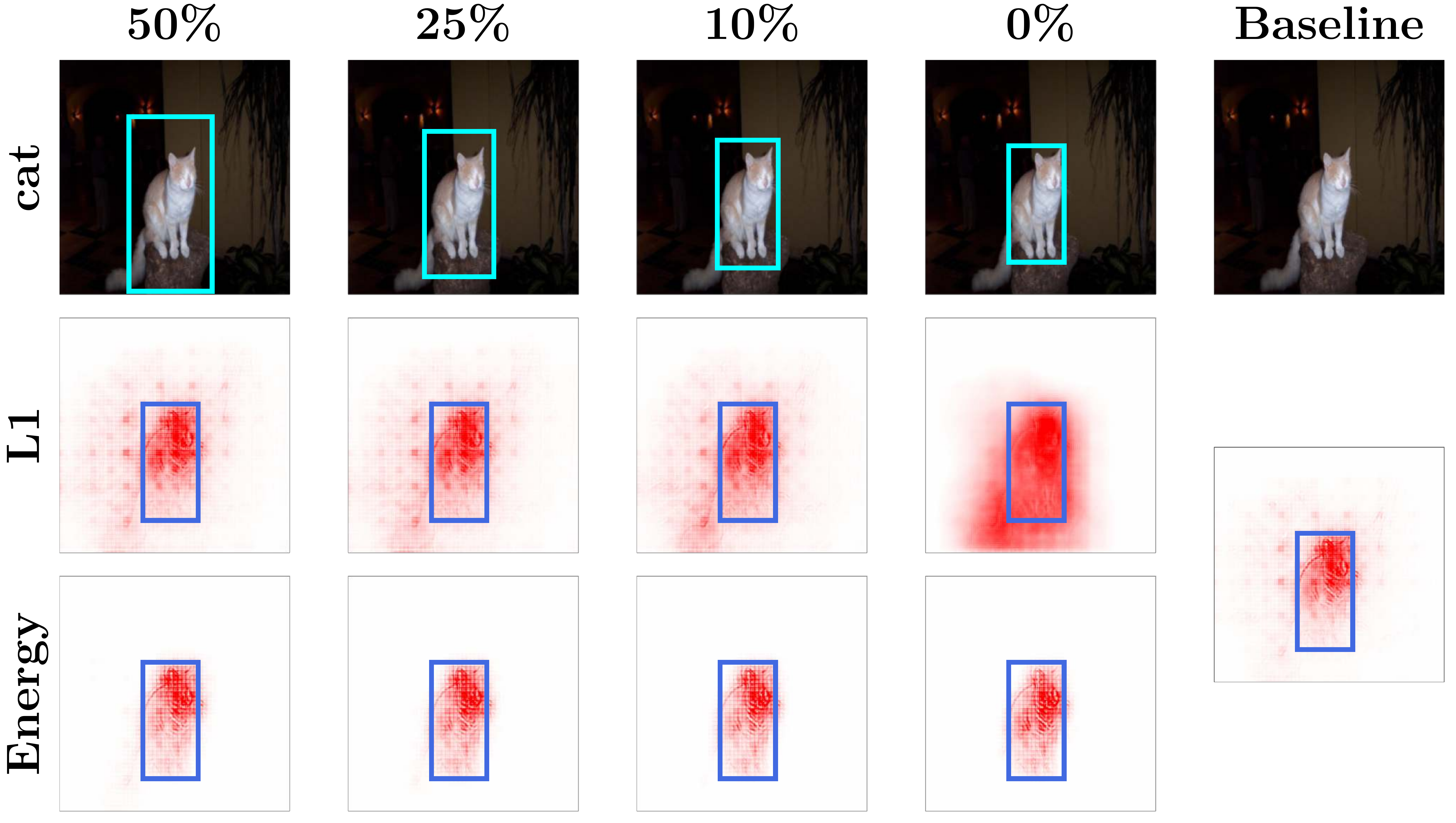}
    \end{subfigure}\hfill
    \begin{subfigure}[c]{.475\textwidth}
    \includegraphics[width=\textwidth]{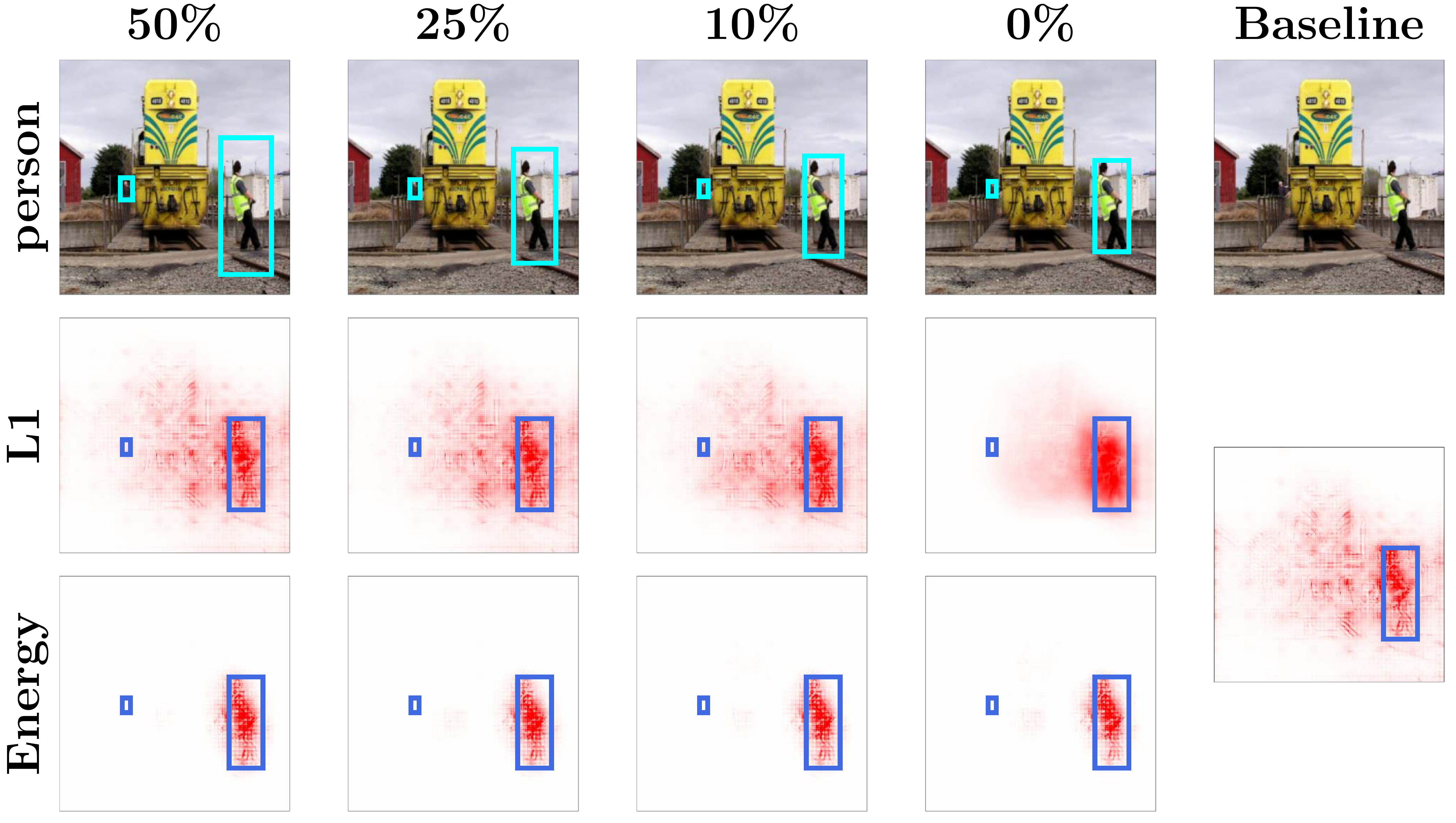}
    \end{subfigure}\hfill
    \begin{subfigure}[c]{.475\textwidth}
    \includegraphics[width=\textwidth]{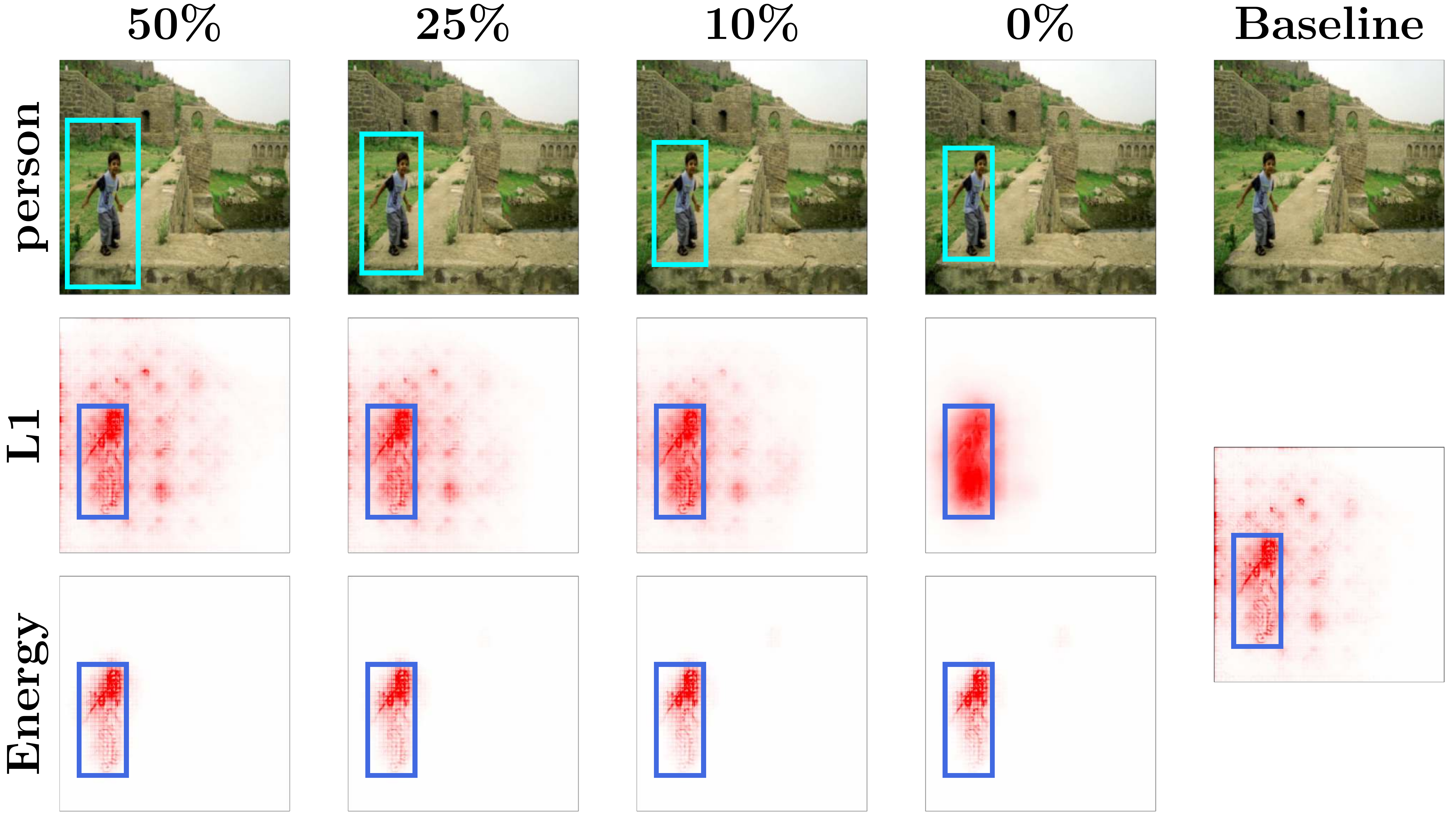}
    \end{subfigure}\hfill
    \caption{\textbf{Qualitative examples of the impact of using coarse bounding boxes for guidance.} We show examples of \bcos attributions from the input layer on the baseline model and on models guided with the \energyloss and \loneloss localization losses with varying degrees of dilation $\{10\%,25\%,50\%\}$ in bounding boxes during training. For each example (\textbf{block} in the figure), we show the image and bounding boxes with varying degrees of dilation (\textbf{top} row), attributions with the \loneloss localization loss (\textbf{middle} row), and attributions with the \energyloss localization loss (\textbf{bottom} row). We find that in contrast to using the \loneloss localization  loss, guidance with \energyloss localization loss maintains localization of attributions to on-object features even with dilated bounding boxes. Note that bounding boxes are dilated only during training, not during evaluation. Bounding boxes in \textbf{light blue} show the extent of dilation that \textit{would have been used} had the image been from the training set, while those in \textbf{dark blue} show undilated bounding boxes that are used during evaluation.}
    \label{fig:supp:dilation_quali}
\end{figure}

In this section, we show more detailed and additional examples of models trained with coarser bounding boxes, \ie with bounding boxes that are purposefully dilated during training by various amounts (10\%, 25\%, or 50\%), see \cref{fig:supp:dilation_quali}. In accordance with our findings in the main paper (cf.~\finding8), we observe that the \epgloss loss is highly robust to such `annotation errors': the attribution maps improve noticeably in all cases (compare the \epgloss row with the respective baseline result). In contrast, the \lone loss seems more dependent on high-quality annotations, which we also observe quantitatively, see \cref{fig:supp:dilation_quanti}.
\clearpage

\section{Additional Quantitative Results (\vocs and \cocos)}
\label{supp:sec:quanti}

In this section, we provide additional quantitative results from our experiments on the \vocs and \cocos datasets. Specifically, in \cref{supp:sec:quantitative:classvsloc}, we show additional results comparing classification and localization performance. In \cref{supp:sec:quantitative:gradcam} we present results for guiding models via \gradcam attributions. In \cref{supp:sec:quantitative:intermediate}, we show that training at intermediate layers can be a cost-effective way approach to performing model guidance.
In \cref{supp:sec:quantitative:segmentepg}, we evaluate how well the attributions localize to on-object features (as opposed to background features) within the bounding boxes, and find that the \energyloss outperforms other localization losses in this regard. In \cref{supp:sec:quantitative:limited}, we provide additional analyses regarding training with a limited number of annotated images. Finally, in \cref{supp:sec:quantitative:dilation}, we provide additional analyses regarding the usage of coarse, dilated bounding boxes during training.

\subsection{Comparing Classification and Localization Performance}
\label{supp:sec:quantitative:classvsloc}
In this section, we discuss additional quantitative findings with respect to localization and classification performance metrics (\iou, \map) for a selected subset of the experiments; for a full comparison of all layers and metrics, please see \cref{fig:supp:voc:f1_results,,fig:supp:coco:f1_results,,fig:supp:voc:map_results:2,,fig:supp:coco:map_results}. 

\begin{figure}[h]
    \centering
    \textbf{\iou results} on {\vocs}.\vspace{.25cm}\\
    \begin{subfigure}[c]{\textwidth}
    \includegraphics[width=\textwidth]{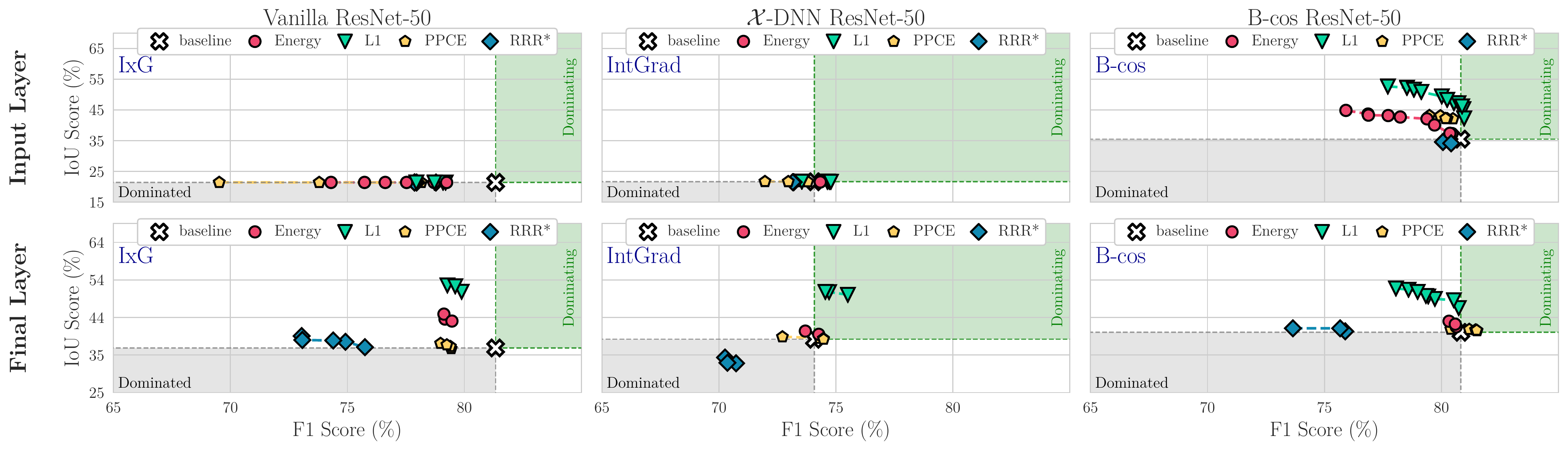}
    \end{subfigure}
    \caption{\textbf{\iou results on \voc.} We show \iou vs.~\fone for all localization loss functions, attribution methods, and layers. In contrast to the consistent improvements observed at the final layer with the \lone loss, the \iou metric only noticeably improves for the \bcos models after model guidance. We attribute this to the high amount of noise present in the attribution maps of \vanilla and \xdnn models, see \eg \cref{fig:supp:quali_voc_1,,fig:supp:quali_coco_1}. For results on the \cocos dataset, please see \cref{fig:sub:iou:coco}.}
    \label{fig:sub:iou:voc}
\end{figure}

\begin{figure}[h]
    \centering
    {\textbf{\iou results} on \cocos}.\vspace{.25cm}\\
    \begin{subfigure}[c]{\textwidth}
    \includegraphics[width=\textwidth]{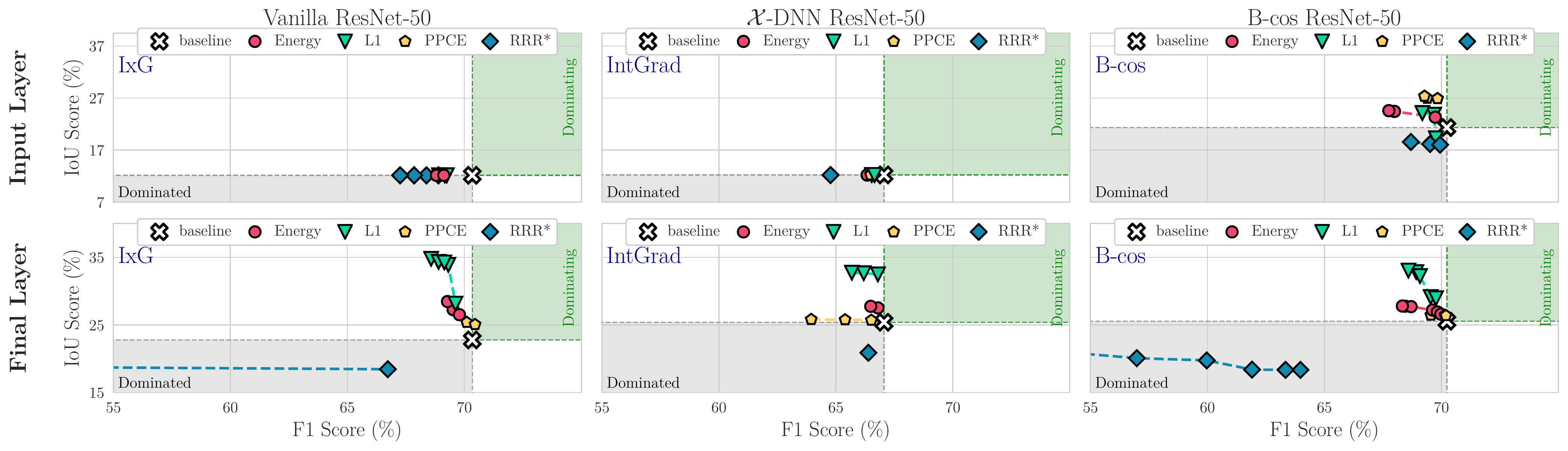}
    \end{subfigure}
    \caption{\textbf{\iou results on \coco.} We show \iou vs.~\fone for all localization loss functions, attribution methods, and layers. In contrast to the consistent improvements observed at the final layer with the \lone loss, the \iou metric only noticeably improves for the \bcos models after model guidance. We attribute this to the high amount of noise present in the attribution maps of \vanilla and \xdnn models, see \eg \cref{fig:supp:quali_voc_1,,fig:supp:quali_coco_1}. For results on the \vocs dataset, please see \cref{fig:sub:iou:voc}.}
    \label{fig:sub:iou:coco}
\end{figure}

\myparagraph{Additional \iou results.} In \cref{fig:sub:iou:voc,fig:sub:iou:coco}, we show the remaining results comparing \iou vs.~\fone scores that were not shown in the main paper for \vocs and \cocos respectively.
Similar to the results in the main paper for the \epg metric (\cref{fig:epg_results}), we find that the results between datasets are highly consistent for the \iou metric. 

In particular, as discussed in \cref{sec:results:epg+iou}, we find that the \lone loss yields the largest improvements in \iou when optimized at the final layer, see bottom rows of \cref{fig:sub:iou:voc,,fig:sub:iou:coco}. At the input layer, we find that \vanilla and \xdnn \resnet models are not improving their \iou scores noticeably, whereas the \bcos models show significant improvements. We attribute this to the noisy patterns in the attribution maps of \vanilla and \xdnn models, which might be difficult to optimize.

\begin{figure}
    \centering
    \vspace{.5cm}
    \textbf{Mean Average Precision (\map) results} on \vocs.\vspace{.25cm}\\
    \begin{subfigure}[c]{.9\textwidth}
    \includegraphics[width=\textwidth]{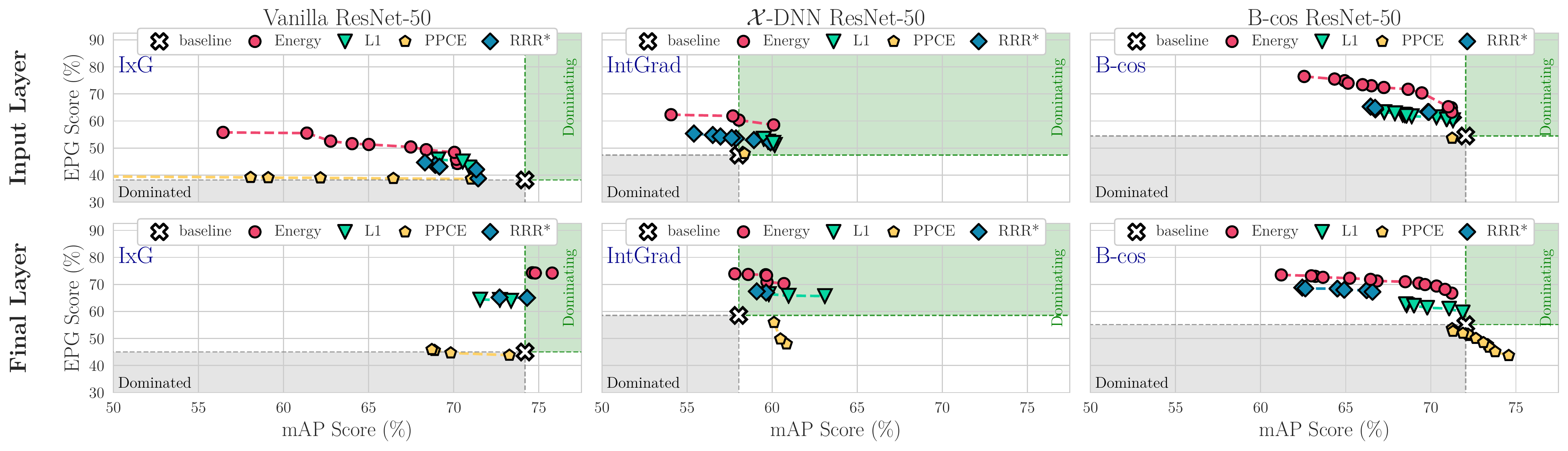}
    \caption{\textbf{\epg vs.~\map.}}
    \end{subfigure}
    \begin{subfigure}[c]{.9\textwidth}
    \includegraphics[width=\textwidth]{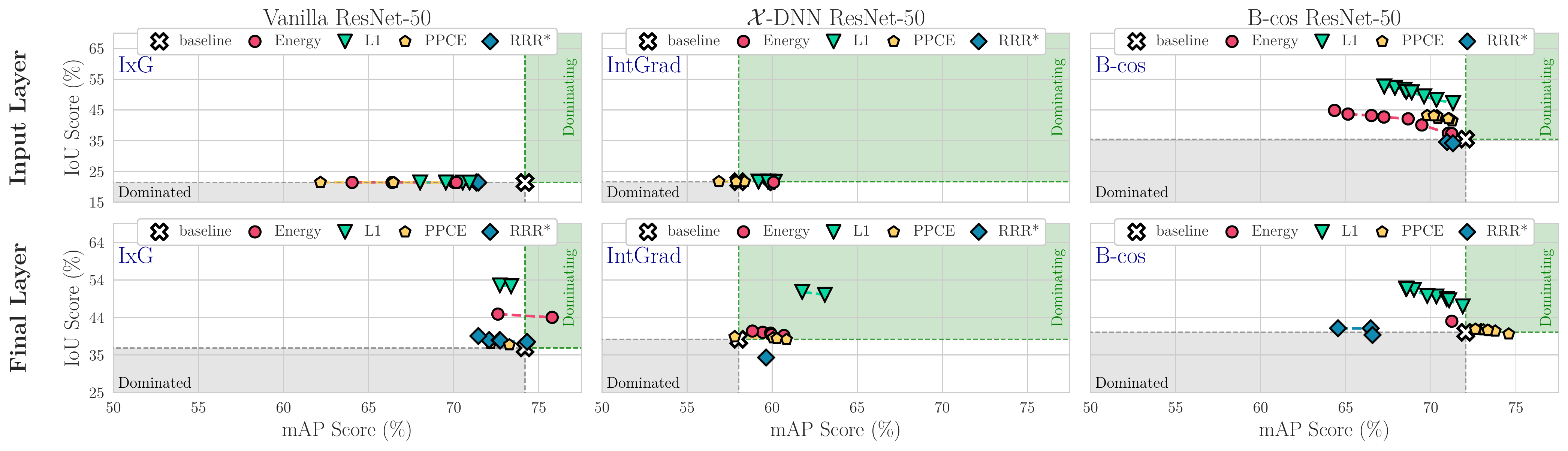}
    \caption{\textbf{\iou vs.~\map.}}
    \end{subfigure}
    \caption{\textbf{Quantitative comparison of \epg and \iou  vs.~\map scores for \vocs.} To ensure that the trends observed and described in the main paper generalize beyond the \fone metric, in this figure we show the \epg and \iou scores plotted against the \map metric. In general, we find the results obtained for the \map metric to be highly consistent with the previously shown results for the \fone metric, see \eg \cref{fig:epg_results,,fig:iou_results}. \Eg, across all configurations, we find the \epgloss to yield the highest gains in \epg score, whereas the \lone loss provides the best trade-offs with respect to the \iou metric. To compare between all results for all datasets and metrics, please see \cref{fig:supp:voc:f1_results,,fig:supp:coco:f1_results,,fig:supp:voc:map_results:1,,fig:supp:coco:map_results}.}
    \label{fig:supp:voc:map_results:1}
\end{figure}

\myparagraph{Using \map to evaluate classification performance.} In all results so far, we plotted the localization metrics (\epg, \iou) versus the \fone score as a measure of classification performance. In order to highlight that the observed trends are independent of this particular choice of metric, in \cref{fig:supp:voc:map_results:1}, we show both \epg as well as \iou results plotted against the \map score. 

In general, we find the results obtained for the \map metric to be highly consistent with the previously shown results for the \fone metric. \Eg, across all configurations, we find the \epgloss to yield the highest gains in \epg score, whereas the \lone loss provides the best trade-offs with respect to the \iou metric. In order to easily compare between all results for all datasets and metrics, please see \cref{fig:supp:voc:f1_results,,fig:supp:coco:f1_results,,fig:supp:voc:map_results:2,,fig:supp:coco:map_results}.

\subsection{Model Guidance via \gradcam}
\label{supp:sec:quantitative:gradcam}
\begin{figure}[h]
    \centering
    {\textbf{Comparison to \gradcam} on {\vocs}.}\vspace{.25cm}\\
    \begin{subfigure}[c]{\textwidth}
    \includegraphics[width=\textwidth]{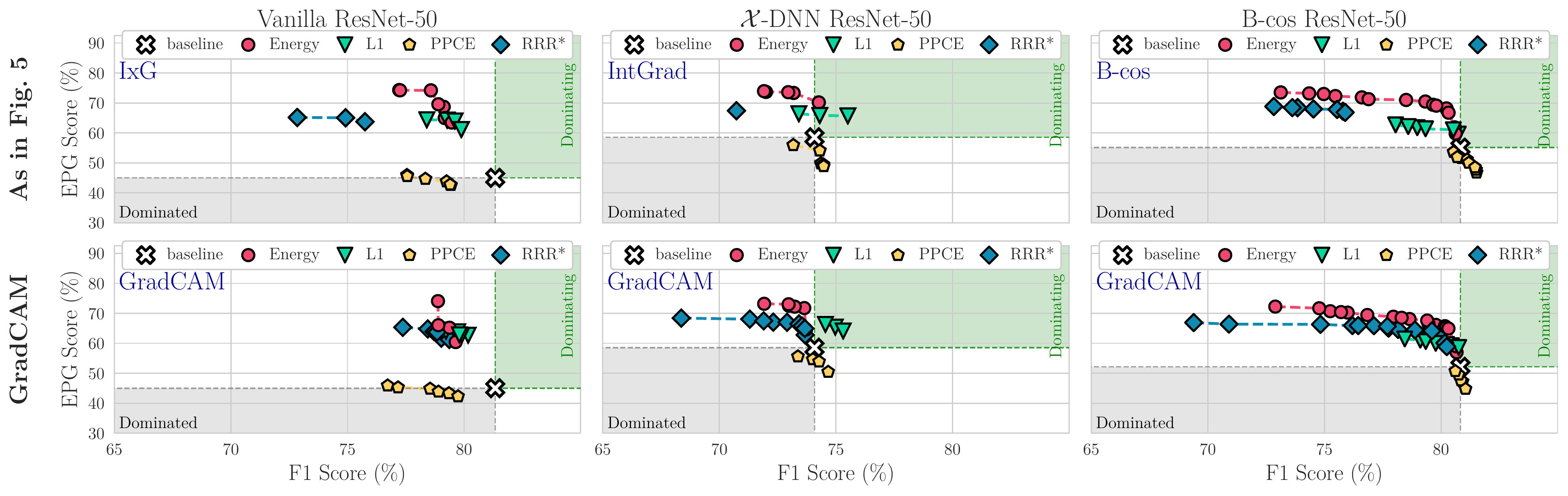}
    \end{subfigure}
    \caption{\textbf{Quantitative results using \gradcam.} We show \epg scores vs.~F1 scores for all localization losses and models using \gradcam at the final layer (\textbf{bottom row}) and compare it to the results shown in the main paper (\textbf{top row}). 
    As expected, \gradcam performs very similarly to \ixg (\vanilla) and \intgrad (\xdnn) used at the final layer---in particular, note that for \resnet architectures, \ixg and \intgrad are very similar to \gradcam for \vanilla and \xdnn models respectively (see \cref{supp:sec:quantitative:gradcam}). Similarly, we find \gradcam to also perform comparably to the \bcos explanations when used at the final layer; for \iou results and results on \cocos, see \cref{fig:supp:gradcam:voc,,fig:supp:gradcam:coco}.
    }
    \label{fig:supp:gradcam_epg_voc}
\end{figure}

\begin{figure}[h]
    \centering
    {\textbf{Comparison to \gradcam} on {\vocs}.}\vspace{.25cm}\\
    \begin{subfigure}[c]{\textwidth}
    \includegraphics[width=\textwidth]{results/VOC/figures/loc/gradcam_f1.pdf}
    \caption{\textbf{\epg vs.~F1.}}
    \end{subfigure}
    \begin{subfigure}[c]{\textwidth}
    \includegraphics[width=\textwidth]{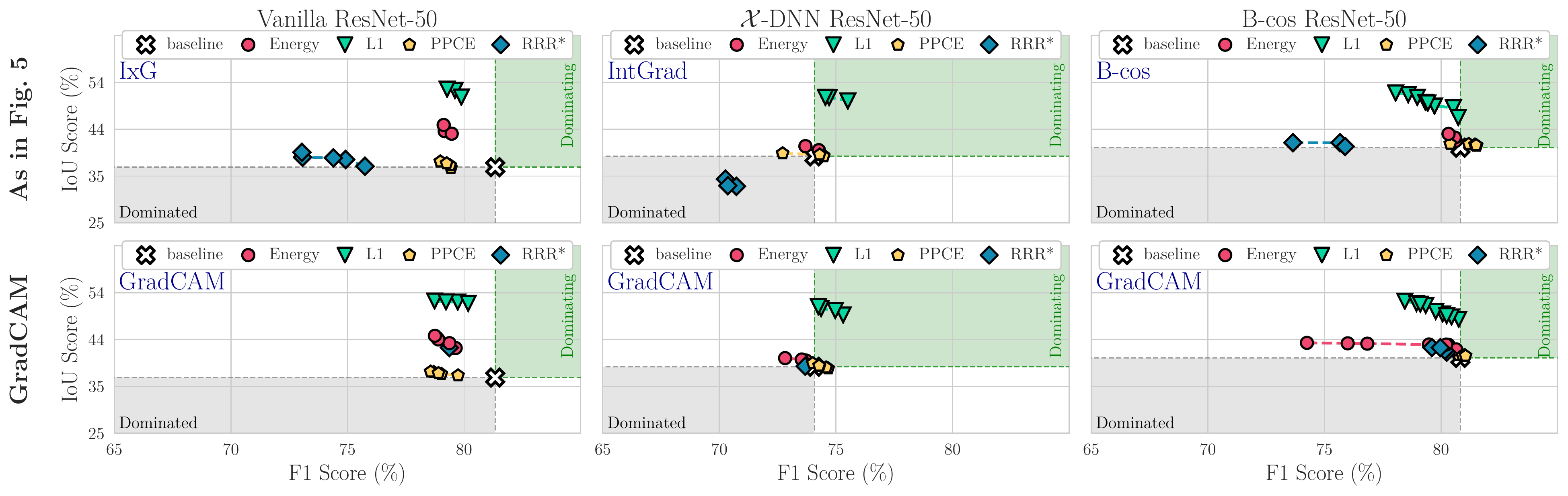}
    \caption{\textbf{\iou vs.~F1.}}
    \end{subfigure}
    \caption{
    \textbf{Quantitative results using \gradcam on \vocs.} We show \epg \textbf{(a)} and  \iou \textbf{(b)} scores vs.~F1 scores  for all localization losses and models using \gradcam at the final layer (\textbf{bottom rows} in (a)+(b) and compare it to the results shown in the main paper (\textbf{top rows}). 
    As expected, \gradcam performs very similarly to \ixg (\vanilla) and \intgrad (\xdnn) used at the final layer---in particular, note that for \resnet architectures, \ixg and \intgrad are very similar to \gradcam for \vanilla and \xdnn models respectively (see \cref{supp:sec:quantitative:gradcam}). Similarly, we find \gradcam to also perform comparably to the \bcos explanations when used at the final layer; for results on \cocos, see \cref{fig:supp:gradcam:coco}.}
    \label{fig:supp:gradcam:voc}
\end{figure}

\begin{figure}[h]
    \centering
    {\textbf{Comparison to \gradcam} on {\cocos}.}\vspace{.25cm}\\
    \begin{subfigure}[c]{\textwidth}
    \includegraphics[width=\textwidth]{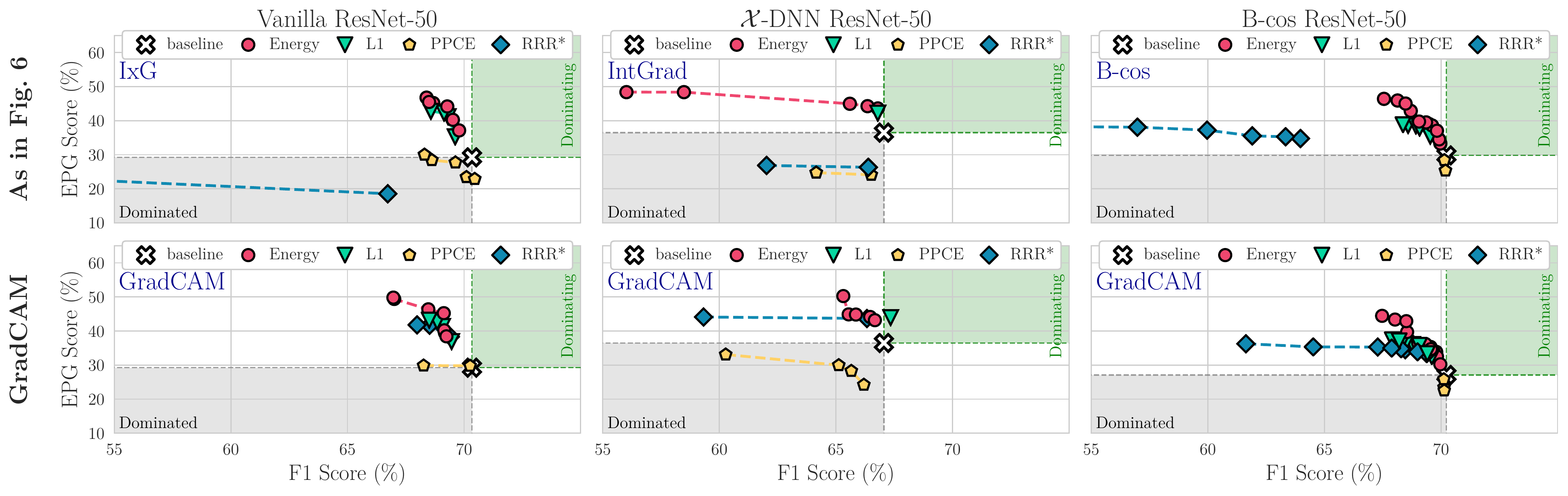}
    \caption{\textbf{\epg vs.~F1.}}
    \end{subfigure}
    \begin{subfigure}[c]{\textwidth}
    \includegraphics[width=\textwidth]{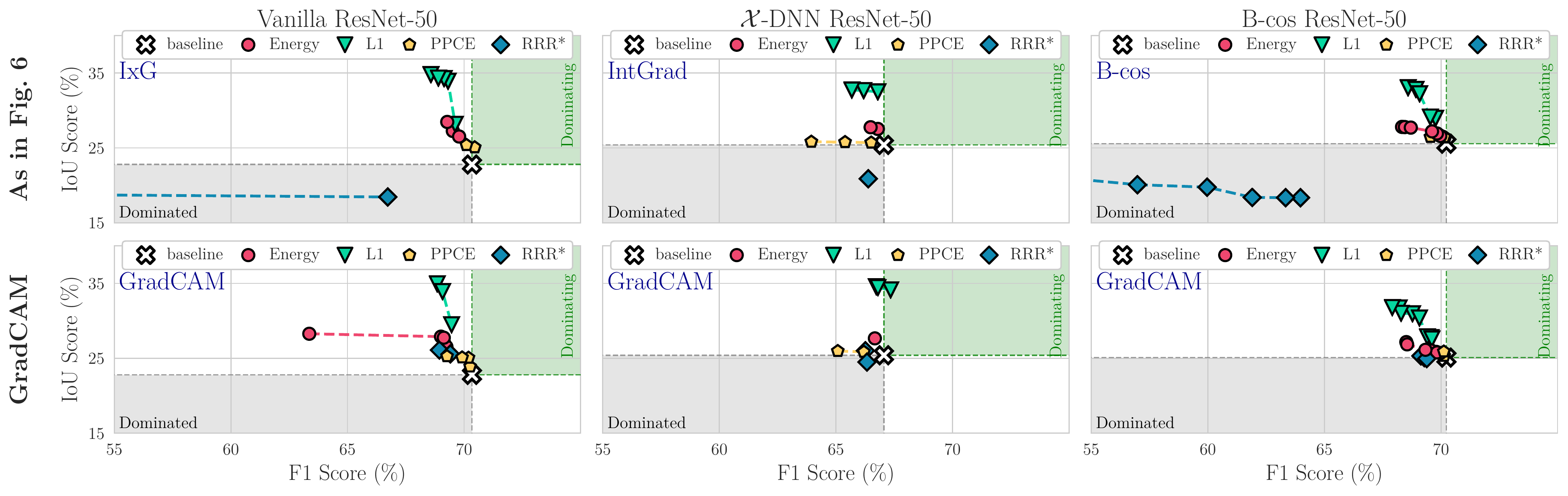}
    \caption{\textbf{\iou vs.~F1.}}
    \end{subfigure}
    \caption{\textbf{Quantitative results using \gradcam on \cocos.} We show \epg \textbf{(a)} and  \iou \textbf{(b)} scores vs.~F1 scores  for all localization losses and models using \gradcam at the final layer (\textbf{bottom rows} in (a)+(b) and compare it to the results shown in the main paper (\textbf{top rows}). 
    As expected, \gradcam performs very similarly to \ixg (\vanilla) and \intgrad (\xdnn) used at the final layer---in particular, note that for \resnet architectures, \ixg and \intgrad are very similar to \gradcam for \vanilla and \xdnn models respectively (see \cref{supp:sec:quantitative:gradcam}). Similarly, we find \gradcam to also perform comparably to the \bcos explanations when used at the final layer; for results on \vocs, see \cref{fig:supp:gradcam:voc}.}
    \label{fig:supp:gradcam:coco}
\end{figure}

In \cref{fig:supp:gradcam_epg_voc}, we show the \epg vs.~\fone results of training models with \gradcam applied at the final layer on the \vocs dataset; for \iou results and results on \cocos, please see \cref{fig:supp:gradcam:voc,fig:supp:gradcam:coco}. When comparing between rows (\textbf{top:} main paper results; \textbf{bottom:} \gradcam), it becomes clear that \gradcam performs very similarly to \ixg\ / \intgrad\  / \bcos attributions on \vanilla\ / \xdnn\ / \bcos models. In fact, note that \gradcam is very similar to \ixg and \intgrad (equivalent up to an additional zero-clamping) for the respective models and any differences in the results can be attributed to the non-deterministic training pipeline and the similarity between the results should thus be expected.

\subsection{Model Guidance at Intermediate Layers}
\label{supp:sec:quantitative:intermediate}

In \cref{sec:results}, we show results for guidance on two `model depths', \ie at the input and the final layer. This corresponds to the two depths at which attributions are typically computed, \eg \ixg and \intgrad are typically computed at the input, while \gradcam is typically computed using final spatial layer activations. Following \citeApp{rao2022towards}, for a fair comparison we optimize using each attribution methods at identical depths. For the final and intermediate layers in the network, this is done by treating the output activations at that layer as effective inputs over which attributions are to be computed. As done with \gradcam \citeApp{selvaraju2017grad}, we then upscale the attribution maps to image dimensions using bilinear interpolation and then use them for model guidance.

\begin{figure}[h]
    \centering    
    {\epg results for \textbf{intermediate layers} on \vocs.}\vspace{.25cm}\\
    \begin{subfigure}[c]{\textwidth}
    \includegraphics[width=\textwidth]{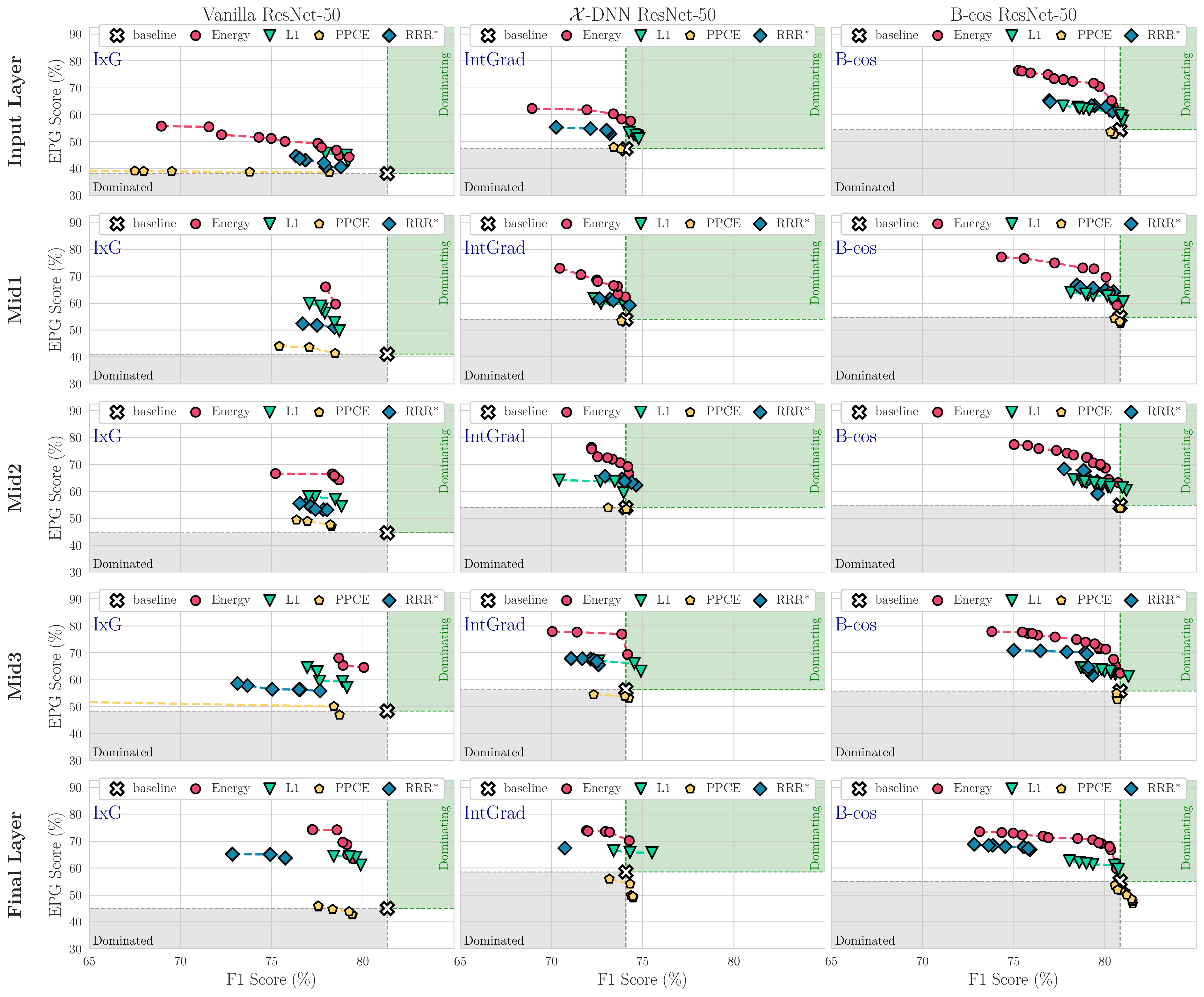}
    \end{subfigure}
    \caption{\textbf{Intermediate layer results comparing \epg vs.~F1.} We compare the effectiveness of model guidance at varying network depths (\textbf{rows}) for each attribution method and model (\textbf{columns}) across localization loss functions. For the \bcos model, we find similar trends at all network depths, with the \energyloss localization loss outperforming all other losses. For the \vanilla and \xdnn models, the \energyloss loss similarly performs the best, but we also observe improved performance across losses when optimizing at deeper layers of the network. Full results can be found in \cref{fig:intermediate:epg,fig:intermediate:iou}.}
    \label{fig:sub:intermediate:epg}
\end{figure}

In \cref{fig:sub:intermediate:epg}, we show results for performing model guidance at additional intermediate layers: Mid1, Mid2, and Mid3. Specifically, for the \resnet models we use, these layers correspond to the outputs of \verb|conv2_x|, \verb|conv3_x|, and \verb|conv4_x| respectively in the ResNet nomenclature (\citeApp{he2016deep}), while the final layer corresponds to the output of \verb|conv5_x|. We find that the \epg performance at these intermediate layers through the network follows the trends when moving from the input to the final layer. Similar results for \iou can be found in \cref{fig:intermediate:iou}.

\subsection{Evaluating On-Object Localization}
\label{supp:sec:quantitative:segmentepg}

\begin{figure}
    \centering    
    {Evaluating \textbf{on-object localization} within bounding boxes.}\vspace{.25cm}\\
    \begin{subfigure}[c]{\textwidth}
    \centering
    \includegraphics[width=.55\textwidth]{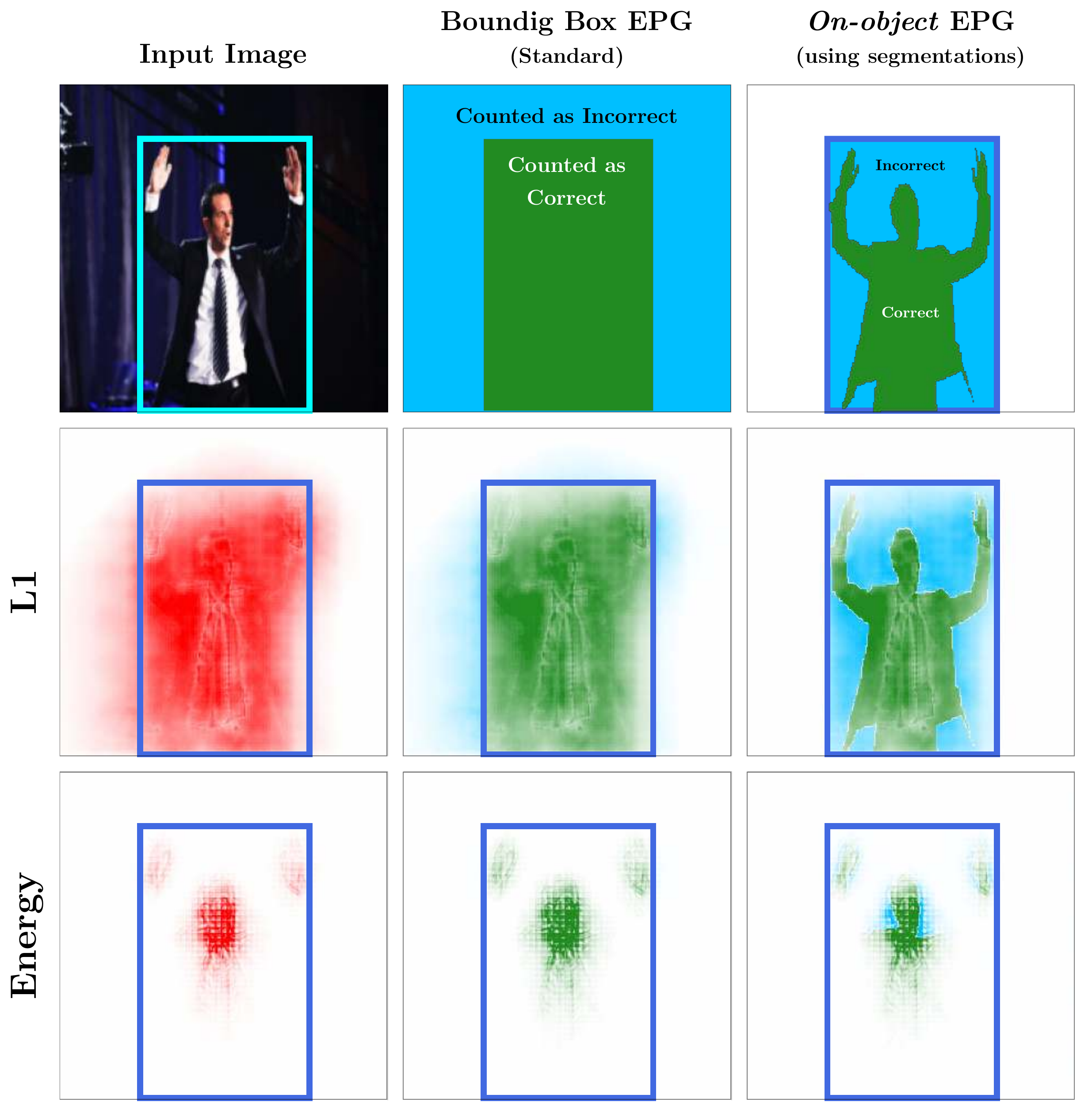}
    \caption{\textbf{Evaluating \emph{on-object} localization within the bounding boxes: On-object \epg.} In the standard \epg metric (\textbf{middle} column), we compute the fraction of positive attributions within the bounding boxes. In other words, attributions within the bounding box (\textbf{green} region) positively impact the metric, while attributions outside (\textbf{blue} region) negatively impact it. Since bounding boxes are coarse annotations and often include background regions, the standard \epg does not evaluate how well attributions localize \textit{on-object} features, \eg the person in the figure. To measure this, we evaluate with an additional Segmentation \epg metric (\textbf{right} column), where we compute the fraction of positive attributions in the bounding box that lie within the segmentation mask of the object. Here, attributions within the segmentation mask (\textbf{green} region) positively impact the metric, and attributions outside the segmentation mask and inside the bounding box (\textbf{blue} region) negatively impact it. Note that attributions outside the bounding box have no effect on Segmentation \epg. As an example and to visualize qualitative differences between losses, in the bottom rows (\lone, \epgloss), we show attributions for a \bcos model guided at the input layer. As becomes clear, by employing a uniform prior on attributions within the bounding box, the \lone loss is effectively  optimized to fill the entire bounding box and thus to not only highlight \emph{on-object} features. This can also be observed quantitatively, see \eg \cref{fig:seg_epg:epg_vs_f1}, right column.}
    \label{fig:seg_epg:schema}
    \end{subfigure}
    \begin{subfigure}[c]{\textwidth}
    \includegraphics[width=\textwidth]{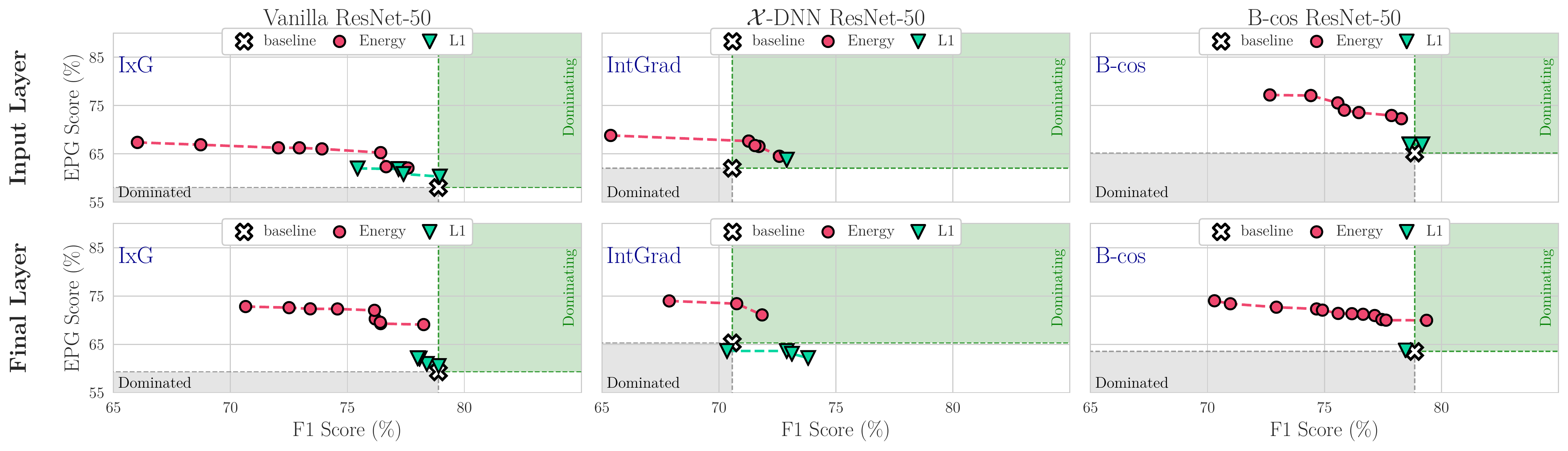}
    \caption{\textbf{On-object \epg results.} We evaluate across models (\textbf{columns}) and layers (\textbf{rows}) for the \energyloss and \loneloss localization losses. As seen qualitatively (\eg \cref{fig:loss_comp}), we find that the \energyloss loss is more effective than the \loneloss loss in localizing attributions to the object as opposed to background regions within the bounding boxes. This is explained by the fact that the \loneloss loss promotes uniformity in attributions within the bounding box, and treats both on-object and background features inside the box with equal importance, while the \energyloss loss only optimizes for attributions to lie within the bounding box without placing any constraints on where they may lie, leaving the model free to decide which regions within the box are important for its decision.}
    \label{fig:seg_epg:epg_vs_f1}
    \end{subfigure}
    \caption{\textbf{Evaluating \emph{on-object} localization via \epg.} We show \textbf{(a)} the schema for the on-object \epg metric and how it differs the standard bounding box \epg metric, and \textbf{(b)} quantitative results on evaluating with on-object \epg.}
    \label{fig:seg_epg}
\end{figure}

The standard \epg metric (\cref{eq:epg}) evaluates the extent to which attributions localize to the bounding boxes. However, since such boxes often include background regions, the \epg score does not distinguish between attributions that focus on the object and attributions that focus on such background regions within the bounding boxes. 

To additionally evaluate for on-object localization, we use a variant of \epg that we call On-object \epg. In contrast to standard \epg, we compute the fraction of positive attributions in pixels contained within the segmentation mask of the object out of positive attributions within the bounding box. This measures how well attributions \textit{within the bounding boxes} localize to the object, and is not influenced by attributions outside the bounding boxes. A visual comparison of the two metrics is shown in \cref{fig:seg_epg}.

We find that the \energyloss localization loss outperforms the \loneloss localization loss both qualitatively (\cref{fig:seg_epg:schema}) and quantitatively (\cref{fig:seg_epg:epg_vs_f1}) on this metric. This is explained by the fact that the \loneloss promotes uniformity in attributions across the bounding box, giving equal importance to on-object and background features within the box. In contrast, the \energyloss loss only optimizes for attributions to lie within the box, without any constraint on \textit{where} in the box they lie. This also corroborates our previous qualitative observations (\eg \cref{fig:loss_comp}).

\subsection{Model Guidance with Limited Annotations}
\label{supp:sec:quantitative:limited}
\begin{figure}
    \centering
    {Additional results for training with \textbf{limited annotations}}\vspace{.5cm}\\
    \begin{subfigure}[c]{\textwidth}
    \begin{subfigure}[c]{.49\textwidth}
    \centering
    \textbf{\epg score}\\\vspace{.2cm}
    \includegraphics[width=\textwidth]{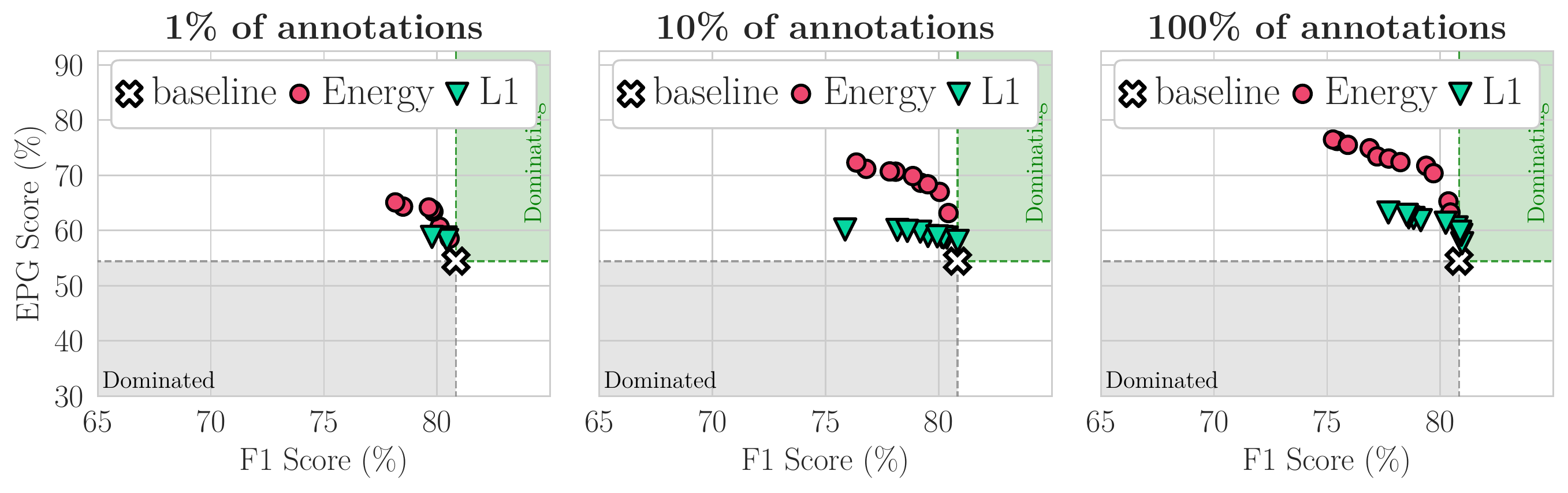}
    \end{subfigure}\hfill
    \begin{subfigure}[c]{.49\textwidth}
    \centering
    \textbf{\iou score}\\\vspace{.2cm}
    \includegraphics[width=\textwidth]{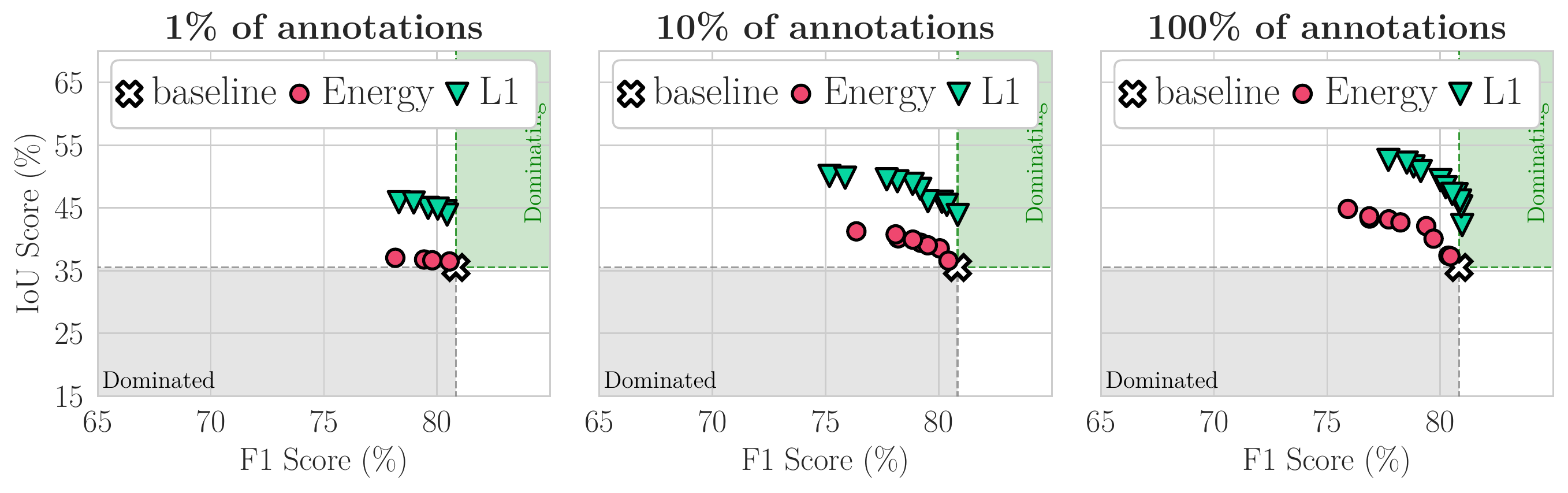}
    \end{subfigure}
    \end{subfigure}
    \caption{\textbf{\epg and \iou scores for limited annotations.} We show \epg vs.~F1 (\textbf{left}) and \iou vs.~F1 (\textbf{right}) for \bcos attributions at the input when optimizing with the \energyloss and \loneloss localization losses, when using $\{1\%,10\%,100\%\}$ training annotations. We find that model guidance is generally effective even when training with annotations for a limited number of images. While the performance slightly worsens when using 1\% annotations, using just 10\% annotated images yields similar gains to using a fully annotated training set. Full results can be found in \cref{fig:supp:limited:full:input,fig:supp:limited:full:final}.}
    \label{fig:supp:limited:sub}
\end{figure}

In \cref{fig:supp:limited:sub}, we show the impact of using limited annotations when training (\cref{sec:results:ablations}) when optimizing with the \energyloss and \loneloss localization losses for \bcos attributions at the input. We find that in addition to \epg, trends in \iou scores also remain consistent even when using bounding boxes for just 1\% of the the training images.

\subsection{Model Guidance with Noisy Annotations} 
\label{supp:sec:quantitative:dilation}
\begin{figure}[h]
    \centering
    {Additional results for training with \textbf{coarse bounding boxes}}\vspace{.5cm}\\
    \begin{subfigure}[c]{.495\textwidth}
    \centering
    \includegraphics[width=\textwidth]{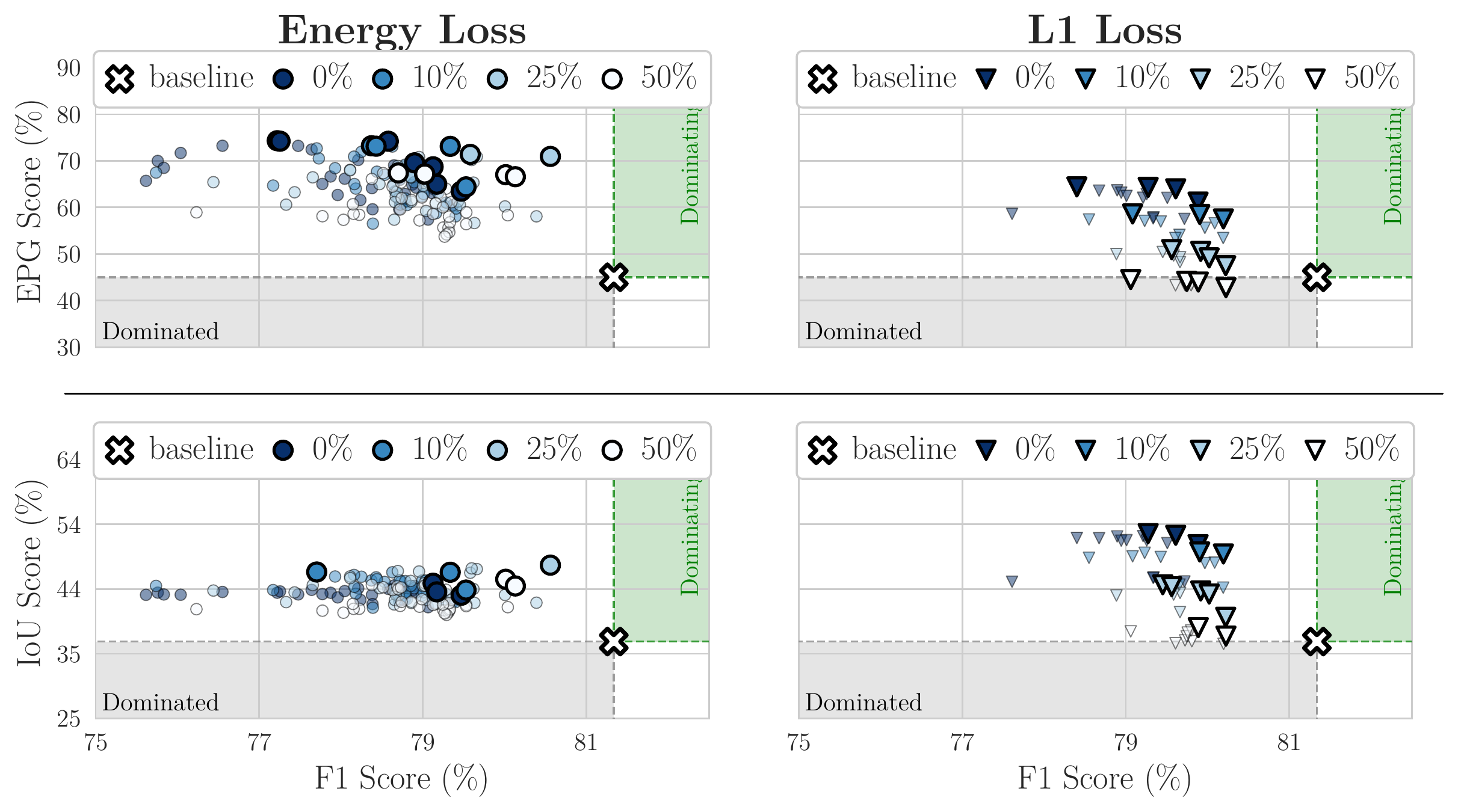}
    \caption{\textbf{\vanilla \resnet  @ Final.}}
    \end{subfigure}\hfill
    \begin{subfigure}[c]{.495\textwidth}
    \centering
    \includegraphics[width=\textwidth]{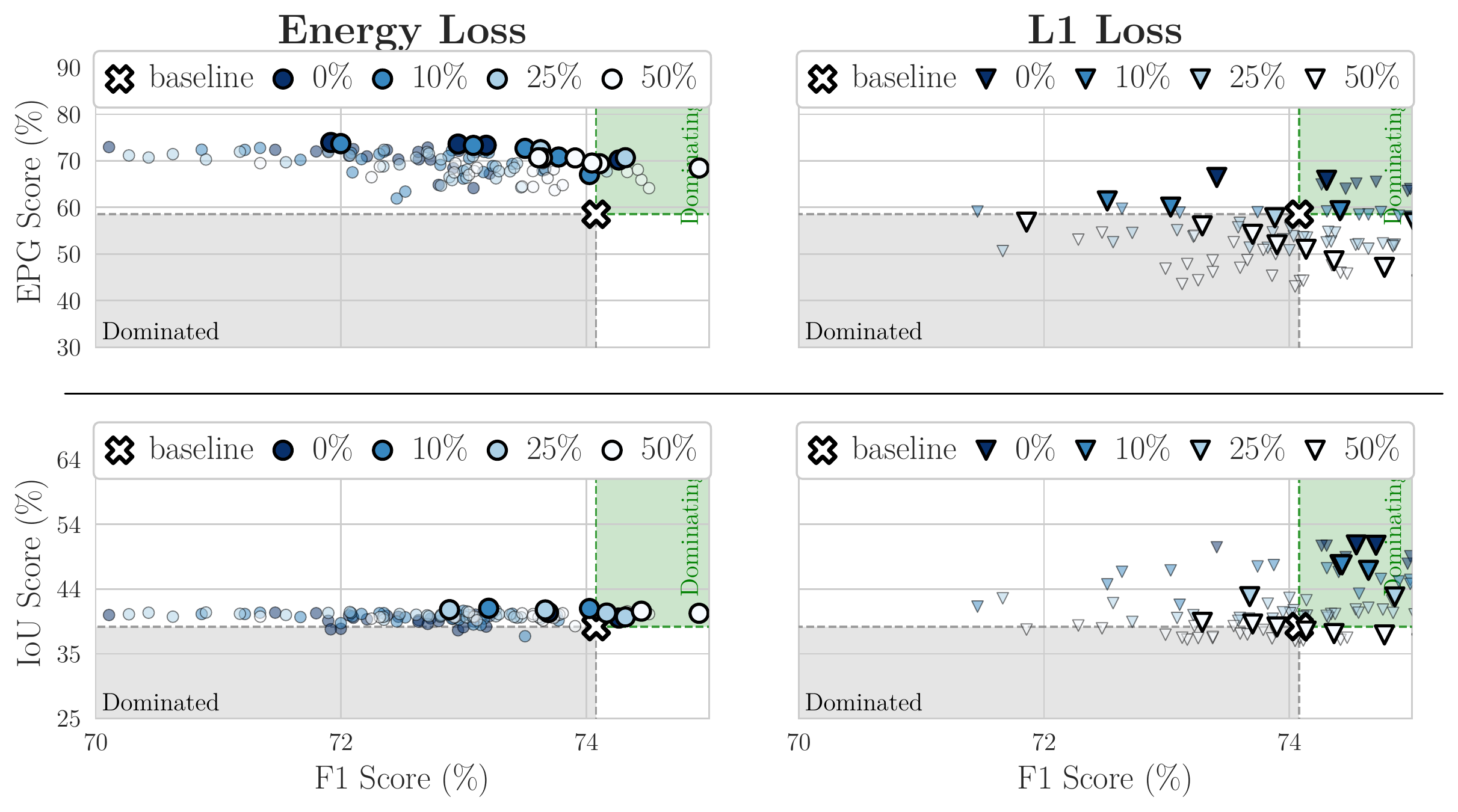}
    \caption{\textbf{\xdnn \resnet  @ Final.}}
    \end{subfigure}
    \caption{\textbf{Coarse bounding box results.} We show the impact of dilating bounding boxes during training for the \textbf{(a)} \vanilla and \textbf{(b)} \xdnn models. Similar to the results seen with \bcos models (\cref{fig:coarse_annotations}), we find that the \energyloss localization loss is generally robust to coarse annotations, while the effectiveness of guidance with the \loneloss localization loss worsens as the extent of coarseness (dilations) increases. Full results in \cref{fig:supp:dilation_quanti:full}.}
    \label{fig:supp:dilation_quanti}
\end{figure}

In \cref{fig:supp:dilation_quanti}, we additionally show the impact of training with coarse, dilated bounding boxes for \ixg attributions on the \vanilla model, and \intgrad attributions on the \xdnn model. Similar to the results seen with \bcos attributions (\cref{fig:coarse_annotations}), we find that the \energyloss localization loss is robust to coarse annotations, while the performance with \loneloss localization loss worsens as the dilations increase.

\subsection{Evaluation on DenseNet and ViT models}
\label{supp:sec:quantitative:morearchs}

\begin{figure}[t]
    \vspace{-.1em}
    \centering
    \begin{subfigure}[c]{.495\textwidth}
    \includegraphics[width=\linewidth]{results/VOC/figures/backbones/architectures_loc.pdf}
    \caption{}
    \label{fig:supp:moremodels:a}
    \end{subfigure}
    \begin{subfigure}[c]{.495\textwidth}
    \includegraphics[width=\linewidth]{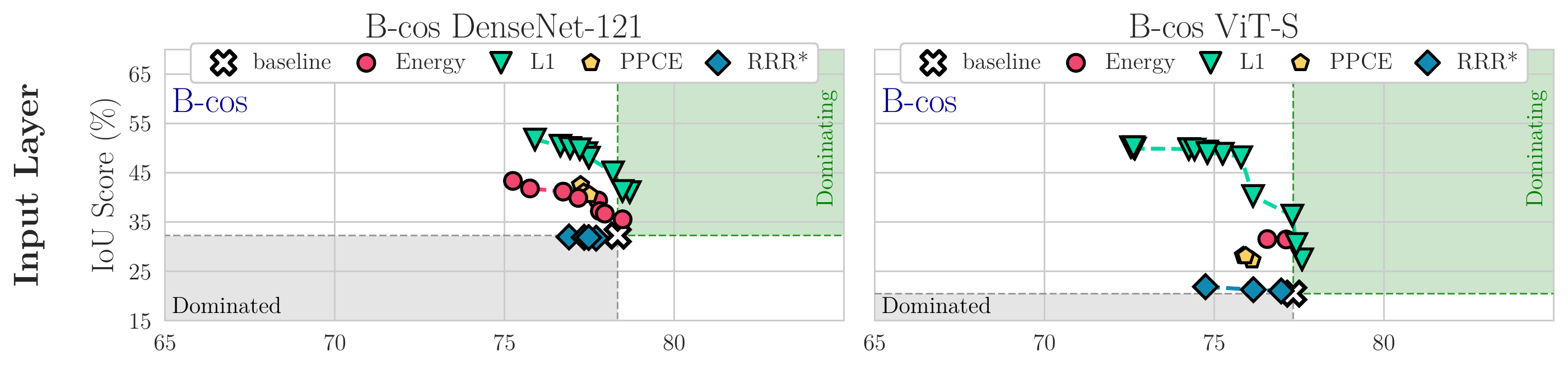}
    \caption{}
    \label{fig:supp:moremodels:b}
    \end{subfigure}
    \caption{\textbf{\epg and \iou vs.~\fone on VOC for two additional \bcos architectures.}
    We find that the trends observed in the main paper for a \bcos ResNet-50 backbone (cf.~\cref{fig:epg_results,fig:iou_results}, right columns) generalize to other backbone architectures. In particular, we find that the \loneloss loss yields the highest gains in \iou, whereas the \energyloss loss yields the highest gains in \epg, both for a DenseNet-121 and a ViT-S model.
    }
    \label{fig:supp:moremodels}
\end{figure}

In \cref{fig:supp:moremodels}, we evaluate the best performing configurations from our study, \ie performing guidance using B-cos attributions at input, on additional model backbones, specifically DenseNet-121 and ViT-S. We find that the trends observed with ResNet-50 models generalizes to these backbones, with the Energy loss yielding the highest gains for EPG, and the $L_1$ loss yielding the highest gains for IoU.

\clearpage
\clearpage
\section{Waterbirds Results}
\label{supp:sec:waterbirds}

\definecolor{lightgrey}{rgb}{.925, .925, .925}
\definecolor{grey}{rgb}{.4, .4, .4}

\setlength\tabcolsep{.125em}
\begin{table}[h!]
\centering
\begin{tabular}{=c+c+c@{\hskip3pt}@{\hskip3pt}+c+c+c+c+c@{\hskip3pt}|@{\hskip3pt}+c+c+c+c+c}
&&&
&\multicolumn{3}{c}{\footnotesize \textbf{Conventional Setting}}&&
&\multicolumn{3}{c}{\footnotesize \textbf{Reversed Setting}}&\\[.25em]
& \footnotesize \bf Layer & \footnotesize \bf Loss      &                             \footnotesize{\bf G1 Acc} &                             \footnotesize{\bf G2 Acc} &                             \footnotesize{\bf G3 Acc} &                             \footnotesize{\bf G4 Acc} &                            \footnotesize{\bf Overall} &                             \footnotesize{\bf G1 Acc} &                             \footnotesize{\bf G2 Acc} &                             \footnotesize{\bf G3 Acc} &                             \footnotesize{\bf G4 Acc} &                            \footnotesize{\bf Overall} \\[.5em]\hline

\multirow{4}{*}{\footnotesize \phantom{\scriptsize0}\rotatebox[origin=c]{90}{\textbf{\bcos}}}\phantom{\scriptsize0} & \phantom{\scriptsize 0}\multirow{2}{*}{\footnotesize Input}\phantom{\scriptsize 0} &\footnotesize Energy &            \footnotesize{ 99.2 \scriptsize($\pm$0.1) } &           \footnotesize{ 40.4 \scriptsize($\pm$1.0) } &  \textbf{\footnotesize{ 56.1 \scriptsize($\pm$4.0) }} &           \footnotesize{ 96.6 \scriptsize($\pm$0.4) } &  \textbf{\footnotesize{ 71.2 \scriptsize($\pm$0.1) }} &  \textbf{\footnotesize{ 99.4 \scriptsize($\pm$0.1) }} &  \textbf{\footnotesize{ 70.2 \scriptsize($\pm$2.1) }} &  \textbf{\footnotesize{ 62.8 \scriptsize($\pm$2.1) }} &           \footnotesize{ 96.5 \scriptsize($\pm$0.6) } &  \textbf{\footnotesize{ 83.6 \scriptsize($\pm$1.1) }} \\
&       & \footnotesize{L1} &           \footnotesize{ 99.3 \scriptsize($\pm$0.1) } &           \footnotesize{ 37.0 \scriptsize($\pm$0.8) } &           \footnotesize{ 51.1 \scriptsize($\pm$1.9) } &  \textbf{\footnotesize{ 97.2 \scriptsize($\pm$0.3) }} &           \footnotesize{ 69.5 \scriptsize($\pm$0.2) } &           \footnotesize{ 99.3 \scriptsize($\pm$0.3) } &           \footnotesize{ 67.7 \scriptsize($\pm$3.3) } &           \footnotesize{ 58.8 \scriptsize($\pm$5.0) } &  \textbf{\footnotesize{ 96.7 \scriptsize($\pm$0.7) }} &           \footnotesize{ 82.2 \scriptsize($\pm$0.9) } \\
       & \phantom{\scriptsize 0}\multirow{2}{*}{\footnotesize Final}\phantom{\scriptsize 0} &\footnotesize Energy &            \footnotesize{ 99.3 \scriptsize($\pm$0.1) } &  \textbf{\footnotesize{ 41.0 \scriptsize($\pm$2.1) }} &           \footnotesize{ 53.1 \scriptsize($\pm$0.8) } &           \footnotesize{ 96.3 \scriptsize($\pm$0.5) } &           \footnotesize{ 71.1 \scriptsize($\pm$0.9) } &  \textbf{\footnotesize{ 99.4 \scriptsize($\pm$0.2) }} &           \footnotesize{ 70.1 \scriptsize($\pm$3.1) } &           \footnotesize{ 60.2 \scriptsize($\pm$3.9) } &           \footnotesize{ 95.8 \scriptsize($\pm$1.1) } &           \footnotesize{ 83.2 \scriptsize($\pm$1.1) } \\
       &       & \footnotesize{L1} &           \footnotesize{ 99.3 \scriptsize($\pm$0.1) } &           \footnotesize{ 34.3 \scriptsize($\pm$3.2) } &           \footnotesize{ 49.4 \scriptsize($\pm$2.6) } &           \footnotesize{ 96.6 \scriptsize($\pm$0.6) } &           \footnotesize{ 68.2 \scriptsize($\pm$1.1) } &  \textbf{\footnotesize{ 99.4 \scriptsize($\pm$0.1) }} &           \footnotesize{ 69.8 \scriptsize($\pm$2.1) } &           \footnotesize{ 56.3 \scriptsize($\pm$1.8) } &           \footnotesize{ 96.1 \scriptsize($\pm$0.7) } &           \footnotesize{ 82.8 \scriptsize($\pm$0.8) } \\
       \rowcolor{lightgrey}\rowstyle{\color{gray}}&\multicolumn{2}{c}{\footnotesize \color{grey} Baseline} &  \textbf{\footnotesize{ 99.4 \scriptsize($\pm$0.1) }} &           \footnotesize{ 37.2 \scriptsize($\pm$0.2) } &           \footnotesize{ 43.4 \scriptsize($\pm$2.4) } &           \footnotesize{ 96.5 \scriptsize($\pm$0.1) } &           \footnotesize{ 68.7 \scriptsize($\pm$0.2) } &  \textbf{\footnotesize{ 99.4 \scriptsize($\pm$0.1) }} &           \footnotesize{ 62.8 \scriptsize($\pm$0.2) } &           \footnotesize{ 56.6 \scriptsize($\pm$2.4) } &           \footnotesize{ 96.5 \scriptsize($\pm$0.1) } &           \footnotesize{ 80.1 \scriptsize($\pm$0.2) } \\
\hline

\multirow{4}{*}{\footnotesize \phantom{\scriptsize0}\rotatebox[origin=c]{90}{\textbf{\xdnn}}}\phantom{\scriptsize0} & \phantom{\scriptsize 0}\multirow{2}{*}{\footnotesize Input}\phantom{\scriptsize 0} &\footnotesize Energy &           \footnotesize{ 99.3 \scriptsize($\pm$0.2) } &  \textbf{\footnotesize{ 47.0 \scriptsize($\pm$9.1) }} &           \footnotesize{ 49.2 \scriptsize($\pm$4.8) } &  \textbf{\footnotesize{ 96.8 \scriptsize($\pm$0.7) }} &  \textbf{\footnotesize{ 73.1 \scriptsize($\pm$3.4) }} &           \footnotesize{ 99.0 \scriptsize($\pm$0.3) } &  \textbf{\footnotesize{ 67.6 \scriptsize($\pm$4.8) }} &  \textbf{\footnotesize{ 63.9 \scriptsize($\pm$3.6) }} &           \footnotesize{ 96.1 \scriptsize($\pm$0.7) } &  \textbf{\footnotesize{ 82.6 \scriptsize($\pm$2.0) }} \\
&       & \footnotesize{L1} &           \footnotesize{ 99.1 \scriptsize($\pm$0.6) } &           \footnotesize{ 40.4 \scriptsize($\pm$7.3) } &           \footnotesize{ 41.8 \scriptsize($\pm$3.8) } &           \footnotesize{ 96.5 \scriptsize($\pm$0.6) } &           \footnotesize{ 69.6 \scriptsize($\pm$3.2) } &  \textbf{\footnotesize{ 99.3 \scriptsize($\pm$0.2) }} &           \footnotesize{ 59.1 \scriptsize($\pm$4.7) } &           \footnotesize{ 63.6 \scriptsize($\pm$6.1) } &           \footnotesize{ 96.0 \scriptsize($\pm$0.9) } &           \footnotesize{ 79.3 \scriptsize($\pm$1.3) } \\
       & \phantom{\scriptsize 0}\multirow{2}{*}{\footnotesize Final}\phantom{\scriptsize 0} &\footnotesize Energy &           \footnotesize{ 99.2 \scriptsize($\pm$0.4) } &          \footnotesize{ 42.5 \scriptsize($\pm$10.4) } &  \textbf{\footnotesize{ 54.2 \scriptsize($\pm$3.2) }} &           \footnotesize{ 96.6 \scriptsize($\pm$0.9) } &           \footnotesize{ 71.9 \scriptsize($\pm$4.2) } &           \footnotesize{ 99.2 \scriptsize($\pm$0.2) } &           \footnotesize{ 65.3 \scriptsize($\pm$2.0) } &           \footnotesize{ 62.3 \scriptsize($\pm$3.3) } &           \footnotesize{ 96.0 \scriptsize($\pm$0.5) } &           \footnotesize{ 81.5 \scriptsize($\pm$0.9) } \\
       &       & \footnotesize{L1} &  \textbf{\footnotesize{ 99.4 \scriptsize($\pm$0.1) }} &           \footnotesize{ 45.1 \scriptsize($\pm$4.0) } &           \footnotesize{ 42.8 \scriptsize($\pm$2.8) } &           \footnotesize{ 96.5 \scriptsize($\pm$0.5) } &           \footnotesize{ 71.7 \scriptsize($\pm$1.4) } &  \textbf{\footnotesize{ 99.3 \scriptsize($\pm$0.2) }} &           \footnotesize{ 62.9 \scriptsize($\pm$4.8) } &           \footnotesize{ 59.8 \scriptsize($\pm$4.8) } &           \footnotesize{ 95.8 \scriptsize($\pm$0.7) } &           \footnotesize{ 80.4 \scriptsize($\pm$1.8) } \\
       \rowcolor{lightgrey}\rowstyle{\color{gray}}&\multicolumn{2}{c}{\footnotesize \color{grey} Baseline} &           \footnotesize{ 99.3 \scriptsize($\pm$0.1) } &           \footnotesize{ 39.8 \scriptsize($\pm$0.7) } &           \footnotesize{ 38.6 \scriptsize($\pm$2.5) } &           \footnotesize{ 96.3 \scriptsize($\pm$0.7) } &           \footnotesize{ 69.1 \scriptsize($\pm$0.6) } &           \footnotesize{ \textbf{99.3 \scriptsize($\pm$0.1)} } &           \footnotesize{ 60.2 \scriptsize($\pm$0.7) } &           \footnotesize{ 61.4 \scriptsize($\pm$2.5) } &           \footnotesize{ \textbf{96.3 \scriptsize($\pm$0.7)} } &           \footnotesize{ 79.6 \scriptsize($\pm$0.5) } \\
\hline

\multirow{4}{*}{\footnotesize \phantom{\scriptsize0}\rotatebox[origin=c]{90}{\textbf{\vanilla}}}\phantom{\scriptsize0} & \phantom{\scriptsize 0}\multirow{2}{*}{\footnotesize Input}\phantom{\scriptsize 0} &\footnotesize Energy &          \footnotesize{ 99.4 \scriptsize($\pm$0.2) } &           \footnotesize{ 42.4 \scriptsize($\pm$2.6) } &           \footnotesize{ 47.9 \scriptsize($\pm$3.5) } &           \footnotesize{ 97.1 \scriptsize($\pm$0.4) } &           \footnotesize{ 71.2 \scriptsize($\pm$1.0) } &  \textbf{\footnotesize{ 99.6 \scriptsize($\pm$0.2) }} &           \footnotesize{ 50.7 \scriptsize($\pm$7.3) } &           \footnotesize{ 52.4 \scriptsize($\pm$1.7) } &           \footnotesize{ 97.2 \scriptsize($\pm$0.5) } &           \footnotesize{ 75.1 \scriptsize($\pm$2.9) } \\
&       & \footnotesize{L1} &  \textbf{\footnotesize{ 99.5 \scriptsize($\pm$0.1) }} &           \footnotesize{ 46.1 \scriptsize($\pm$4.4) } &           \footnotesize{ 51.1 \scriptsize($\pm$4.0) } &           \footnotesize{ 97.5 \scriptsize($\pm$0.1) } &           \footnotesize{ 73.1 \scriptsize($\pm$1.6) } &  \textbf{\footnotesize{ 99.6 \scriptsize($\pm$0.1) }} &           \footnotesize{ 48.0 \scriptsize($\pm$7.8) } &           \footnotesize{ 49.7 \scriptsize($\pm$3.7) } &           \footnotesize{ 96.8 \scriptsize($\pm$0.6) } &           \footnotesize{ 73.7 \scriptsize($\pm$2.7) } \\
       & \phantom{\scriptsize 0}\multirow{2}{*}{\footnotesize Final}\phantom{\scriptsize 0} &\footnotesize Energy & \textbf{\footnotesize{ 99.5 \scriptsize($\pm$0.0) }} &           \footnotesize{ 56.1 \scriptsize($\pm$7.0) } &  \textbf{\footnotesize{ 60.7 \scriptsize($\pm$5.5) }} &           \footnotesize{ 97.0 \scriptsize($\pm$0.5) } &  \textbf{\footnotesize{ 78.1 \scriptsize($\pm$2.6) }} &           \footnotesize{ 99.5 \scriptsize($\pm$0.1) } &           \footnotesize{ 59.4 \scriptsize($\pm$5.9) } &  \textbf{\footnotesize{ 56.5 \scriptsize($\pm$3.7) }} &           \footnotesize{ 97.2 \scriptsize($\pm$0.5) } &  \textbf{\footnotesize{ 78.9 \scriptsize($\pm$1.9) }} \\
       &       & \footnotesize{L1} &  \textbf{\footnotesize{ 99.5 \scriptsize($\pm$0.1) }} &  \textbf{\footnotesize{ 57.1 \scriptsize($\pm$2.9) }} &           \footnotesize{ 55.4 \scriptsize($\pm$2.5) } &           \footnotesize{ 96.7 \scriptsize($\pm$0.6) } &           \footnotesize{ 77.8 \scriptsize($\pm$1.0) } &           \footnotesize{ 99.5 \scriptsize($\pm$0.1) } &           \footnotesize{ 56.3 \scriptsize($\pm$6.7) } &           \footnotesize{ 51.6 \scriptsize($\pm$3.1) } &           \footnotesize{ 97.3 \scriptsize($\pm$0.6) } &           \footnotesize{ 77.1 \scriptsize($\pm$2.5) } \\
       \rowcolor{lightgrey}\rowstyle{\color{gray}}&\multicolumn{2}{c}{\footnotesize \color{grey} Baseline} &           \footnotesize{ 99.4 \scriptsize($\pm$0.0) } &           \footnotesize{ 39.6 \scriptsize($\pm$0.7) } &           \footnotesize{ 53.7 \scriptsize($\pm$2.1) } &  \textbf{\footnotesize{ 97.7 \scriptsize($\pm$0.0) }} &           \footnotesize{ 70.8 \scriptsize($\pm$0.0) } &           \footnotesize{ 99.4 \scriptsize($\pm$0.0) } &  \textbf{\footnotesize{ 60.4 \scriptsize($\pm$0.7) }} &           \footnotesize{ 46.3 \scriptsize($\pm$2.1) } &  \textbf{\footnotesize{ 97.7 \scriptsize($\pm$0.0) }} &           \footnotesize{ 78.1 \scriptsize($\pm$0.1) } \\
       \hline
\end{tabular}
\caption{\textbf{Classification performance on Waterbirds} after model guidance with the \lone and the \epgloss loss. We find that both losses consistently improve the models' classification performance over the baseline model (\ie a model without guidance). These improvements are particularly pronounced in the groups \emph{not seen during training}, \ie landbirds on water (``G2'') and waterbirds on land (``G3''). For qualitative visualizations of the effect of model guidance on the waterbirds dataset, see \cref{fig:supp:waterbirds_quali}.}
\label{tab:supp:waterbirds_table}
\end{table}

As discussed in section \cref{sec:results:waterbirds}, we use the \waterbirds dataset \citeApp{sagawa2019distributionally,petryk2022guiding} to evaluate the effectiveness of model guidance in a setting where strong spurious correlations are present in the training data. This dataset consists of four groups---\textit{Landbird} on \textit{Land} (\textbf{G1}), \textit{Landbird} on \textit{Water} (\textbf{G2}), \textit{Waterbird} on \textit{Land} (\textbf{G3}), and \textit{Waterbird} on \textit{Water} (\textbf{G4})---of which only groups \textbf{G1} and \textbf{G4} appear during training and the background is thus perfectly correlated with the type of bird (\eg Landbird on land).

To evaluate the effectiveness of model guidance, we train the models on two binary classification tasks: to classify the type of birds (the \emph{conventional setting}) or the background (the \emph{reversed setting}, as described in \citeApp{petryk2022guiding}) and evaluate models without guidance (baselines), as well as with guidance: specifically, for guiding the models, we evaluate different models (\vanilla, \xdnn, \bcos) with different guidance losses (\epgloss, \lone) applied at different layers (Input and Final), see \cref{tab:supp:waterbirds_table}. {For each model, we use its corresponding attribution method, \ie \ixg for \vanilla, \intgrad for \xdnn, and \bcos for \bcos.}

In \cref{tab:supp:waterbirds_table} we present the classification performance for the individual groups (\textbf{G1-G4}) as well as the average over all samples  (`Overall') across all configurations; note that the group sizes differ in the test set and the average over the individual group acccuracies thus differs from the overall accuracy. For each row, the results are averaged over 4 runs (2 random training seeds and 2 different sets of 1\% annotated samples) with the exception of the baseline results being an average over 2 runs.

In almost all cases, we find that both of the evaluated losses (\epgloss, \lone) improve the models' classification performance over the baseline. As expected, these improvements are particularly pronounced in the groups not seen during training, \ie landbirds on water (\textbf{G2}) and waterbirds on land (\textbf{G3}). 

\begin{figure}[h]
    \centering
    {\textbf{Additional qualitative results} on the \waterbirds dataset.}\vspace{.25cm}\\
    \begin{subfigure}[c]{.46\textwidth}
    \includegraphics[width=\textwidth]{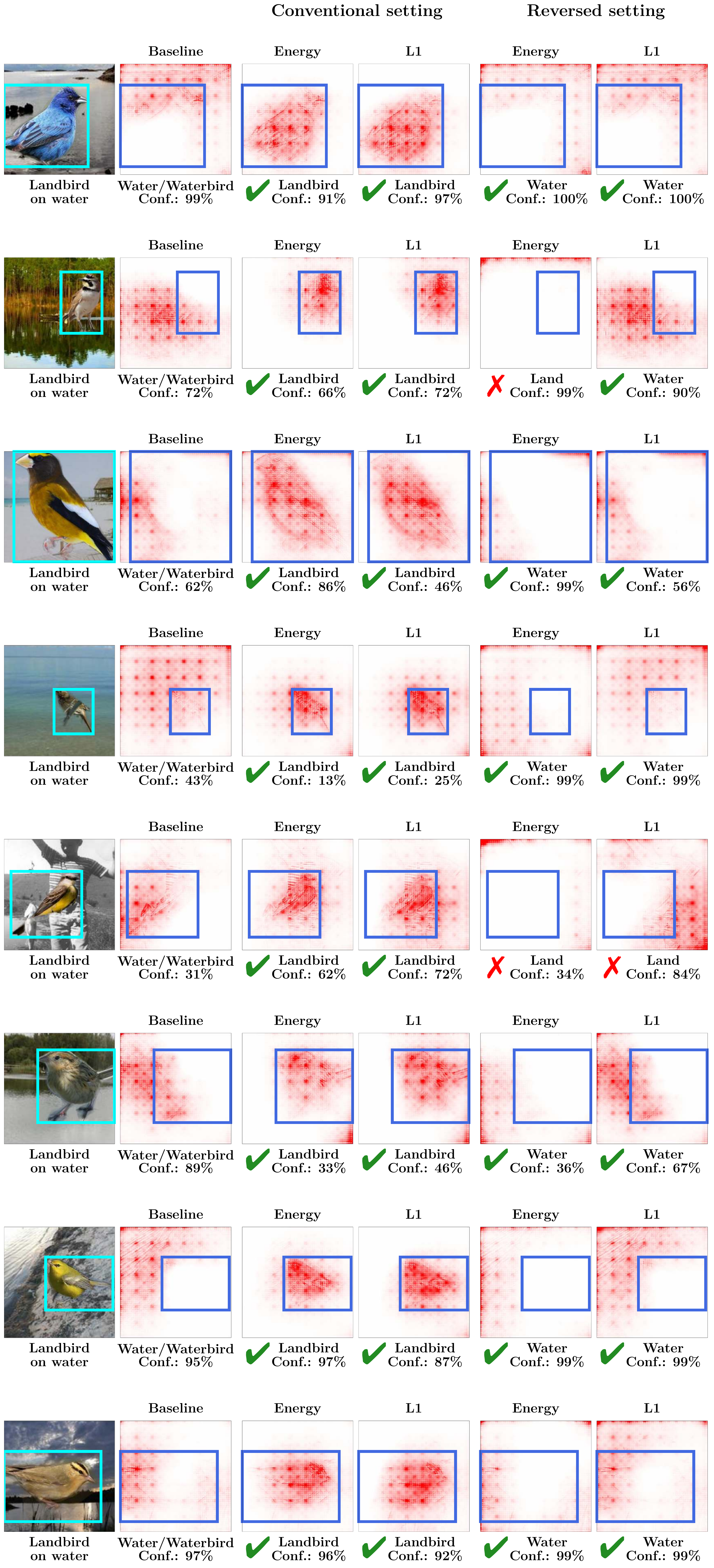}
    \caption{\textbf{Landbirds on Water.}}
    \label{fig:supp:waterbirds_quali:a}
    \end{subfigure}\hfill
    \begin{subfigure}[c]{.46\textwidth}
    \includegraphics[width=\textwidth]{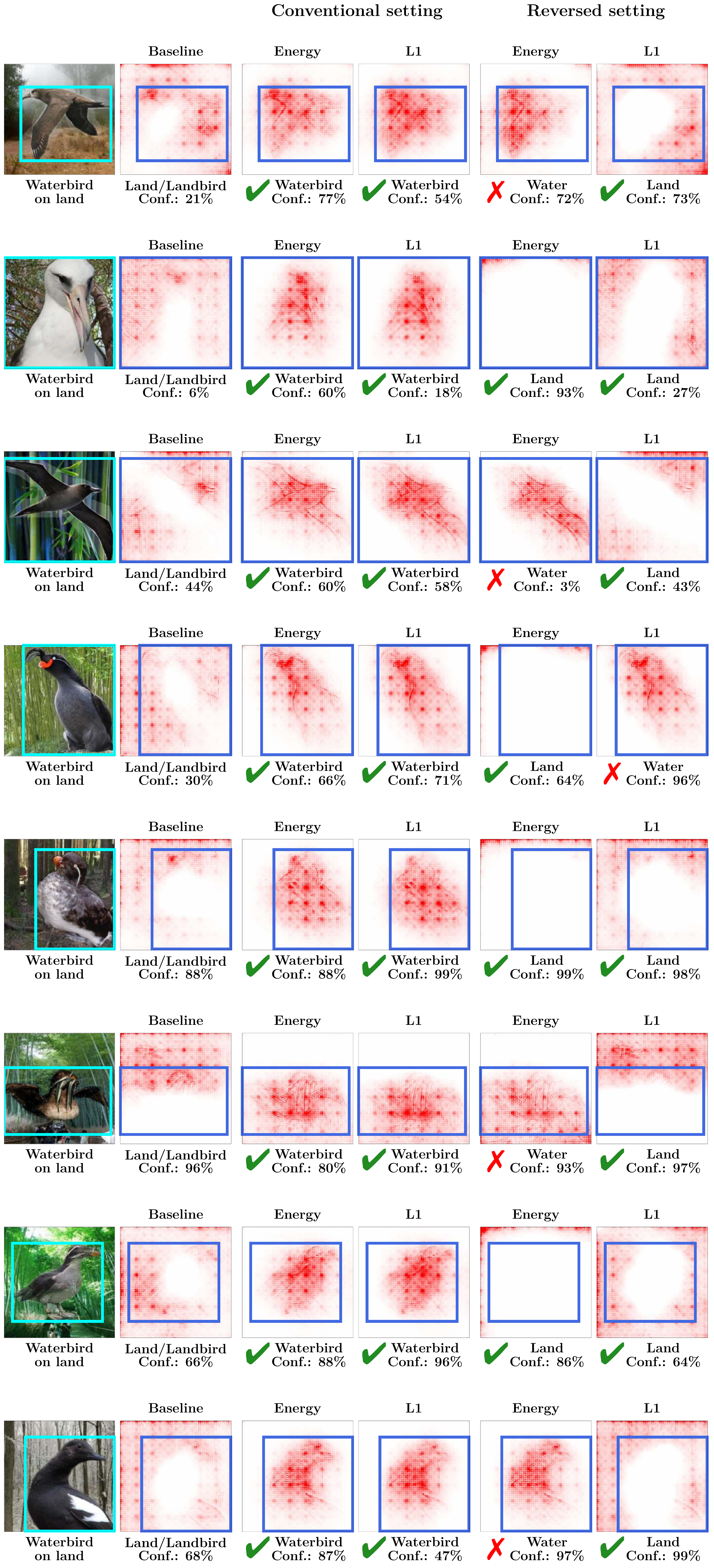}
    \caption{\textbf{Waterbirds on Land.}}
    \label{fig:supp:waterbirds_quali:b}
    \end{subfigure}
    \caption{\textbf{Qualitative results for the Waterbirds dataset.} Specifically, we show input layer attributions for \bcos models trained without guidance (`Baseline') as well as guided via the \epgloss or \lone loss. We find that model guidance can be effective both for focusing on the bird and the background. For example, in the top row of (a), the model originally focuses on the background (col.~2) and classifies the image (col.~1) as Water/Waterbird. In the conventional setting, both the \energyloss and \loneloss localization losses are effective in redirecting the focus to the bird (cols.~3-4), changing the model's prediction to Landbird with high confidence. Similarly, in the reversed setting, both localization losses direct the focus to the background (cols.~5-6), which increases the model's confidence in classifying the image as Water.}
    \label{fig:supp:waterbirds_quali}
\end{figure}

Further, in \cref{fig:supp:waterbirds_quali}, we show attribution maps of the baseline models, as well as the guided models. As can be seen, model guidance not only improves the accuracy, but is also reflected in the attribution maps: \eg, in row 1 of \cref{fig:supp:waterbirds_quali:a}, we see that while the baseline model originally focused on the background (water) to classify the image, it is possible to guide the model to use the desired features (\ie the bird in conventional setting and the background in the reversed setting) and consequently arrive at the desired classification decision. As this guidance is `soft', we also observe cases in which the model still focused on the wrong feature and thus arrived at the wrong prediction: \eg in \cref{fig:supp:waterbirds_quali:b} row 1 (reversed setting), the \epgloss-guided model still focuses on the bird and thus incorrectly predicts `Water', similar to the \lone-guided model in row 4. 

\clearpage
\section{Implementation Details}
\label{supp:sec:implementation}

\subsection{Training and Evaluation Details}
\label{supp:sec:implementation:training}

\myparagraph{Implementations:} We implement our code using PyTorch\footnote{\url{https://github.com/pytorch/pytorch}} \citeApp{paszke2019pytorch}. The \voc \citeApp{everingham2009pascal} and \coco \citeApp{lin2014microsoft} datasets and the \vanilla \resnet model were obtained from the Torchvision library\footnote{\url{https://github.com/pytorch/vision}} \citeS{paszke2019pytorch_2,torchvision2016_2}. Official implementations were used for the \bcos\footnote{\label{footnote:bcoscode}\url{https://github.com/B-cos/B-cos-v2}} \citeApp{bohle2023b} and \xdnn\footnote{\url{https://github.com/visinf/fast-axiomatic-attribution}} \citeApp{hesse2021fast} networks. Some of the utilities for data loading and evaluation were derived from NN-Explainer\footnote{\url{https://github.com/stevenstalder/NN-Explainer}} \citeApp{stalder2022wyswyc}, and for visualization from the Captum library\footnote{\url{https://github.com/pytorch/captum}} \citeApp{kokhlikyan2020captum}.

\subsubsection{Experiments with \vocs and \cocos}

\myparagraph{Training baseline models:} We train starting from models pre-trained on \imagenet \citeApp{imagenet}. We fine-tune with fixed learning rates in $\{10^{-3},10^{-4},10^{-5}\}$ using an Adam optimizer \citeApp{kingma2014adam} and select the checkpoint with the best validation F1-score. For \vocs, we train for 300 epochs, and for \cocos, we train for 60 epochs.

\myparagraph{Training guided models:} We train the models jointly optimized for classification and localization (\cref{eq:overall}) by fine-tuning the baseline models. We use a fixed learning rate of $10^{-4}$ and a batch size of $64$. For each configuration (given by a combination of attribution method, localization loss, and layer), we train using three different values of $\lambda_{\text{loc}}$, as detailed in \cref{tab:supp:lambdas}. For \vocs, we train for 50 epochs, and for \cocos, we train for 10 epochs.

\myparagraph{Selecting models to visualize:} As described in \cref{sec:experiments}, we select and evaluate on the set of Pareto-dominant models for each configuration after training. Each model on the Pareto front represents the extent of trade-off made between classification (F1) and localization (\epg) performance. In practice, the `best' model to choose would depend on the requirements of the end user. However, to evaluate the effectiveness of model guidance (\eg \cref{fig:teaser,fig:teaser2,fig:loss_comp}), we select a representative model on the front whose attributions we visualize. This is done by selecting the model with the highest \epg score with an at most 5 p.p. drop in F1-score.

\myparagraph{Efficient Optimization:} As described in \cref{sec:method:efficient}, for each image in a batch, we optimize for localization of a single class selected at random. This approximation allows us to perform model guidance efficiently and keeps the training cost tractable. However, to accurately evaluate the impact of this optimization, we evaluate the localization of all classes in the image at test time.

\myparagraph{Training with Limited Annotations:} As described in \cref{sec:results:ablations}, we show that training with a limited number of annotations can be a cost effective way of performing model guidance. In order to maintain the relative magnitude of $\mathcal{L}_{\text{loc}}$ as compared to $\mathcal{L}_{\text{class}}$ in this setting, we scale up the values of $\lambda_{\text{loc}}$ when training. The values of $\lambda_{\text{loc}}$ we use are shown in \cref{tab:supp:dilation:lambdas}.

\subsubsection{Experiments with \waterbirds}

\myparagraph{Data distributions:} The conventional binary classification task includes classifying \textit{Landbird} from \textit{Waterbird}, irrespective of their backgrounds.  We use the same splits generated and published by \citeApp{petryk2022guiding}. As discussed in \cref{supp:sec:waterbirds}, at training time there are no samples from \textbf{G2} or \textbf{G3}, making the bird type and the background perfectly correlated.  In contrast, both the validation and test sets are balanced across foregrounds and backgrounds, \ie a landbird is equally likely to occur on land or water, and vice-versa. However, as noted by \citeApp{sagawa2019distributionally}, using a validation set with the same distribution as the test set leaks information on the test distribution in the process of hyperparameter and checkpoint selection during training. Therefore, we modify the validation split to avoid such information leakage; in particular, we use a validation set with the same distribution as the training set, and only use examples of groups \textbf{G1} and \textbf{G4}. Note that \cref{tab:waterbirds} refers to \textbf{G3} as the ``Worst Group''. \\

\myparagraph{Training details:} We train starting from models pre-trained on \imagenet \citeApp{imagenet}. We fine-tune with fixed learning rate of $10^{-5}$ with $\lambda_{\text{loc}}$ of $5\times10^{-2}$ ($5\times10^{-4} \times 100$ for using 1\% of annotations) using an Adam optimizer \citeApp{kingma2014adam} . We train for 350 epochs with random cropping and horizontal flipping and select the checkpoint with the highest accuracy on the modified validation set.

\begin{table}[t]
    \def\arraystretch{1.2}
    \centering
    \begin{tabular}{c@{\hskip10pt}|@{\hskip10pt}c}
	    \bf Localization Loss & \bf Values of $\lambda_{\text{loc}}$ \\
     \hline
        \energyloss & \footnotesize$5\mytimes10^{-4}$, \phantom{0} $1\mytimes10^{-3}$, \phantom{0} $5\mytimes10^{-3}$ \\
        \loneloss &\footnotesize $1\mytimes10^{-3}$, \phantom{0} $5\mytimes10^{-3}$, \phantom{0} $1\mytimes10^{-2}$ \\
        \ppceloss &\footnotesize $1\mytimes10^{-4}$, \phantom{0} $5\mytimes10^{-4}$, \phantom{0} $1\mytimes10^{-3}$ \\
        \rrrloss &\footnotesize $5\mytimes10^{-6}$, \phantom{0} $1\mytimes10^{-5}$, \phantom{0} $5\mytimes10^{-5}$  \\
	\end{tabular}
	\caption{\textbf{Hyperparameter $\lambda_{\text{loc}}$: Default training.} used for when training on \vocs and \cocos with each localization loss. Different values are used for different loss functions since the magnitudes of each loss varies.}
    \label{tab:supp:lambdas}
\end{table}

\begin{table}[t]
    \def\arraystretch{1.2}
    \centering
    \begin{tabular}{c@{\hskip10pt}|@{\hskip10pt}c}
	    \bf Localization Loss & \bf Values of $\lambda_{\text{loc}}$ \\
     \hline
        \energyloss & \footnotesize$0.05$, \phantom{0} $0.100$, \phantom{0} $0.50$ \\
        \loneloss &\footnotesize $0.01$, \phantom{0} $0.100$, \phantom{0} $1.00$ \\
	\end{tabular}
	\caption{\textbf{Hyperparameter $\lambda_{\text{loc}}$: Limited annotations.} used for when training on \vocs and \cocos with \textbf{limited data} for each localization loss. Different values are used for different loss functions since the magnitudes of each loss varies. We use larger values of $\lambda_{\text{loc}}$ when training with limited annotations to maintain the relative magnitudes of the classification and localization losses during training.}
    \label{tab:supp:dilation:lambdas}
\end{table}

\subsection{Optimizing \bcos Attributions}
\label{supp:sec:implementation:bcos}

Training for optimizing the localization of attributions (\cref{eq:overall}) requires backpropagating through the attribution maps, which implies that they need to be differentiable. While \bcos attributions \citeApp{bohle2022b} as formulated are mathematically differentiable, the original implementation\footnoteref{footnote:bcoscode} \citeApp{bohle2023b} for computing them involves detaching the dynamic weights from the computational graph, which prevents them from being used for optimization. In this work, to use them for model guidance, we develop a twice-differentiable implementation of \bcos attributions.

\clearpage

\setcounter{section}{23}
\section{Full Results}
\label{supp:sec:full}
\begin{figure}[h]
    \centering
    {Full results on \textbf{\voc} (\fone score).}\vspace{.25cm}\\
    \begin{subfigure}[c]{.9\textwidth}
    \includegraphics[width=\textwidth]{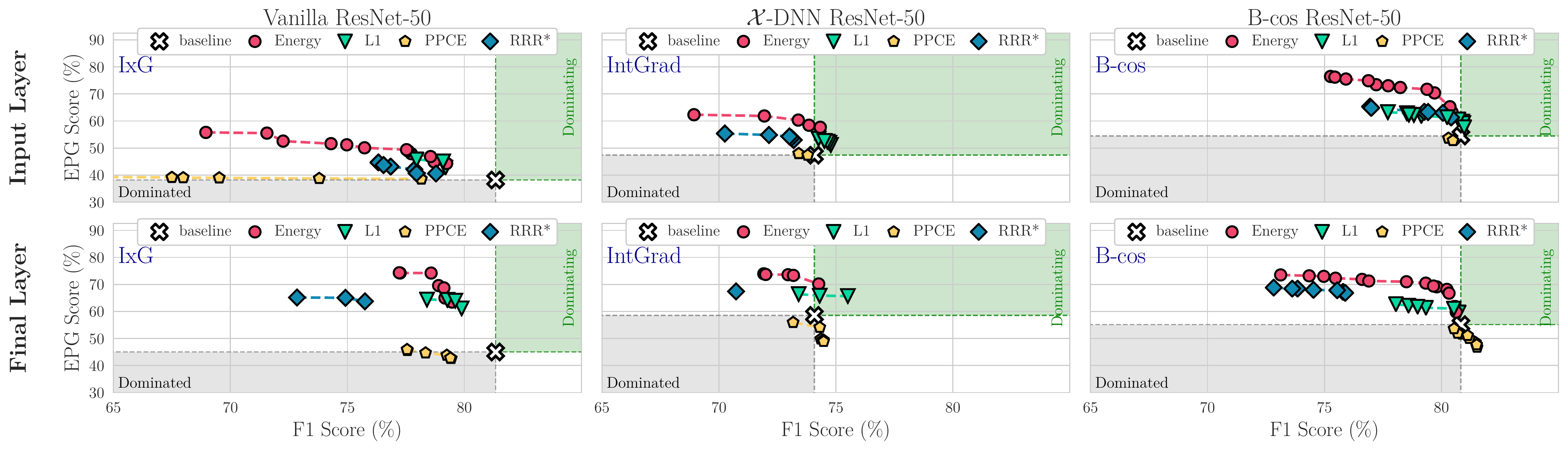}
    \caption{\textbf{\epg vs.~\fone.}}
    \end{subfigure}
    \begin{subfigure}[c]{.9\textwidth}
    \includegraphics[width=\textwidth]{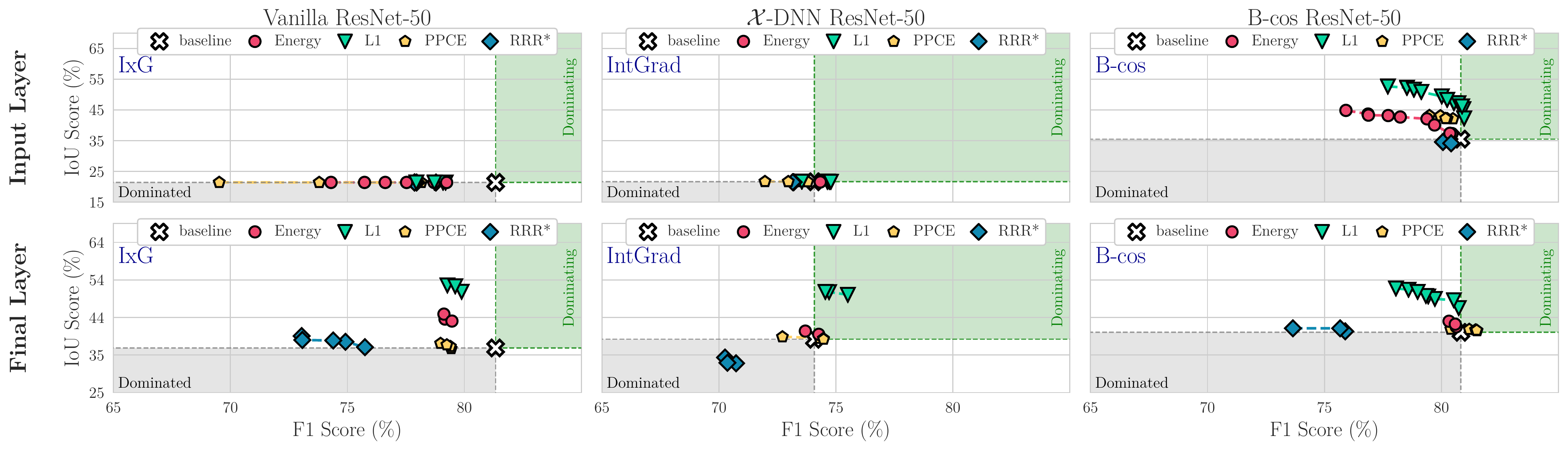}
    \caption{\textbf{\iou vs.~\fone.}}
    \end{subfigure}
    \caption{\textbf{EPG (a) and \iou (b) vs.~\fone on \vocs,} for different losses (\textbf{markers}) and models (\textbf{columns}), optimized at different layers (\textbf{rows}); additionally, we show the performance of the baseline model before fine-tuning and demarcate regions that strictly dominate (are strictly dominated by) the baseline performance in green (grey). 
    For each configuration, we show the Pareto fronts (cf.\ \cref{fig:pareto_example}) across regularization strengths $\lambda_\text{loc}$ and epochs (cf.\ \cref{sec:results} and \cref{fig:pareto_example}). 
    We find the \epgloss loss to give the best trade-off between \epg and \fone, whereas the \lone loss (especially at the final layer) provides the best trade-off between \iou and \fone. We further find these results to be consistent across datasets, see \cref{fig:supp:coco:f1_results}.}
    \label{fig:supp:voc:f1_results}
\end{figure}

\begin{figure}[h]
    \centering
    {Full results on \textbf{\coco} (\fone score).}\vspace{.25cm}\\
    \begin{subfigure}[c]{.9\textwidth}
    \includegraphics[width=\textwidth]{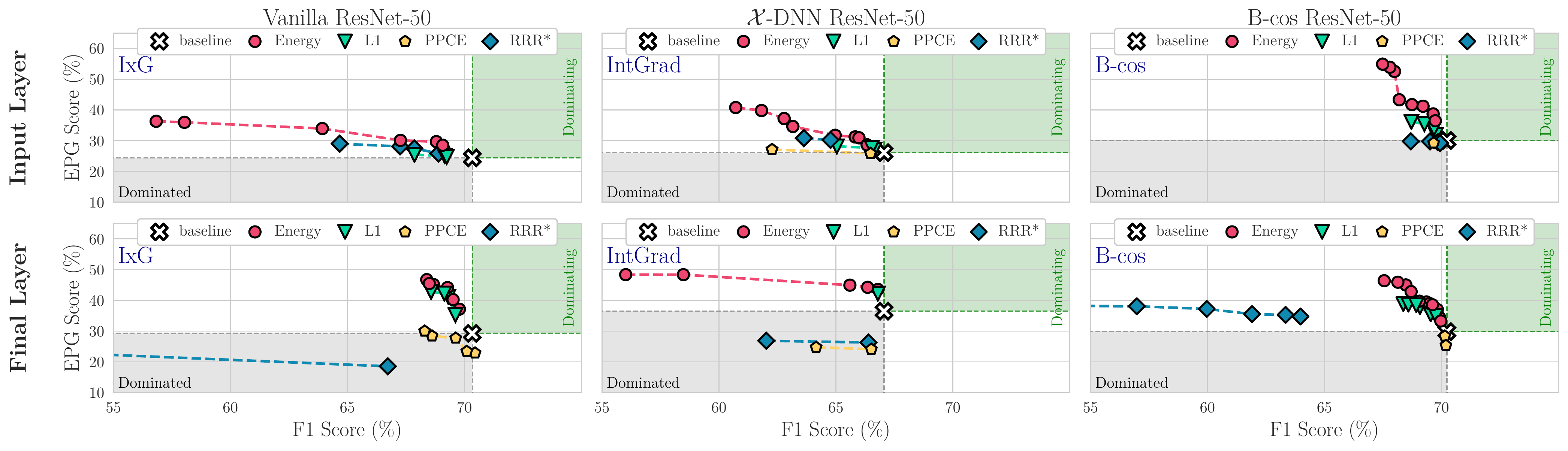}
    \caption{\textbf{\epg vs.~\fone.}}
    \end{subfigure}
    \begin{subfigure}[c]{.9\textwidth}
    \includegraphics[width=\textwidth]{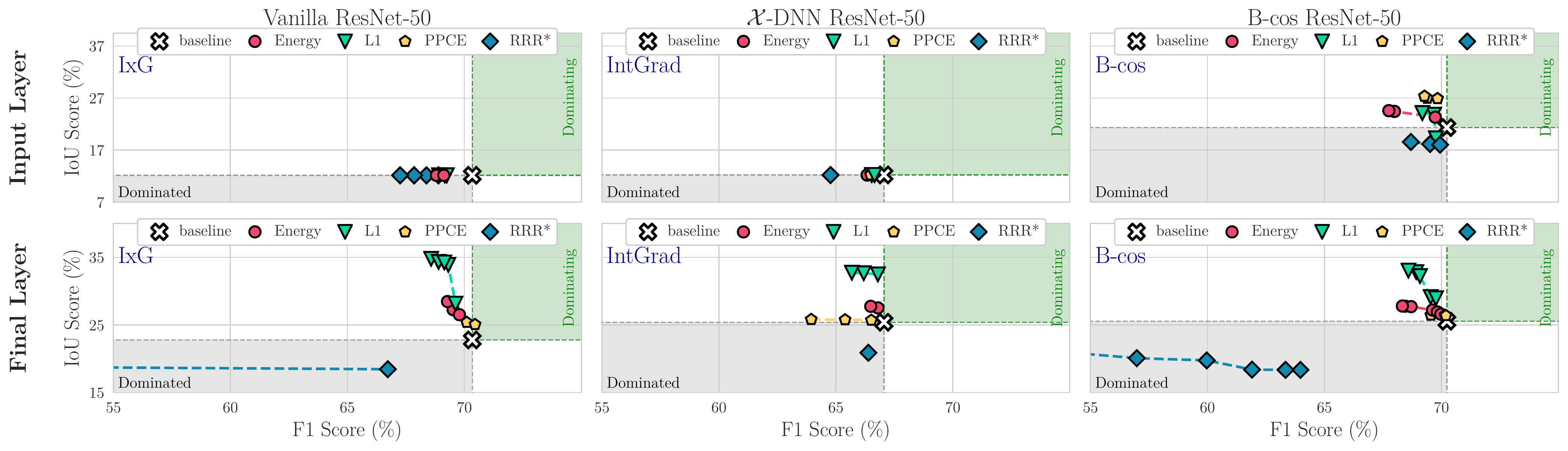}
    \caption{\textbf{\iou vs.~\fone.}}
    \end{subfigure}
    \caption{\textbf{EPG (a) and \iou (b) vs.~\fone on \cocos,} for different losses (\textbf{markers}) and models (\textbf{columns}), optimized at different layers (\textbf{rows}); additionally, we show the performance of the baseline model before fine-tuning and demarcate regions that strictly dominate (are strictly dominated by) the baseline performance in green (grey). 
    For each configuration, we show the Pareto fronts (cf.\ \cref{fig:pareto_example}) across regularization strengths $\lambda_\text{loc}$ and epochs (cf.\ \cref{sec:results} and \cref{fig:pareto_example}). 
    We find the \epgloss loss to give the best trade-off between \epg and \fone, whereas the \lone loss (especially at the final layer) provides the best trade-off between \iou and \fone. We further find these results to be consistent across datasets, see \cref{fig:supp:voc:f1_results}.}
    \label{fig:supp:coco:f1_results}
\end{figure}
\begin{figure}
    \centering
    \textbf{Mean Average Precision (\map) results} on \vocs.\vspace{.25cm}\\
    \begin{subfigure}[c]{.9\textwidth}
    \includegraphics[width=\textwidth]{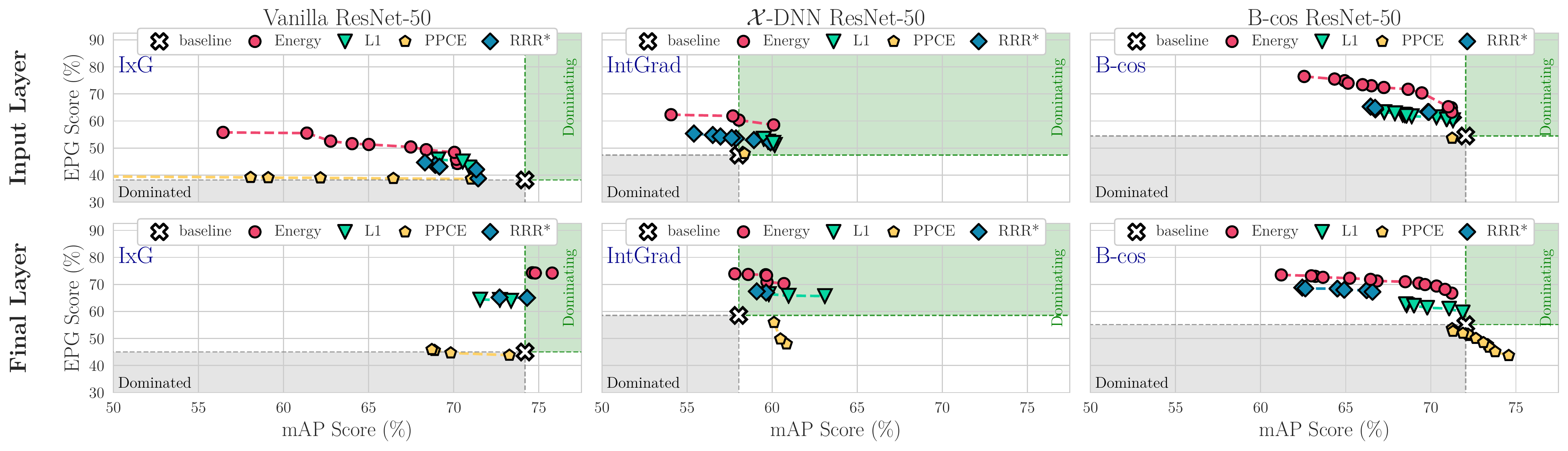}
    \caption{\textbf{\epg vs.~\map.}}
    \end{subfigure}
    \begin{subfigure}[c]{.9\textwidth}
    \includegraphics[width=\textwidth]{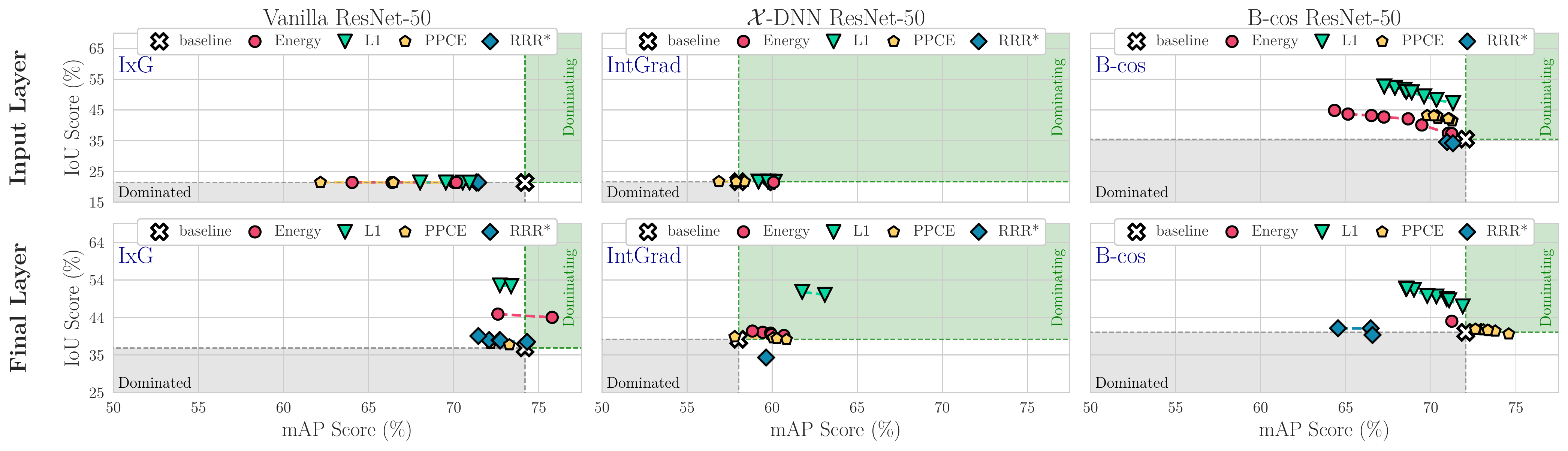}
    \caption{\textbf{\iou vs.~\map.}}
    \end{subfigure}
    \caption{\textbf{Quantitative comparison of \epg and \iou  vs.~\map scores for \vocs.} To ensure that the trends observed and described in the main paper generalize beyond the \fone metric, in this figure we show the \epg and \iou scores plotted against the \map metric. In general, we find the results obtained for the \map metric to be highly consistent with the previously shown results for the \fone metric, see \cref{fig:supp:voc:f1_results}. \Eg, across all configurations, we find the \epgloss to yield the highest gains in \epg score, whereas the \lone loss provides the best trade-offs with respect to the \iou metric. These results are further also consistent with those observed on \cocos, see \cref{fig:supp:coco:map_results}.}
    \label{fig:supp:voc:map_results:2}
\end{figure}
\begin{figure}[h]
    \centering
    {Full results on \textbf{\coco} (\map).}\vspace{.25cm}\\
    \begin{subfigure}[c]{.9\textwidth}
    \includegraphics[width=\textwidth]{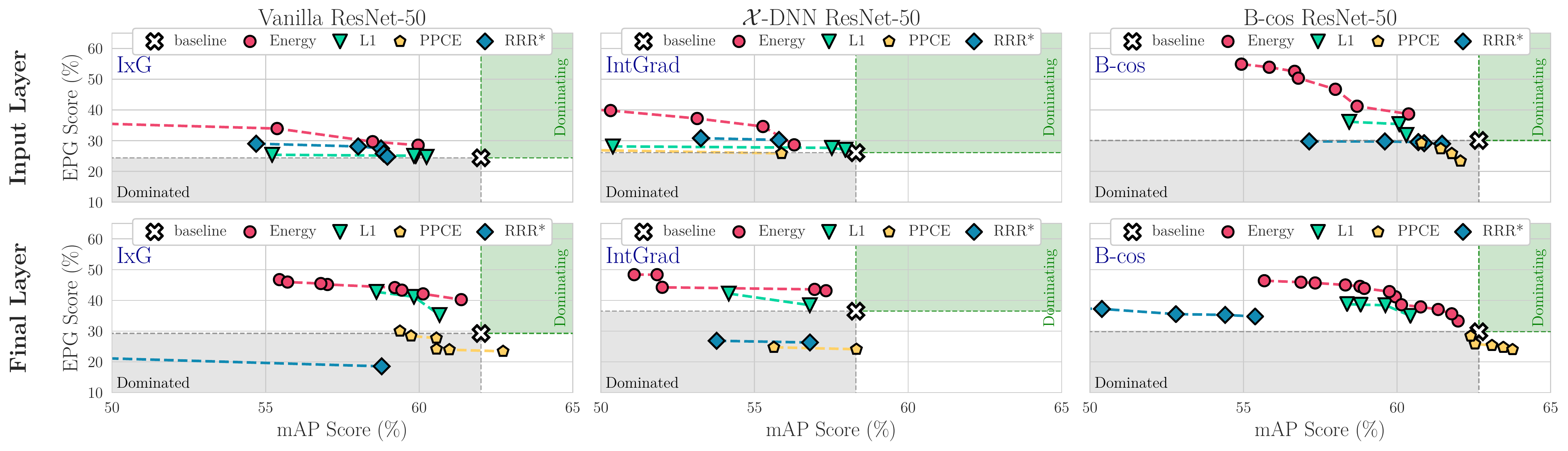}
    \caption{\textbf{\epg vs.~\map.}}
    \end{subfigure}
    \begin{subfigure}[c]{.9\textwidth}
    \includegraphics[width=\textwidth]{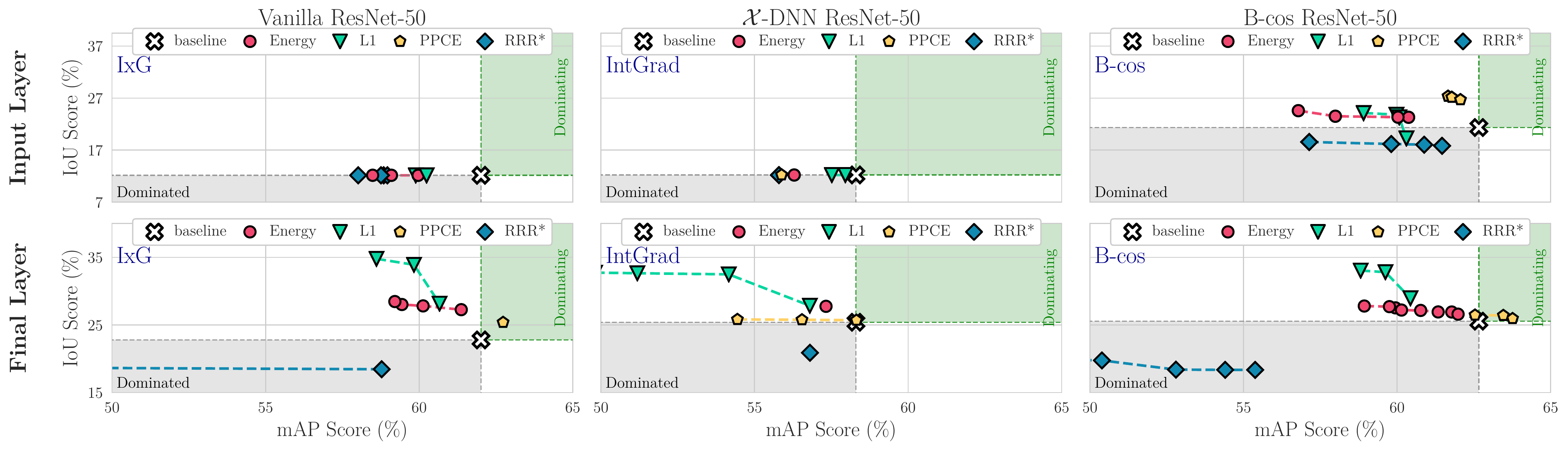}
    \caption{\textbf{\iou vs.~\map.}}
    \end{subfigure}
    \caption{\textbf{Quantitative comparison of \epg and \iou  vs.~\map scores for \cocos.} To ensure that the trends observed and described in the main paper generalize beyond the \fone metric, in this figure we show the \epg and \iou scores plotted against the \map metric. In general, we find the results obtained for the \map metric to be highly consistent with the previously shown results for the \fone metric, see \cref{fig:supp:coco:f1_results}. \Eg, across all configurations, we find the \epgloss to yield the highest gains in \epg score, whereas the \lone loss provides the best trade-offs with respect to the \iou metric. These results are further also consistent with those observed on \vocs, see \cref{fig:supp:voc:map_results:2}.}
    \label{fig:supp:coco:map_results}
\end{figure}
\begin{figure}[h]
    \centering    
    {\epg results for \textbf{intermediate layers} on \vocs.}\vspace{.25cm}\\
    \begin{subfigure}[c]{\textwidth}
    \includegraphics[width=\textwidth]{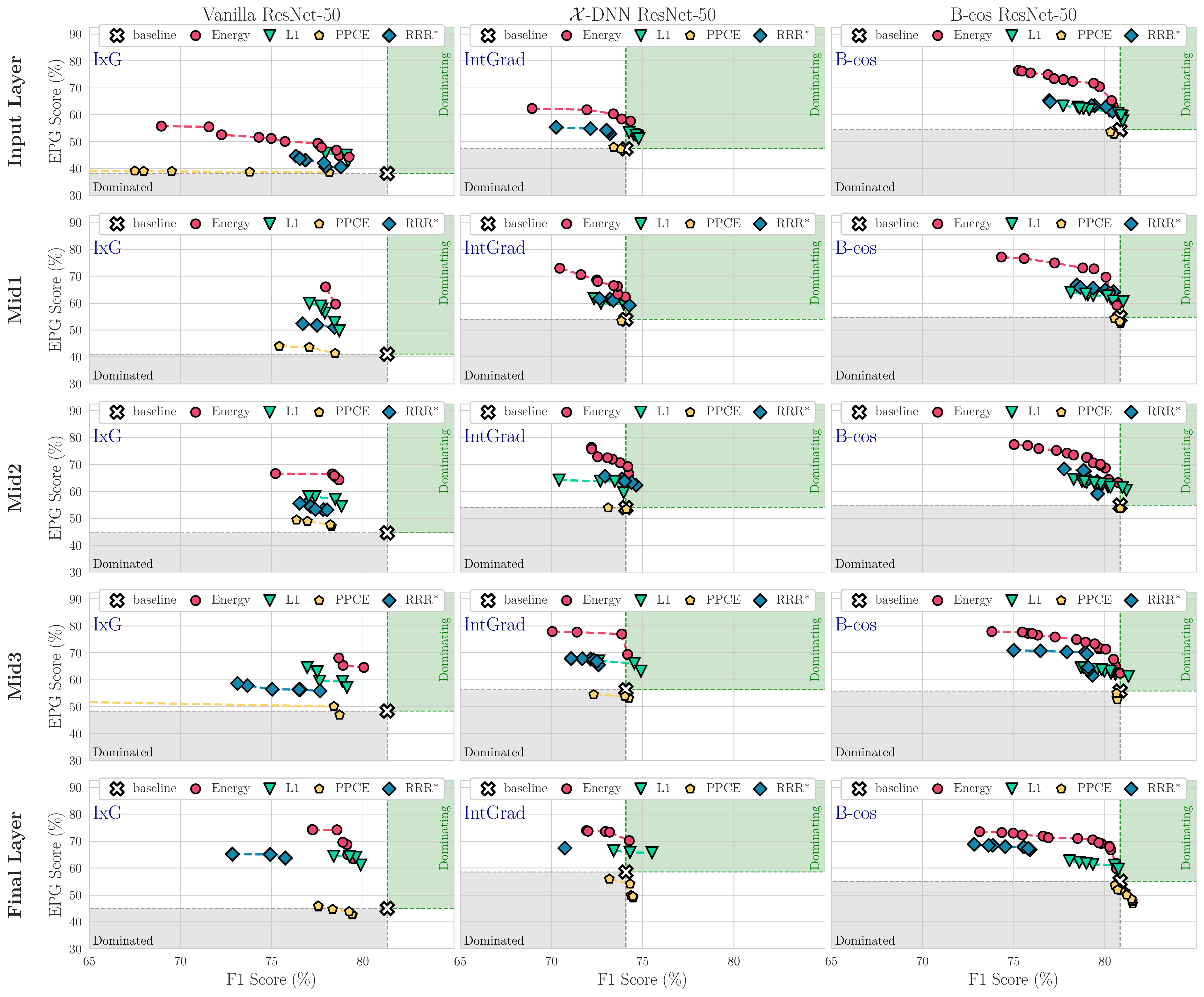}
    \end{subfigure}
    \caption{\textbf{Intermediate layer results comparing \epg vs.~F1.} We compare the effectiveness of model guidance at varying network depths (\textbf{rows}) for each attribution method and model (\textbf{columns}) across localization loss functions. For the \bcos model, we find similar trends at all network depths, with the \energyloss localization loss outperforming all other losses. For the \vanilla and \xdnn models, the \energyloss loss similarly performs the best, but we also observe improved performance across losses when optimizing at deeper layers of the network. Results for \iou can be found in \cref{fig:intermediate:iou}.}
    \label{fig:intermediate:epg}
\end{figure}
\begin{figure}[h]
    \centering    
    {\iou results for \textbf{intermediate layers} on \vocs.}\vspace{.25cm}\\
    \begin{subfigure}[c]{\textwidth}
    \includegraphics[width=\textwidth]{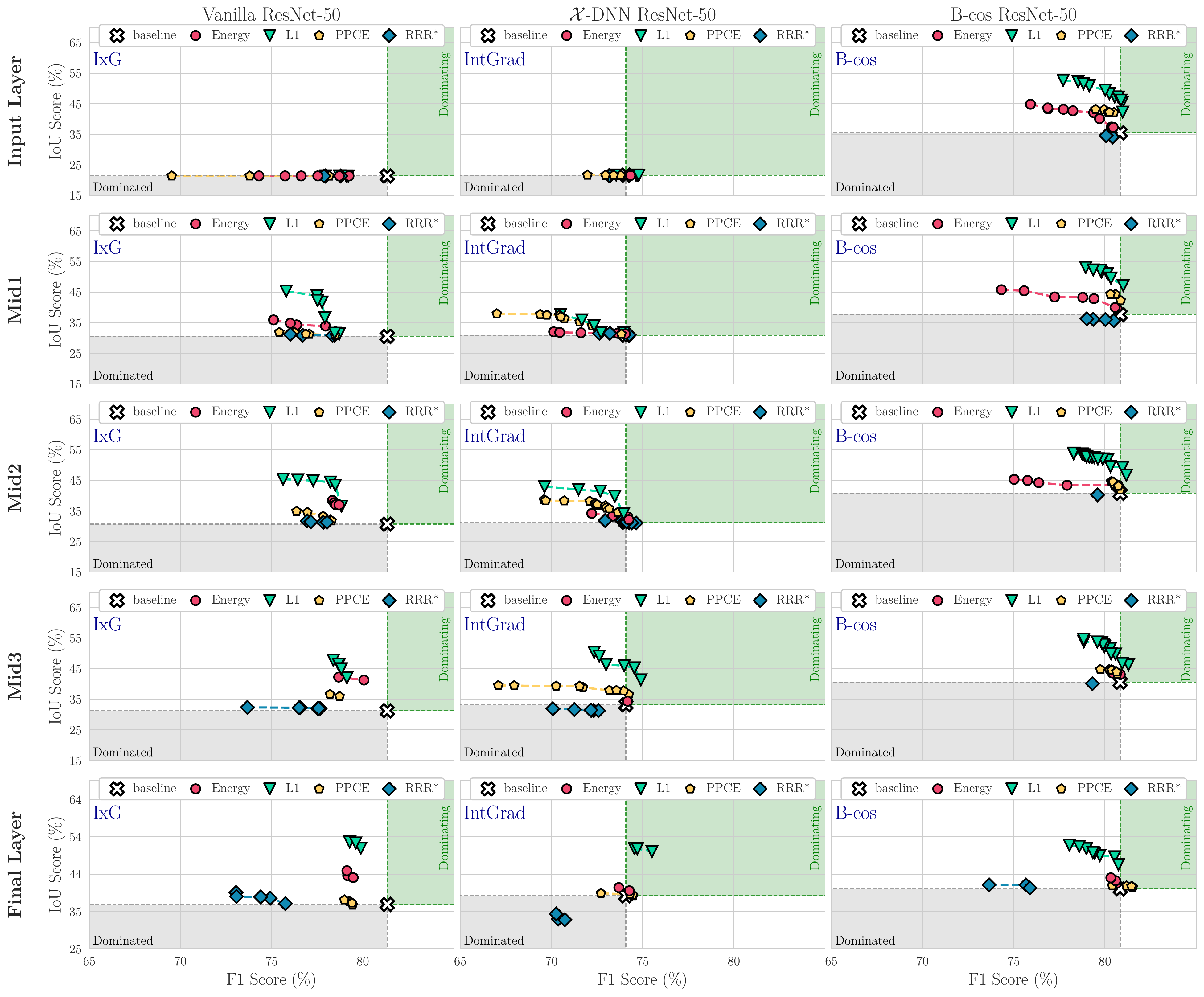}
    \end{subfigure}
    \caption{\textbf{Intermediate layer results comparing \iou vs.~F1.} We compare the effectiveness of model guidance at varying network depths (\textbf{rows}) for each attribution method and model (\textbf{columns}) across localization loss functions. We find similar trends across all configurations, with the \lone  loss outperforming all other losses. For the \vanilla and \xdnn models,  we observe improved performance across losses when optimizing at deeper layers of the network, whereas the results seem very stable for the \bcos models. For \epg results, see \cref{fig:intermediate:epg}.}
    \label{fig:intermediate:iou}
\end{figure}
\begin{figure}
    \centering
    {\large \textbf{Limited annotations --- Input layer}}\vspace{.5cm}\\
    \begin{subfigure}[c]{\textwidth}
    \begin{subfigure}[c]{.485\textwidth}
    \centering
    \textbf{\epg score}\\\vspace{.2cm}
    \includegraphics[width=\textwidth]{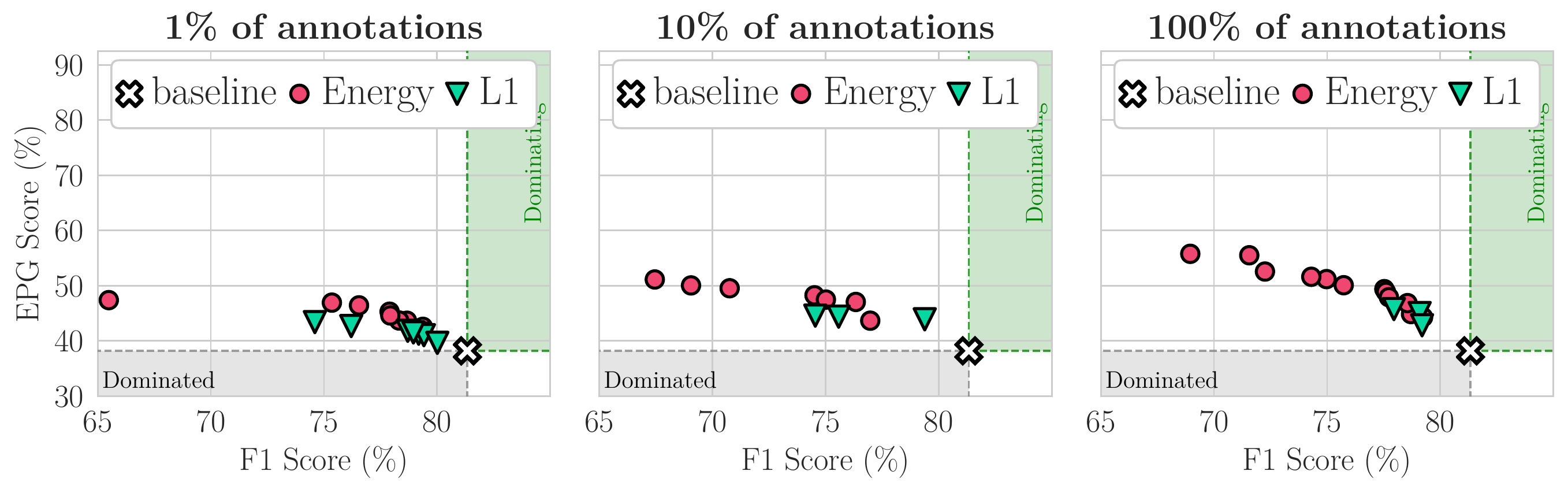}
    \end{subfigure}\hfill
    \begin{subfigure}[c]{.485\textwidth}
    \centering
    \textbf{\iou score}\\\vspace{.2cm}
    \includegraphics[width=\textwidth]{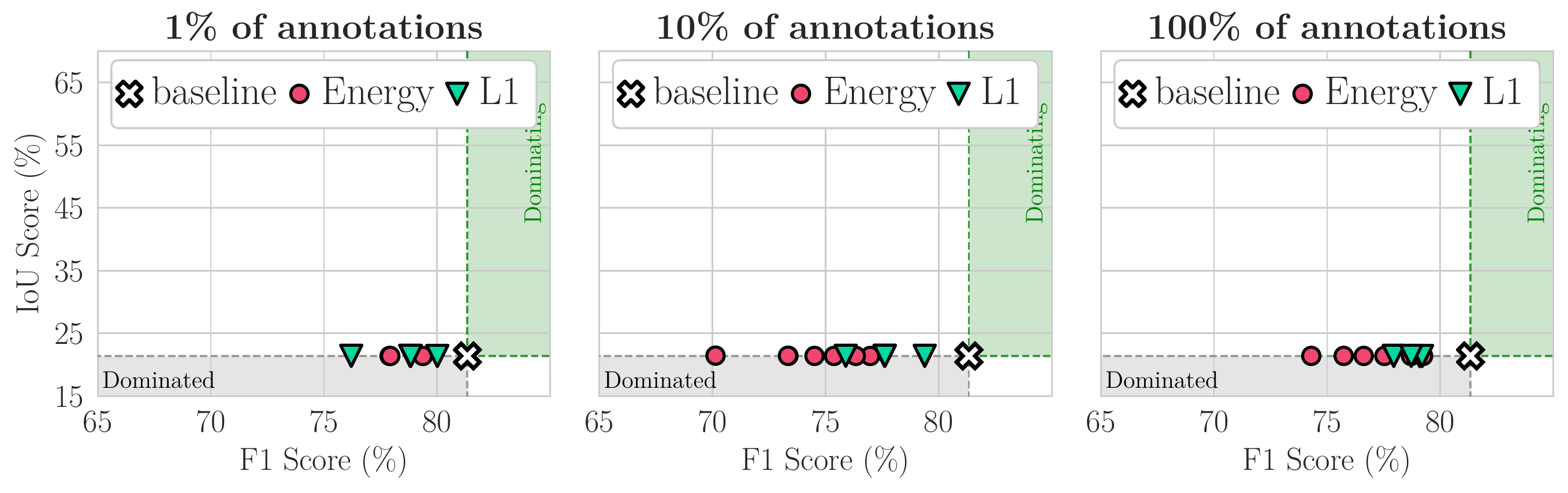}
    \end{subfigure}
    \caption{\textbf{\vanilla \resnet}}\vspace{.5cm}
    \end{subfigure}
    \begin{subfigure}[c]{\textwidth}
    \begin{subfigure}[c]{.485\textwidth}
    \centering
    \textbf{\epg score}\\\vspace{.2cm}
    \includegraphics[width=\textwidth]{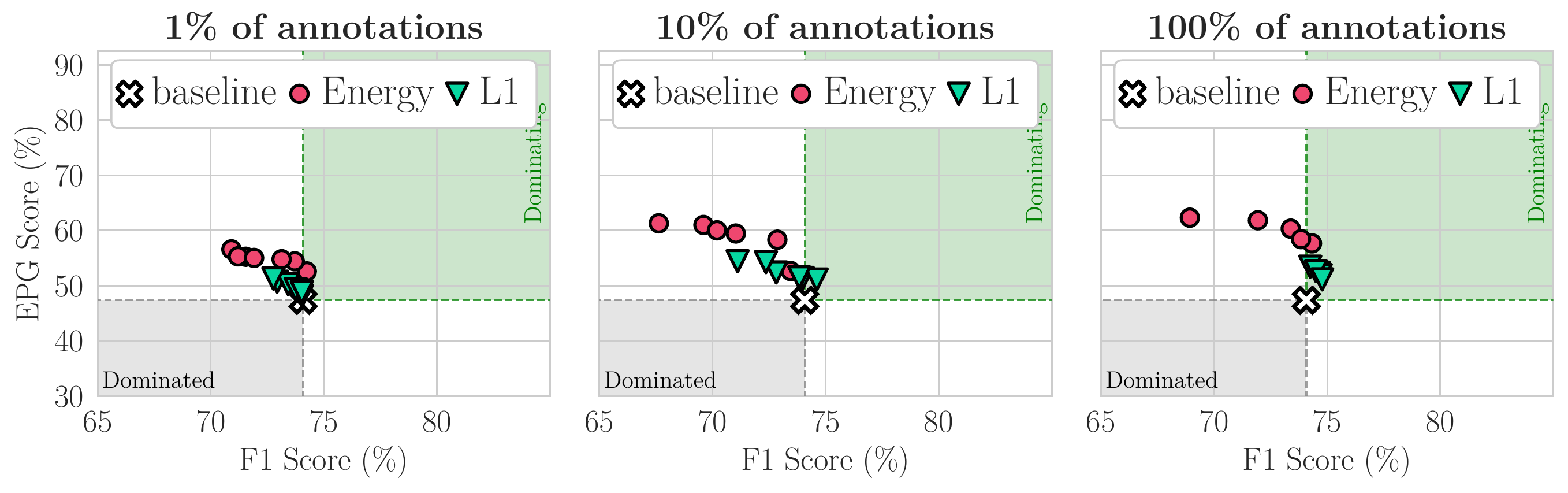}
    \end{subfigure}\hfill
    \begin{subfigure}[c]{.485\textwidth}
    \centering
    \textbf{\iou score}\\\vspace{.2cm}
    \includegraphics[width=\textwidth]{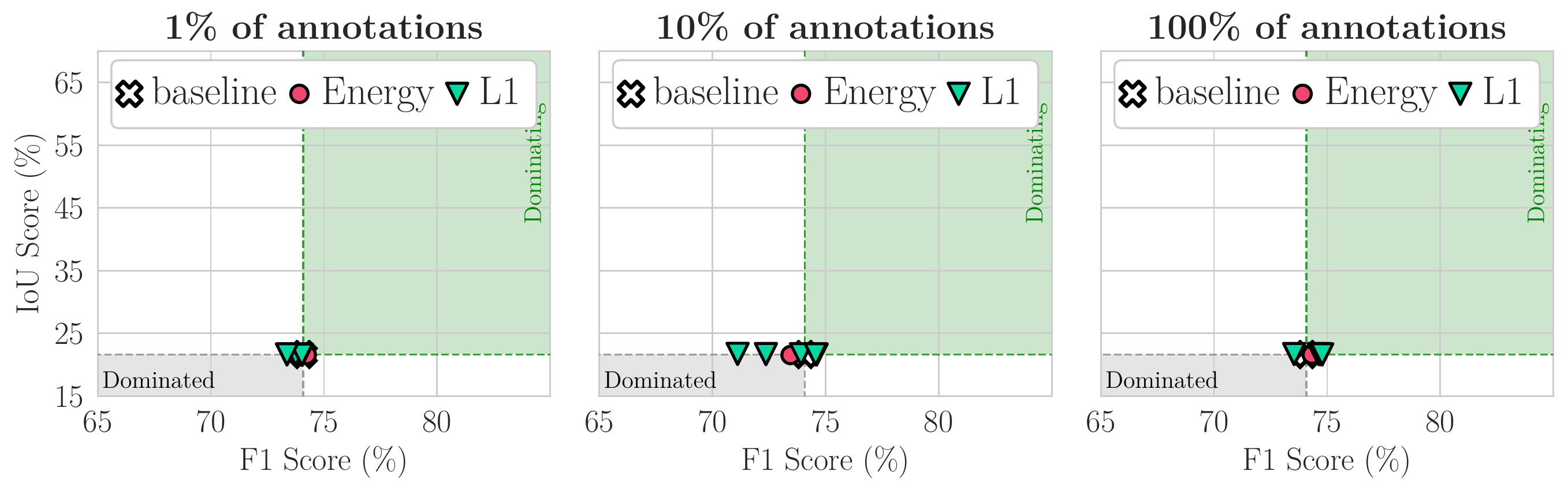}
    \end{subfigure}
    \caption{\textbf{\xdnn \resnet}}\vspace{.5cm}
    \end{subfigure}
    \begin{subfigure}[c]{\textwidth}
    \begin{subfigure}[c]{.485\textwidth}
    \centering
    \textbf{\epg score}\\\vspace{.2cm}
    \includegraphics[width=\textwidth]{results/VOC/figures/loc/input/limited_annotation_results_bcos.pdf}
    \end{subfigure}\hfill
    \begin{subfigure}[c]{.485\textwidth}
    \centering
    \textbf{\iou score}\\\vspace{.2cm}
    \includegraphics[width=\textwidth]{results/VOC/figures/iou/input/limited_annotation_results_bcos.pdf}
    \end{subfigure}
    \caption{\textbf{\bcos \resnet}}
    \end{subfigure}
    \caption{\textbf{\epg and \iou scores for model guidance at the input layer with a limited number of annotations.} We show \epg vs.~F1 (\textbf{left}) and \iou vs.~F1 (\textbf{right}) for all models, optimized with the \energyloss and \loneloss localization losses, when using $\{1\%,10\%,100\%\}$ training annotations. We find that model guidance is generally effective even when training with annotations for a limited number of images. While the performance slightly worsens when using 1\% annotations, using just 10\% annotated images yields similar gains to using a fully annotated training set. Results at the final layer can be found in \cref{fig:supp:limited:full:final}.}
    \label{fig:supp:limited:full:input}
\end{figure}
\begin{figure}[h]
    \centering
    {\large \textbf{Limited annotations --- Final layer}}\vspace{.5cm}\\
    \begin{subfigure}[c]{\textwidth}
    \begin{subfigure}[c]{.485\textwidth}
    \centering
    \textbf{\epg score}\\\vspace{.2cm}
    \includegraphics[width=\textwidth]{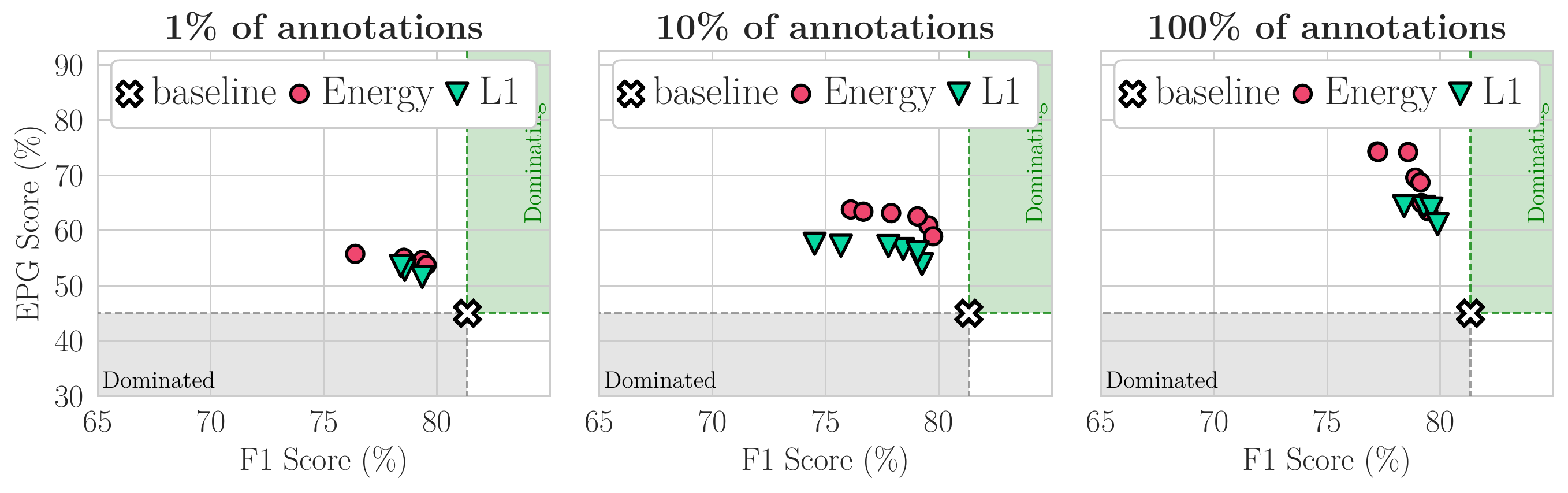}
    \end{subfigure}\hfill
    \begin{subfigure}[c]{.485\textwidth}
    \centering
    \textbf{\iou score}\\\vspace{.2cm}
    \includegraphics[width=\textwidth]{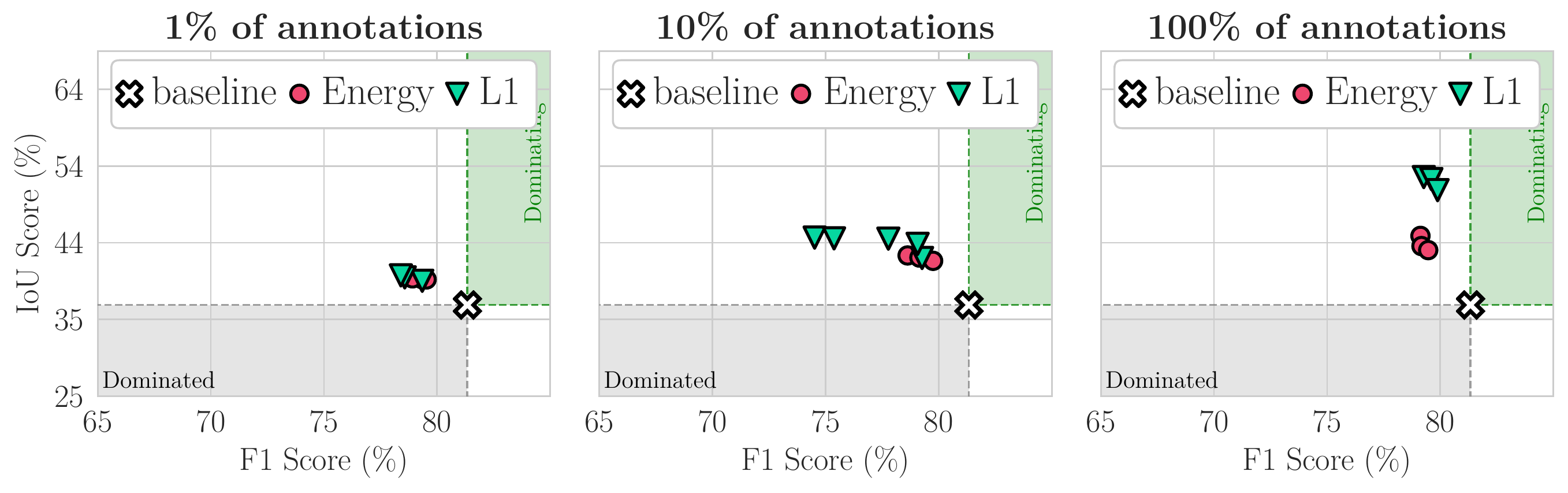}
    \end{subfigure}
    \caption{\textbf{\vanilla \resnet}}\vspace{.5cm}
    \end{subfigure}
    \begin{subfigure}[c]{\textwidth}
    \begin{subfigure}[c]{.485\textwidth}
    \centering
    \textbf{\epg score}\\\vspace{.2cm}
    \includegraphics[width=\textwidth]{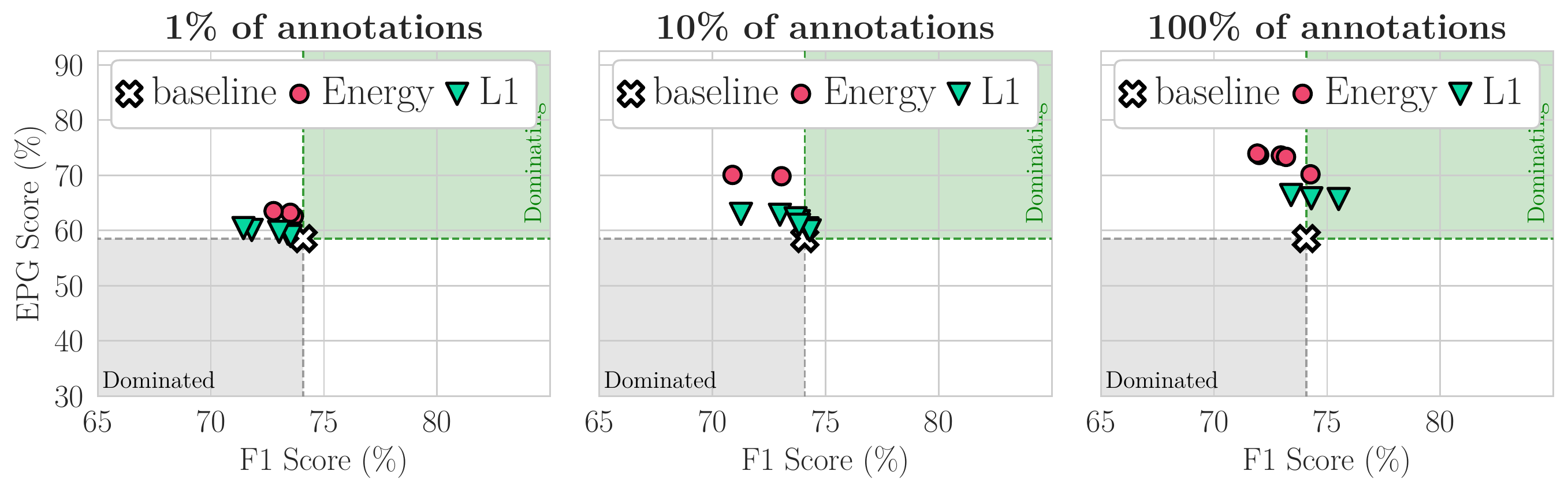}
    \end{subfigure}\hfill
    \begin{subfigure}[c]{.485\textwidth}
    \centering
    \textbf{\iou score}\\\vspace{.2cm}
    \includegraphics[width=\textwidth]{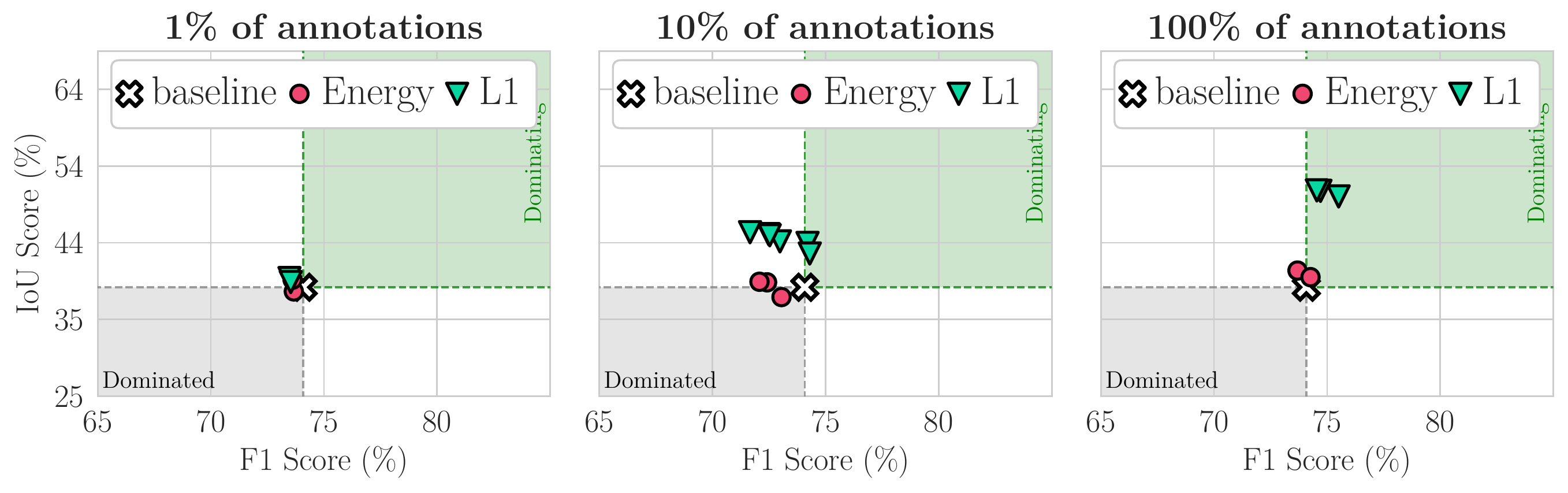}
    \end{subfigure}
    \caption{\textbf{\xdnn \resnet}}\vspace{.5cm}
    \end{subfigure}
    \begin{subfigure}[c]{\textwidth}
    \begin{subfigure}[c]{.485\textwidth}
    \centering
    \textbf{\epg score}\\\vspace{.2cm}
    \includegraphics[width=\textwidth]{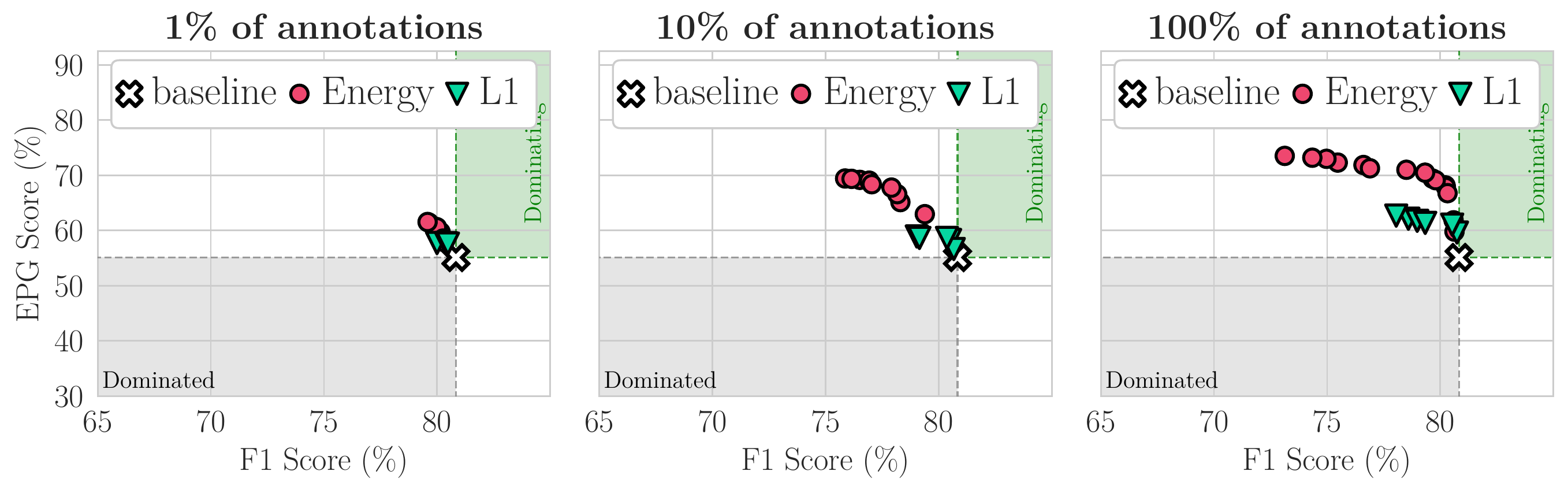}
    \end{subfigure}\hfill
    \begin{subfigure}[c]{.485\textwidth}
    \centering
    \textbf{\iou score}\\\vspace{.2cm}
    \includegraphics[width=\textwidth]{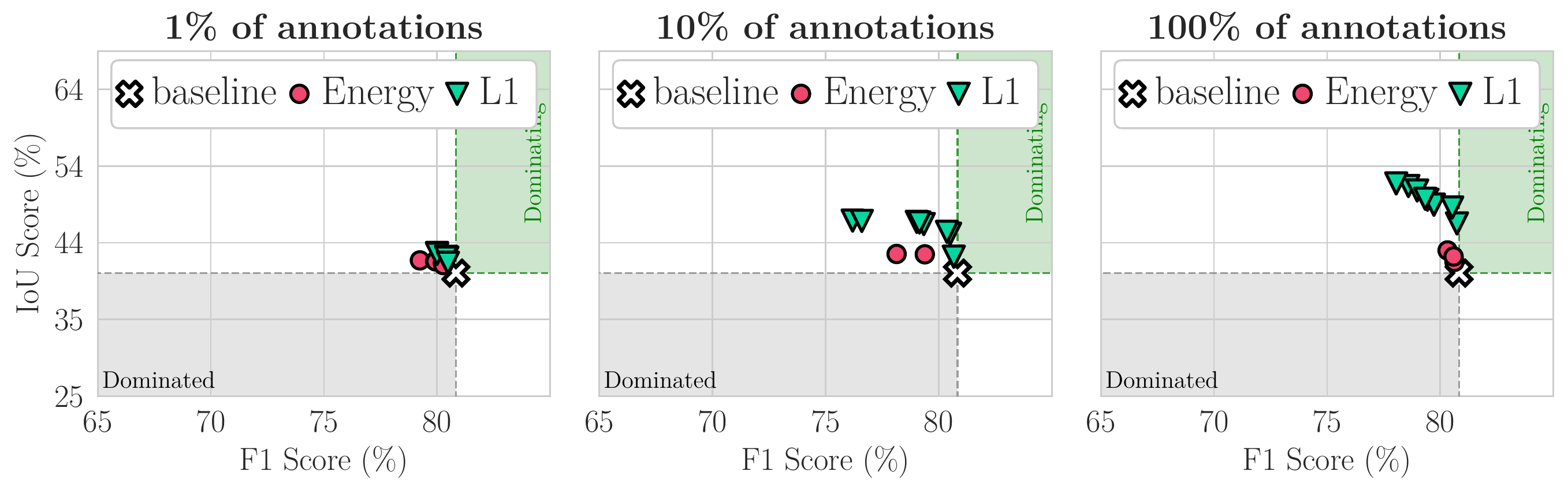}
    \end{subfigure}
    \caption{\textbf{\bcos \resnet}}
    \end{subfigure}
    \caption{\textbf{\epg and \iou scores for model guidance at the final layer with a limited number of annotations.} We show \epg vs.~F1 (\textbf{left}) and \iou vs.~F1 (\textbf{right}) for all models, optimized with the \energyloss and \loneloss localization losses, when using $\{1\%,10\%,100\%\}$ training annotations. We find that model guidance is generally effective even when training with annotations for a limited number of images. While the performance worsens when using 1\% annotations, using just 10\% annotated images yields similar gains to using a fully annotated training set. Results at the input layer can be found in \cref{fig:supp:limited:full:input}.}
    \label{fig:supp:limited:full:final}
\end{figure}
\begin{figure}
    \centering
    {Additional results for training with \textbf{coarse bounding boxes}}\vspace{.5cm}\\
    \begin{subfigure}[c]{.495\textwidth}
    \centering
    \includegraphics[width=\textwidth]{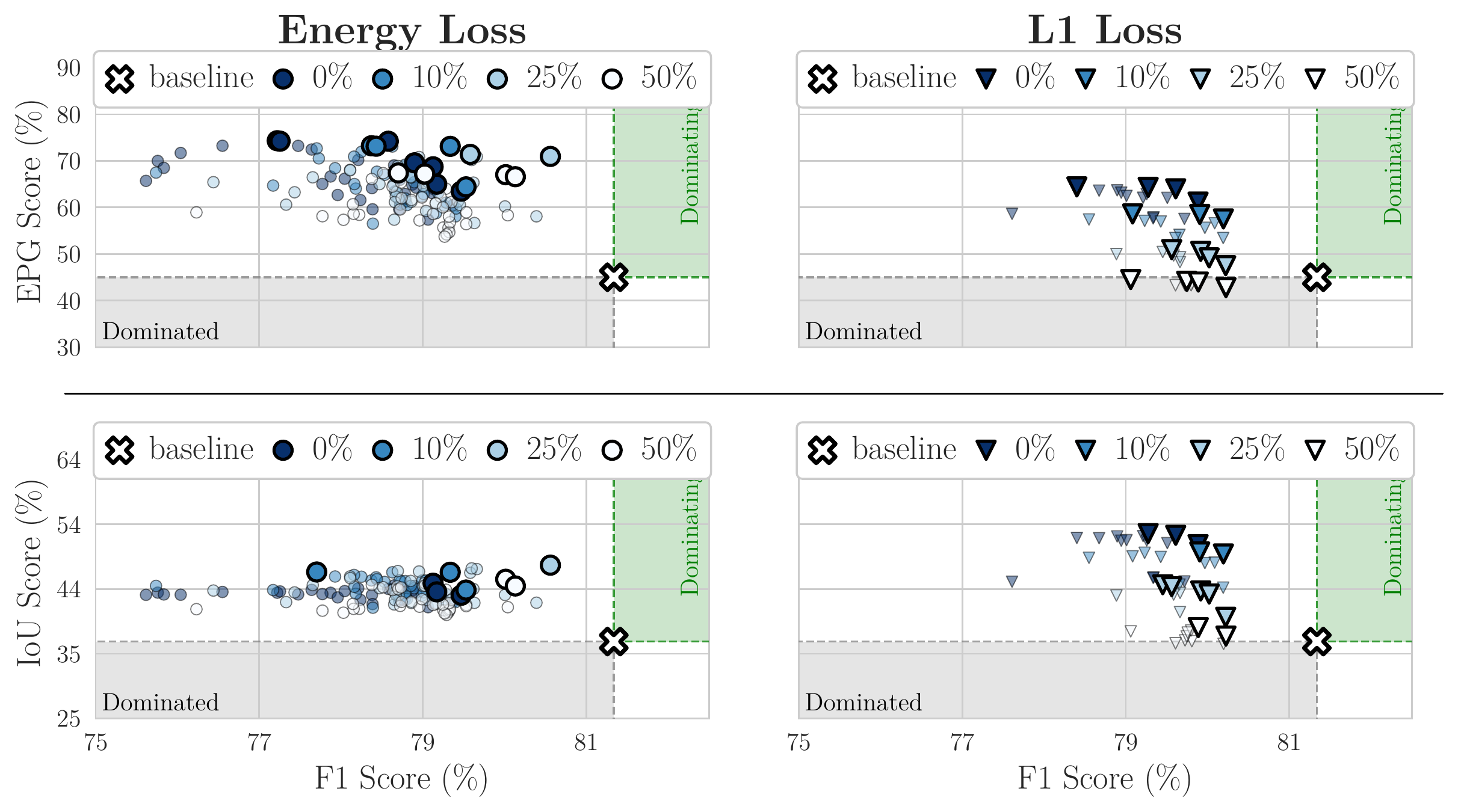}
    \caption{\textbf{\vanilla \resnet  @ Final.}}
    \end{subfigure}\hfill
    \begin{subfigure}[c]{.495\textwidth}
    \centering
    \includegraphics[width=\textwidth]{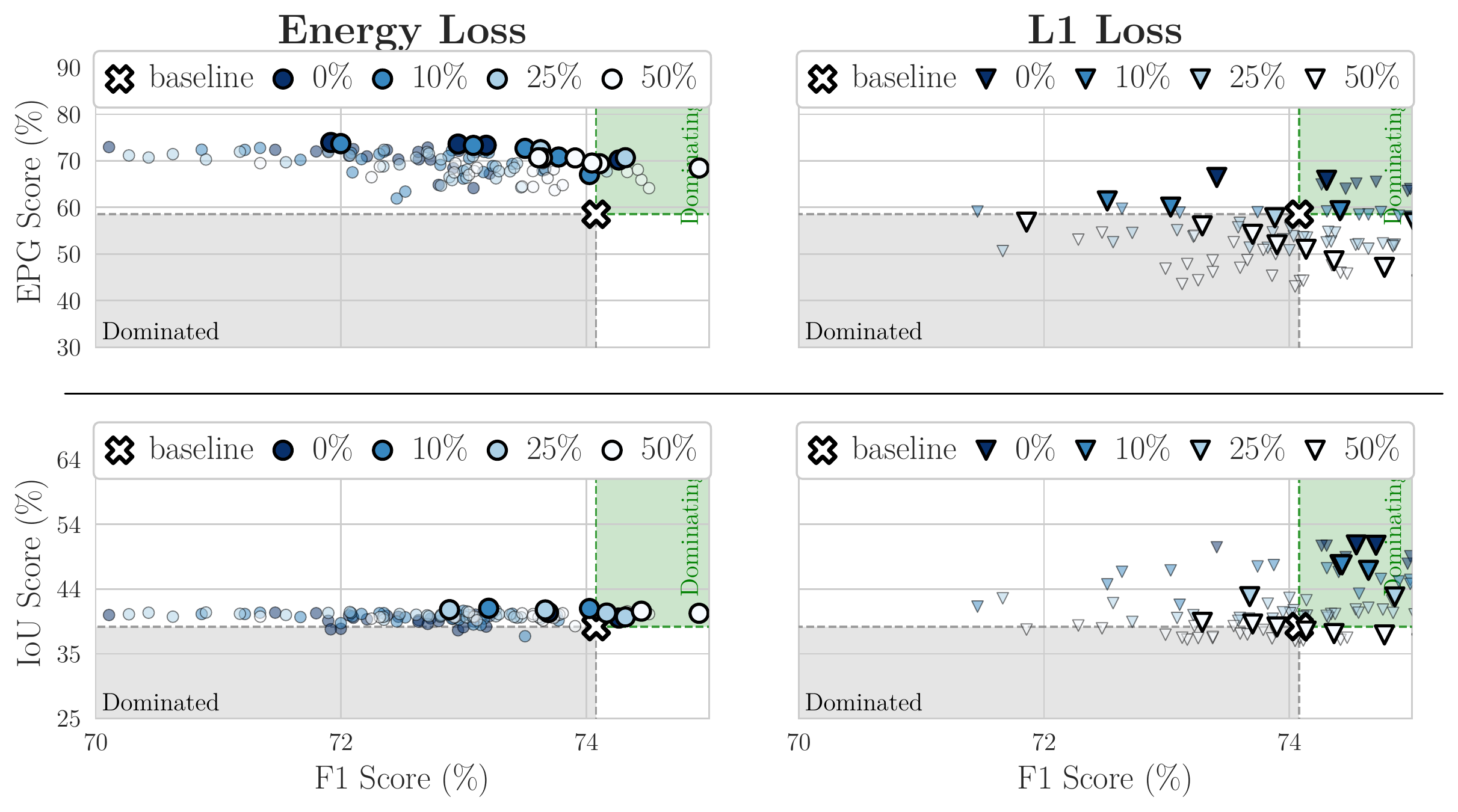}
    \caption{\textbf{\xdnn \resnet  @ Final.}}
    \end{subfigure}
    \begin{subfigure}[c]{.495\textwidth}
    \centering
    \includegraphics[width=\textwidth]{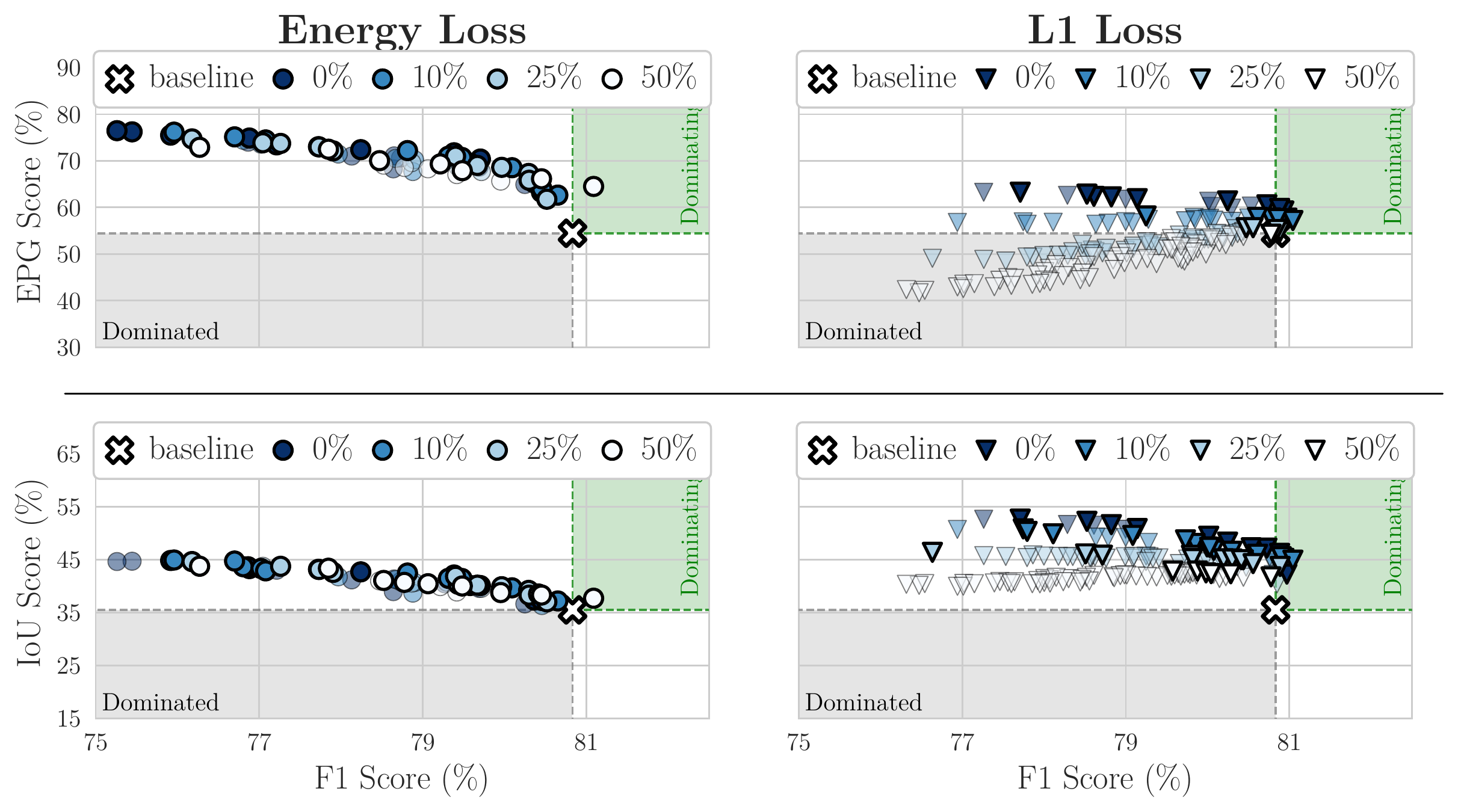}
    \caption{\textbf{\bcos \resnet @ Input.}}
    \end{subfigure}
    \caption{\textbf{Coarse bounding box results}. We show the impact of dilating bounding boxes during training for the \textbf{(a)} \vanilla and \textbf{(b)} \xdnn, and \textbf{(c)} \bcos models. Similar to the results seen with \bcos models (c), we find that the \energyloss localization loss is generally robust to coarse annotations, while the effectiveness of guidance with the \loneloss localization loss worsens as the extent of coarseness increases.   
    }
    \label{fig:supp:dilation_quanti:full}
\end{figure}

\clearpage

{\small
\bibliographystyleS{ieee_fullname}
\bibliographyS{references_supp}
}

\end{document}